\documentclass[10pt,twocolumn,letterpaper]{article}

\usepackage[table,pdftex]{xcolor}
\usepackage{iccv}
\usepackage{times}
\usepackage{epsfig}
\usepackage{amsmath}
\usepackage{amssymb}
\usepackage{algorithm,algpseudocode}
\usepackage{multirow,graphicx}

\usepackage{caption}
\usepackage{comment}
\usepackage{color}
\usepackage{subfigure}
\usepackage{gensymb}
\usepackage{soul}
\usepackage[title]{appendix}

\usepackage[pagebackref=true,breaklinks=true,letterpaper=true,colorlinks,bookmarks=false]{hyperref}

\iccvfinalcopy 

\definecolor{sgmcolor}{RGB}{10,110,210}

\definecolor{sarahcolor}{rgb}{0.858,0.188,0.478}

\definecolor{gscolor}{RGB}{10,240,10}

\definecolor{alcolor}{RGB}{240,10,10}

\definecolor{Gray}{gray}{0.9}

\makeatletter
\newcommand{\algmargin}{\the\ALG@thistlm}
\makeatother
\newlength{\whilewidth}
\algdef{SE}[parWHILE]{parWhile}{EndparWhile}[1]
  {\parbox[t]{\dimexpr\linewidth-\algmargin}{%
     \hangindent\whilewidth\strut\algorithmicwhile\ #1\ \algorithmicdo\strut}}{\algorithmicend\ \algorithmicwhile}%
\algnewcommand{\parState}[1]{\State%
  \parbox[t]{\dimexpr\linewidth-\algmargin}{\strut #1\strut}}

\makeatletter
\def\blfootnote{\xdef\@thefnmark{}\@footnotetext}
\makeatother

\ificcvfinal\fi

\begin{document}

\title{\vspace{-2.0cm}Formulating Camera-Adaptive Color Constancy \\ as a Few-shot  Meta-Learning Problem}




\author{Steven McDonagh\textsuperscript{*}, \,\,
Sarah Parisot\textsuperscript{*}, \,\,
Fengwei Zhou, \,\,
Xing Zhang, \,\, \\
Ales Leonardis, \,\,
Zhenguo Li, \,\,
Gregory Slabaugh \\
Huawei Noah's Ark Lab\\ 
{\tt\small \{steven.mcdonagh, sarah.parisot, zhoufengwei, zhang.xing1} \\
{\tt\small ales.leonardis, li.zhenguo, greg.slabaugh\}} \\
{\tt\small @huawei.com} } 

\maketitle

\blfootnote{* Authors contributed equally}

\begin{abstract}




Digital camera pipelines employ color constancy methods to estimate an unknown scene illuminant, in order to re-illuminate images as if they were acquired under an achromatic light source. Fully-supervised learning approaches exhibit state-of-the-art estimation accuracy with camera-specific labelled training imagery. Resulting models typically suffer from domain gaps and fail to generalise across imaging devices. In this work, we propose a new approach that affords fast adaptation to previously unseen cameras, and robustness to changes in capture device by leveraging annotated samples across different cameras and datasets. We present a general approach that utilizes the concept of color temperature to frame color constancy as a set of distinct, homogeneous few-shot regression tasks, each associated with an intuitive physical meaning. We integrate this novel formulation within a meta-learning framework, enabling fast generalisation to previously unseen cameras using only handfuls of camera specific training samples. Consequently, the time spent for data collection and annotation substantially diminishes in practice whenever a new sensor is used. To quantify this gain, we evaluate our pipeline on three publicly available datasets comprising $12$ different cameras and diverse scene content. Our approach delivers competitive results both qualitatively and quantitatively while requiring a small fraction of the camera-specific samples compared to standard approaches.

\end{abstract}

\section{Introduction}
\label{sec:intro}

The colors of an image captured by a digital camera are always affected by the prevailing light source color in the scene. Accounting for the effect of scene illuminant to produce images of canonical appearance (as if captured under an achromatic light source) is an essential component of digital photography pipelines, and is of great importance for many practical high-level computer vision applications including image classification, semantic segmentation and machine vision quality control~\cite{ramanath2005color,vrhel2005color,funt1998machine}.
Such applications commonly require that input images are device independent and illuminant color-unbiased. Extraction of the intrinsic color information from scene surfaces by compensating for scene illuminant color is commonly referred to as ``Color Constancy'' (CC) or ``Automatic White Balance'' (AWB).
The process of computational CC can be defined as the transformation of the source image, captured under an unknown illuminant, to a target image representing the same scene under a canonical illuminant. CC algorithms typically consist of two stages; first, estimation of the scene illuminant color and second, transformation of the source image, accounting for the illuminant, such that the resulting image illumination appears achromatic. 


\begin{figure}[t]
\centering
    \subfigure[Input image]{
    \label{fig:sec1:example_result:subfig1}
    \includegraphics[width=0.45\linewidth]{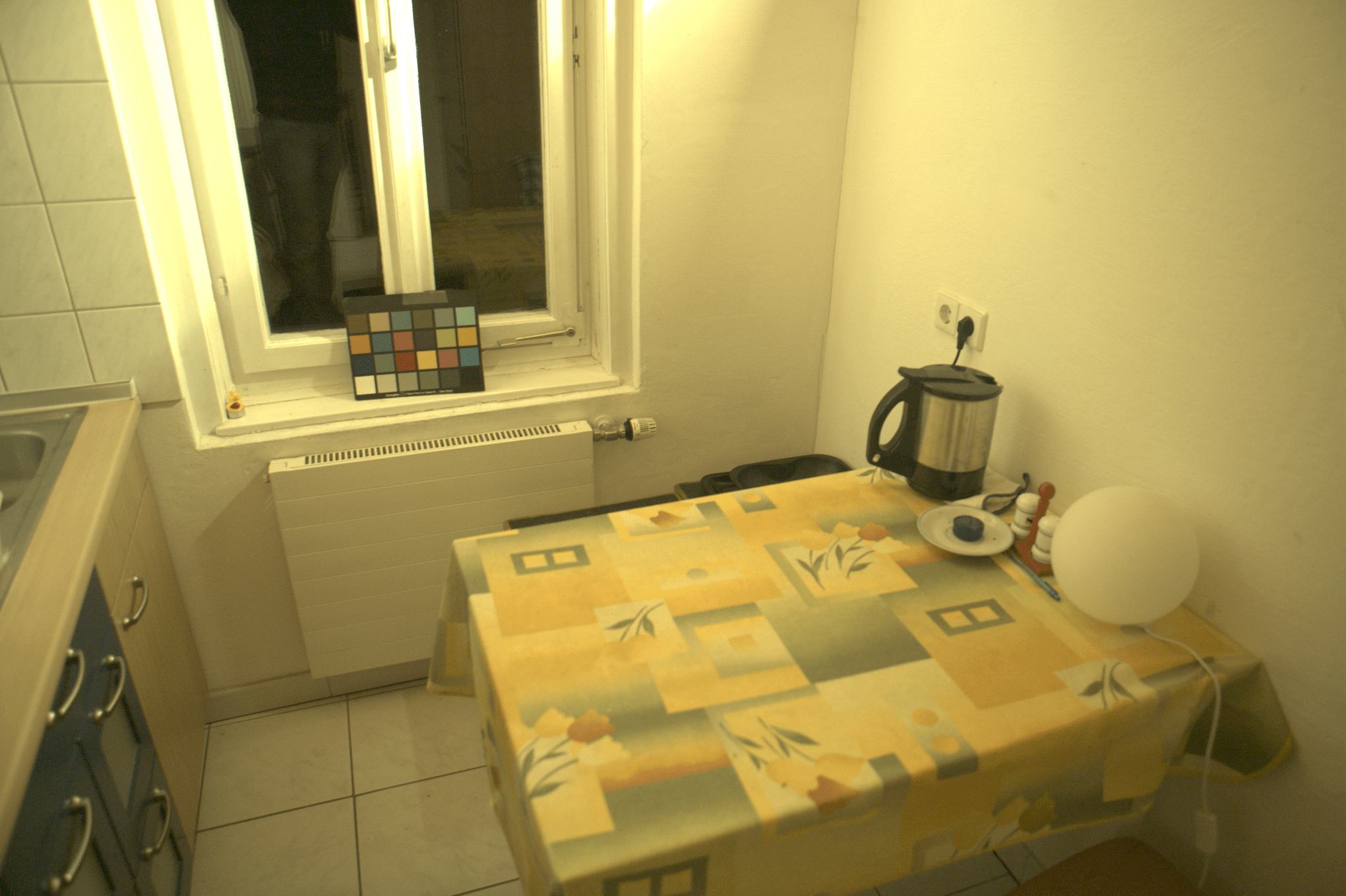}
    }
    \subfigure[Ground-truth]{
    \label{fig:sec1:example_result:subfig2}
    \includegraphics[width=0.45\linewidth]{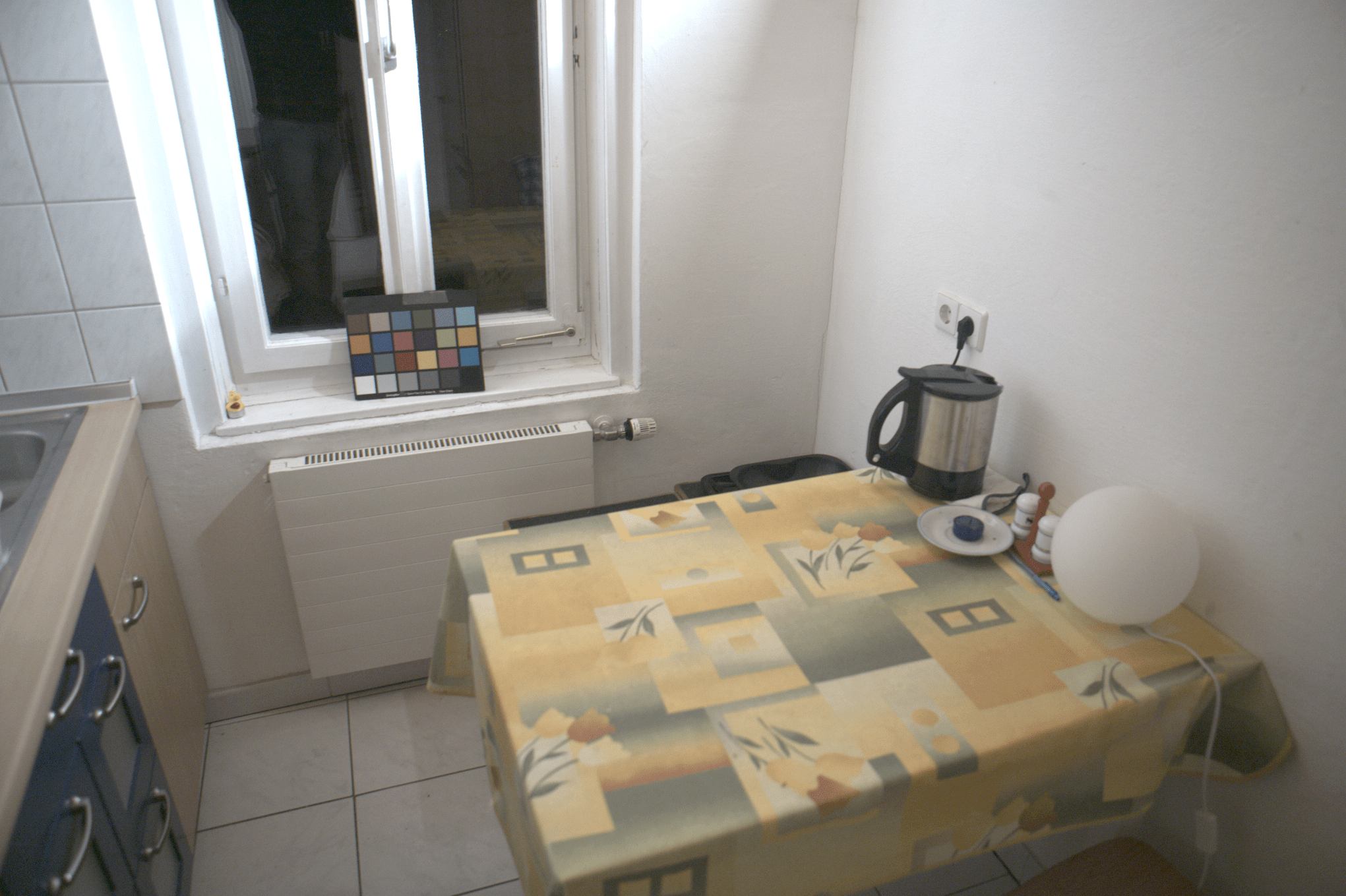}
    }\\
    \subfigure[Standard fine tuning of a pre-trained model with access to only $10$~images from a test camera.]{
    \label{fig:sec1:example_result:subfig4}
    \includegraphics[width=0.45\linewidth]{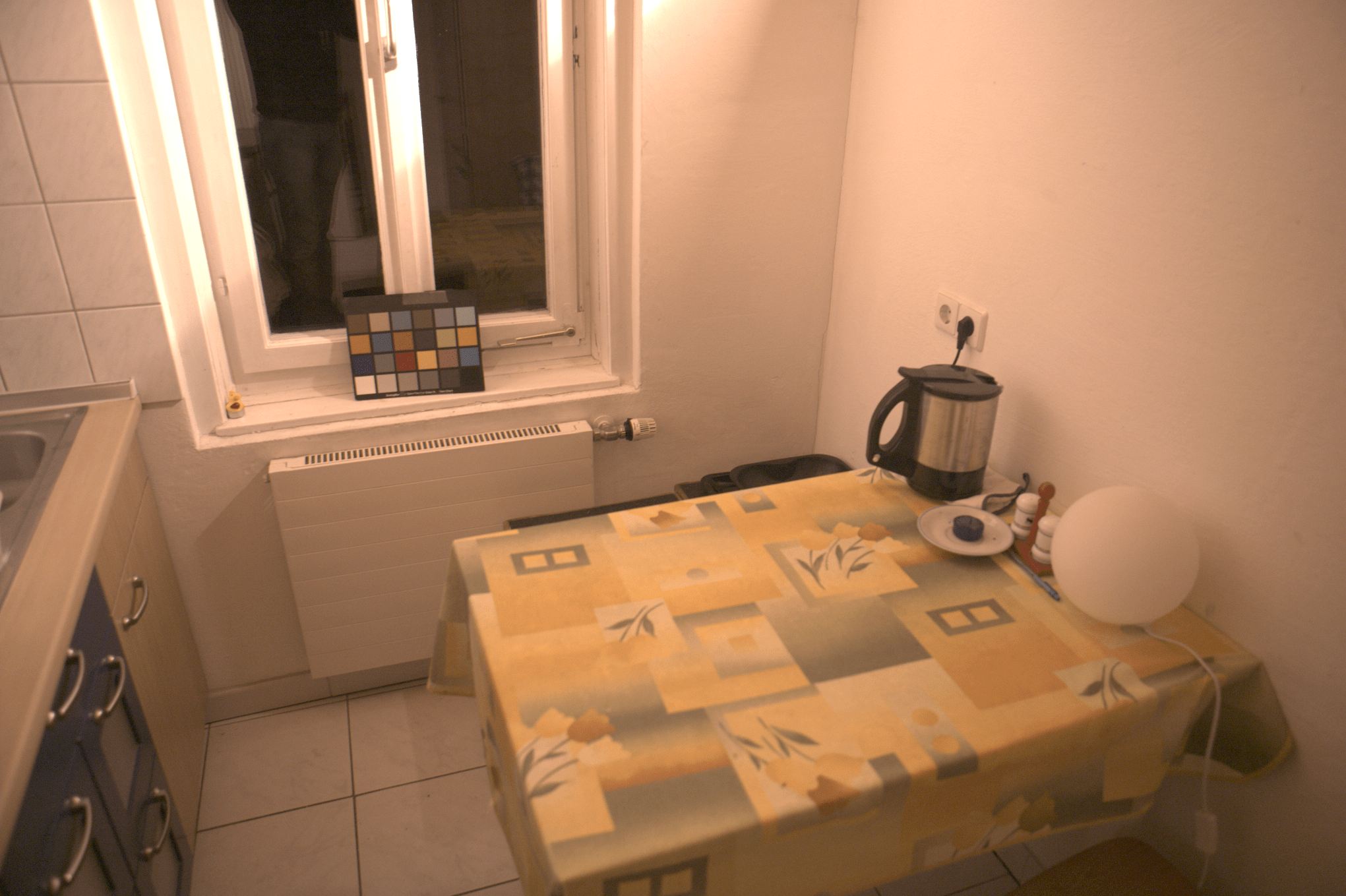}
    }
    \subfigure[Ours, $k$-shot learning (10~images from a test camera)]{
    \label{fig:sec1:example_result:subfig3}
    \includegraphics[width=0.45\linewidth]{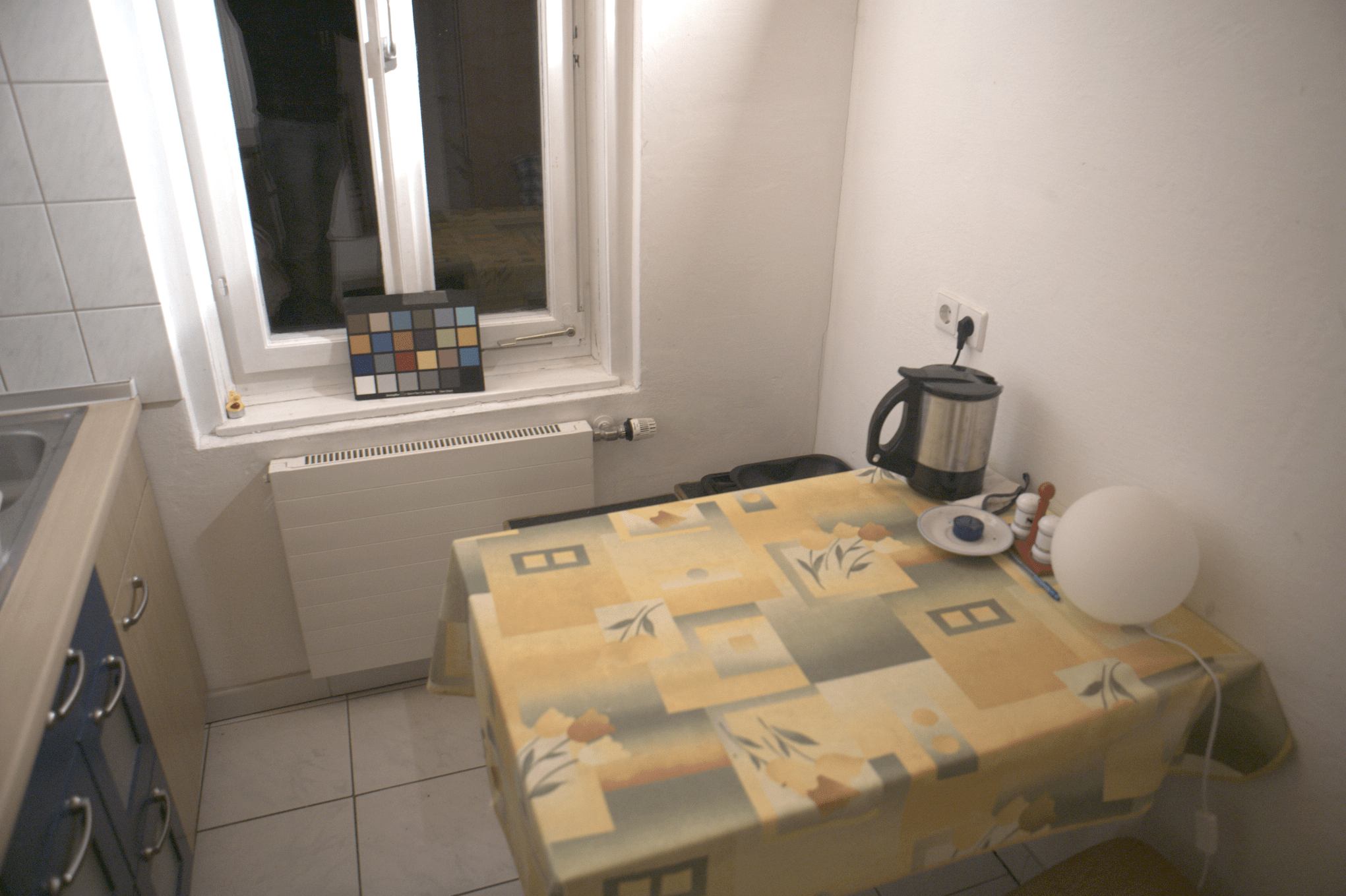}
    }
\caption{
An example image before and after color constancy correction. Our approach can quickly adapt to new unseen camera sensors using few samples where a pre-trained model, fine-tuned naively, fails to adapt well.   
}
\label{fig:sec1:example_result}
\end{figure}

\begin{figure*}[t]
\begin{center}
\includegraphics[width=0.95\linewidth,trim={0cm 0.5cm 0cm 0.5cm}, clip]{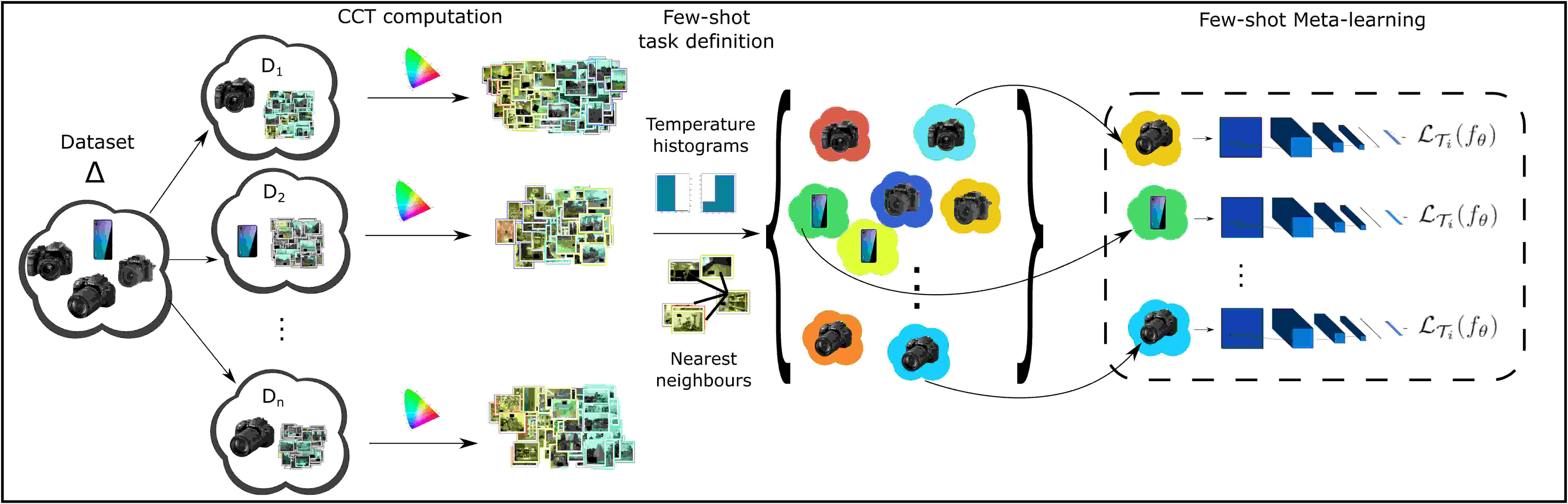}
\end{center}
\caption{Overview of the proposed strategy defining task distribution $\mathcal{T}_i \sim p(\mathcal{T})$. Considering a set of cameras and camera specific images, we separate images into subtasks based on illuminant color. This is done by computing color temperature for each image, and building a CCT histogram for each camera. Images in the same task are defined as images captured using the same camera and belonging to the same CCT histogram bin.}
\label{fig:sec:1:overview}
\end{figure*}

The perceived color of surfaces are determined by the intrinsic surface reflectance properties of objects in the scene, the spectral power distribution of the light(s) illuminating them and physical capture device properties encompassing image sensor, camera spectral sensitivity (CSS) and lens effects. 
This combination of properties makes the problem locally underdetermined. 
In practice, if no device calibration prior is available, we can only observe a product of these factors, as measured in the digital image. More formally, we model a tri-chromatic photosensor response in the standard way such that:
\begin{equation}
    \rho_k({X}) = \int_{\Omega} E(\lambda) S(\lambda, X) R_k(\lambda) d \lambda \quad k \in \{R, G, B\}.
    \label{eq:per_pixel_appearance}
\end{equation}
where $\rho_k(X)$ is the intensity of color channel $k$ at pixel location $X$, the wavelength of light $\lambda$ such that $E(\lambda)$ represents the spectrum of the illuminant, $S(\lambda, X)$ the surface reflectance at pixel location $X$ and $ R_k(\lambda)$ the CSS for channel $k$, considered over the spectrum of visible wavelengths $\Omega$. 
The goal of computational CC then becomes estimation of the global illumination color $\rho^E_k$ where:

\begin{equation}
    \rho^E_k = \int_{\Omega} E(\lambda) R_k(\lambda) d \lambda \hspace{5mm} k \in \{R, G, B\}.
    \label{eq:illuminant_eq}
\end{equation}

Given that there exist infinitely many combinations of illuminant color and surface reflectance that result in identical (pixel value) observations $\rho_k(X)$, the problem of finding $\rho^E_k$ is (locally) ill-posed. 

Modern supervised learning techniques can be used to infer this global image illuminant color and currently provide state-of-the-art estimation accuracy~\cite{barron2017fast}. However, such approaches are typically CSS specific (i.e. consistent $R_k(\lambda)$) and therefore require, for each camera considered, acquisition of large sets of manually labelled images comprising a variety of scenes and illumination colors. This poses a barrier preventing such tools from providing highly accurate and robust illuminant estimation for new, previously unseen cameras in a manner that can be regarded as both quick and cheap.



In this paper we propose a new approach that removes the expensive, yet necessary for standard approaches, requirement of large amounts of labelled, sensor-specific image acquisition 
by decomposing the illuminant estimation problem such that it is robust with respect to variation in capture device. Using the concept of color temperature to infer the nature of scene light source, we frame the CC problem as a set of related yet distinct few-shot regression tasks, where each task is camera and illuminant specific. This enables us
to exploit small image datasets, captured from disparate sources, and construct models with a capacity to learn camera-specific color biases quickly and cheaply using only a handful of target-device labelled images. Integrating our task definition approach within a meta-learning framework~\cite{finn2017model}, we are able to train a joint model capable of quickly adapting to new unseen capture devices and report performance competitive with the fully-supervised,  state of the art using a handful of camera-specific training samples. An overview of the proposed approach is depicted in Fig.~\ref{fig:sec:1:overview}.






The main contributions of this work are:
\begin{enumerate}
\item Our work constitutes the first few-shot learning approach for color constancy and enables the use of order(s) of magnitude fewer device-specific training images in comparison to contemporary work. 

\item We introduce color temperature to the CC problem, demonstrate how it allows to estimate the type of light source from a photograph, and use it to frame CC as a set of simpler physically intuitive  problems.




\item We provide extensive experiments on three public datasets and provide a comparative analysis of three meta-learning algorithm variants on a real-world image regression problem.
\end{enumerate}

\section{Related Work}
\label{sec:lit_review}



Our contributions are closely related to previous learning based color constancy work, inter-camera considerations and few-shot learning techniques. We now provide brief review of these topics.

\noindent\textbf{Fully supervised methods.}
Prior work 
can broadly be divided into statistics-based and learning-based methods~\cite{gijsenij2011computational}. Classical methods utilise low-level statistics that are fast and typically contain few free parameters. However, performance is highly dependent on strong scene content assumptions and these methods falter in cases where assumptions fail to hold.
Early learning-based work~\cite{funt2004estimating,xiong2006estimating,wang2009edge,rosenberg2004bayesian,gehler2008bayesian} comprised of combinational and direct approaches, typically relying on hand-crafted image features which limited their overall performance. 

Recent fully supervised convolutional CC work now offers state-of-the-art estimation accuracy. Both local patch-based~\cite{bianco2015color,shi2016deep,bianco2017single,hu2017fc4} and full image~\cite{barron2015convolutional,lou2015color,barron2017fast} input have been considered, investigating different model architectures~\cite{bianco2015color, bianco2017single,shi2016deep} and the use of semantic information~\cite{hu2017fc4, lou2015color}. 
Barron~\cite{barron2015convolutional, barron2017fast} alternatively frames computational CC as a 2D spatial localisation problem. He represents image data using log-chroma histograms for which a single convolutional layer learns to evaluate illuminant color solutions in the chroma plane. 
Despite strong performance, fully supervised deep-learning techniques require large amounts of calibrated and hand-labelled sensor specific data to learn robust models for each target device~\cite{aytekin2018data}. This makes collection and calibration of imagery for data driven color constancy both restrictive and costly, commonly requiring placement of physical calibration objects in a large variety of scenes and illuminants, and subsequent manual segmentation to measure ground-truth illuminants.  


Image augmentation, \emph{eg.} synthetic relighting~\cite{bianco2017single}, and transfer-learning~\cite{bianco2015color} using models pre-trained for alternative tasks, \emph{eg.} ImageNet classification~\cite{krizhevsky2012imagenet}, have been previously employed to mitigate lack of available data. The former strategy commonly struggles with synthetic-data domain gap issues and may not generalise well to real-world image manifolds at inference time, while 
the misalignment between object classification and computational CC likely results in learning features invariant to appearance attributes of critical importance for CC, limiting the performance of the latter strategy.  

\noindent\textbf{Inter-camera and unsupervised approaches.}
Few color constancy works have attempted to mitigate the costs of sensor-specific data collection, calibration and image labelling. 
The work of~\cite{gao2017improving} learns a transformation between pairs of camera CSS, but requires a priori knowledge of the cameras' CSS and is limited to pairs of sensors. 
Early unsupervised work~\cite{tieu2003unsupervised} introduces a linear statistical model learned on a single sensor from video frame observations. Banic et al.~\cite{banic2018unsupervised} use classical statistical approaches to learn parameters that approximate the unknown ground-truth illumination of the training images, avoiding calibration and image labelling. Despite promising inter-camera performance ~\cite{banic2018unsupervised}, unsupervised techniques still require the collection of a large amount of unlabelled images under varying light sources, and yield subpar performance when compared to fully-supervised methods. 



Our approach proposes to strongly restrict the data requirements via a few-shot formulation that is robust to variations in camera sensor, and bridges the gap between fully-supervised and unsupervised performances.  


\noindent\textbf{Few-shot Learning.}
Few-shot learning problems consist of learning a new task or concept using only a handful of data points (typically $1$-$10$ samples per task) and have recently received considerable attention~\cite{vinyals2016matching,ravi2017optimization,snell2017prototypical, finn2017model}. This promises a number of advantages with regard to efficient model building for new tasks; reducing the need for data acquisition and labelling by order(s) of magnitude, and decreasing effort spent on fine-tuning and adaptation of existing models to novel problems. A popular meta-learning strategy consists of finding model initialisations that allow fast adaptation to new, previously unseen tasks~\cite{finn2017model}. 
The strategy has since been widely adopted for classification tasks ~\cite{jamal2018task} and several recently proposed extensions report increases to efficiency~\cite{nichol2018first} and performance~\cite{li2017meta, antoniou2018train}. While a natural separation of few-shot tasks exists for image classification problems, in contrast, problems framed as a regression (e.g. image illuminant estimation), require a careful task definition process so as to provide distinct yet homogeneous tasks that aid fast and accurate model adaption to new problem instances using limited training data. 

To take advantage of such tools for color constancy, an important research question emerges, namely, how to decompose our problem in a set of few-shot tasks? This is crucially important when camera specific data is sparse. 
\section{Camera-Adaptive Color Constancy}
\label{sec:method}

\begin{figure}[!t]
\begin{center}
\subfigure[]{
    \label{fig:planck}
    \includegraphics[width=0.42\linewidth]{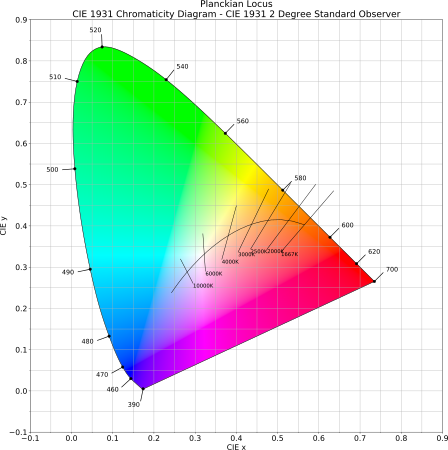}
    }
    \subfigure[]{
    \label{fig:scale}
    \includegraphics[width=0.24\linewidth]{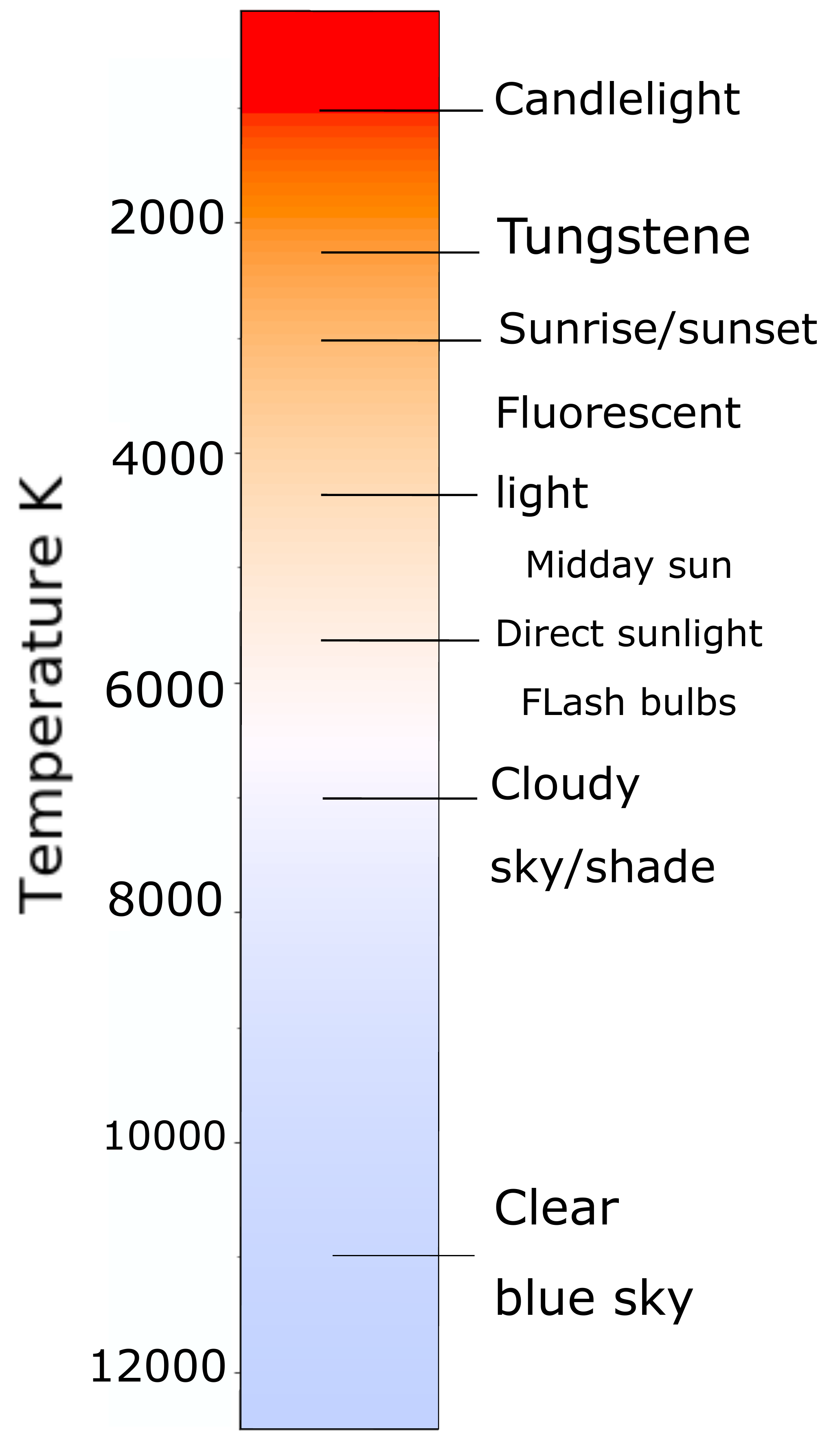}
    }
    \subfigure[]{
    \label{fig:img}
    \includegraphics[width=0.24\linewidth,trim={8cm 2.5cm 8cm 3cm},clip]{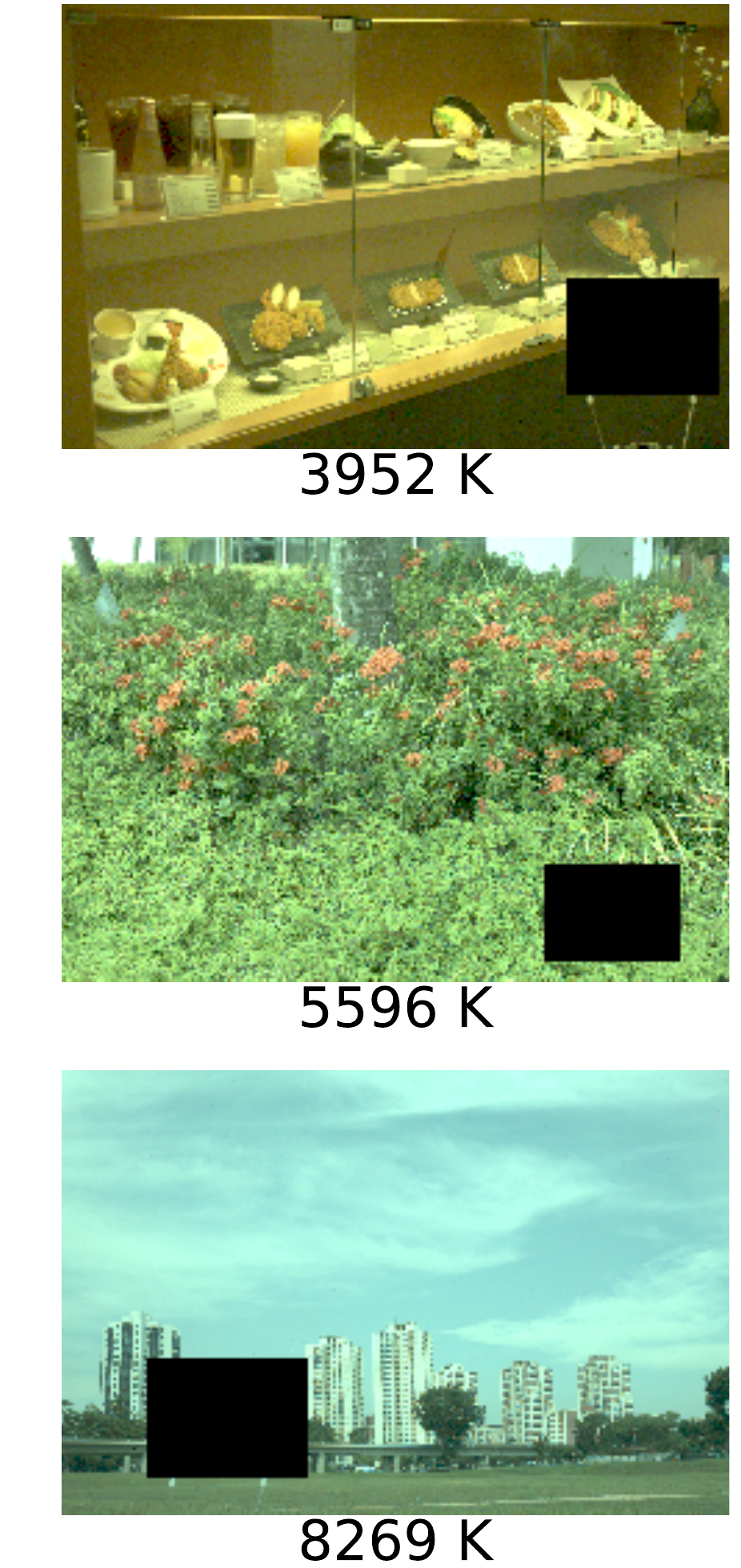}
    }
 \end{center}
\caption{ (a) Chromaticity space with Planckian locus. (b) Color temperature chart and types of light source associated with specific temperatures (c) Examples of images and corresponding color temperature $K$.}
\end{figure}

An overview of our proposed Meta-AWB method is shown in Fig. \ref{fig:sec:1:overview}. We consider a set of datasets $\Delta = \{\mathcal{D}_j\}$, where each dataset $\mathcal{D}_j = \{C_j, \{I_i\}_j\}$ comprises images acquired using a single camera $C_j$, representing various scenes under varying illumination conditions. Our objective is to provide a color constancy framework that is robust to variations in capture device, so as to leverage all the data available in $\Delta$ in order to adapt to previously unseen cameras with limited new training samples.
We propose a physically intuitive model that casts CC as a set of simple illuminant and camera specific regression tasks. We use the concept of color temperature to approximate the type of light source illuminating each image in our datasets, allowing us to separate images based on light source in an unsupervised way. As tasks are illuminant specific, estimated illuminant corrections have limited variability and can be evaluated accurately with limited training samples. This allows us to frame each task as a few-shot regression problem, which can be addressed using recent few-shot learning strategies, such as meta-learning. Section~\ref{sec3:CCT} reviews the concept of color temperature and its computation from images before introducing our few-shot task definition strategy in Section~\ref{sec3:illu_cam_tasks}. Section~\ref{sec3:meta} details how our novel formulation is integrated in a meta-learning framework.

\subsection{Correlated Color Temperature}
\label{sec3:CCT}

Color Temperature (CT) is a common measurement in photography, often used in high-end camera software to describe the color of the illuminant for setting white balance~\cite{jacobson2000manual}. By definition, CT measures, in degrees Kelvin, the temperature that is required to heat a Planckian (or black body) radiator to produce light of a particular color. A Planckian radiator is defined as a theoretical object that is a perfect radiator of visible light~\cite{schubert2018light}. The Planckian locus, illustrated in Fig.~\ref{fig:planck}, is the path that the color of a black body radiator would take in chromaticity space as the temperature increases, effectively illustrating all possible CTs.

In practice, the chromaticity of most light sources is off the Planckian locus, so the Correlated Color Temperature (CCT) is computed. CCT is the point on the Planckian locus closest to the non-Planckian light source \cite{schanda2007colorimetry,schubert2018light}. Intuitively, CCT describes the color of the light source and can be approximated from photos taken under this light. As shown in Fig.~\ref{fig:scale}, different temperatures can be associated with different types of light~\cite{jacobson2000manual}.
For each image, we can compute CCT using standard approaches that map CIE 1931 chromaticities $x$ and $y$ to CCT~\cite{Hernandez1999}. Chromaticities $x$, $y$ are coordinates in the chromaticity space which can easily be estimated from the image's RGB values \cite{schanda2007colorimetry}.



\begin{figure}
    \centering
    \subfigure[Pre-white balance images (Canon1D camera), marking their individual ground-truth gain corrections in $\lbrack\frac{r}{g},\frac{b}{g}\rbrack$ space. Image frame border color (red, blue) indicates corresponding image temperature histogram bin membership.]
    {
    \includegraphics[width=0.455\linewidth]{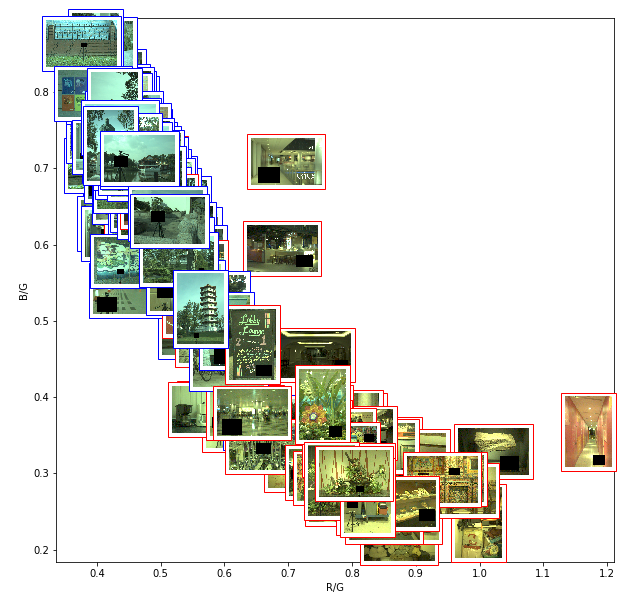}
    \label{fig:sec:3:dist_subfiga}
    }
    \hspace{2mm}%
    \subfigure[Ground-truth gain corrections for images observing identical scenes under similar illumination yet captured with distinct cameras (Canon1D, Sony; NUS-9~\cite{cheng2014illuminant}).]
    {
    \includegraphics[width=0.455\linewidth,trim={0.5cm 0.5cm 0.5cm 0.5cm},clip]{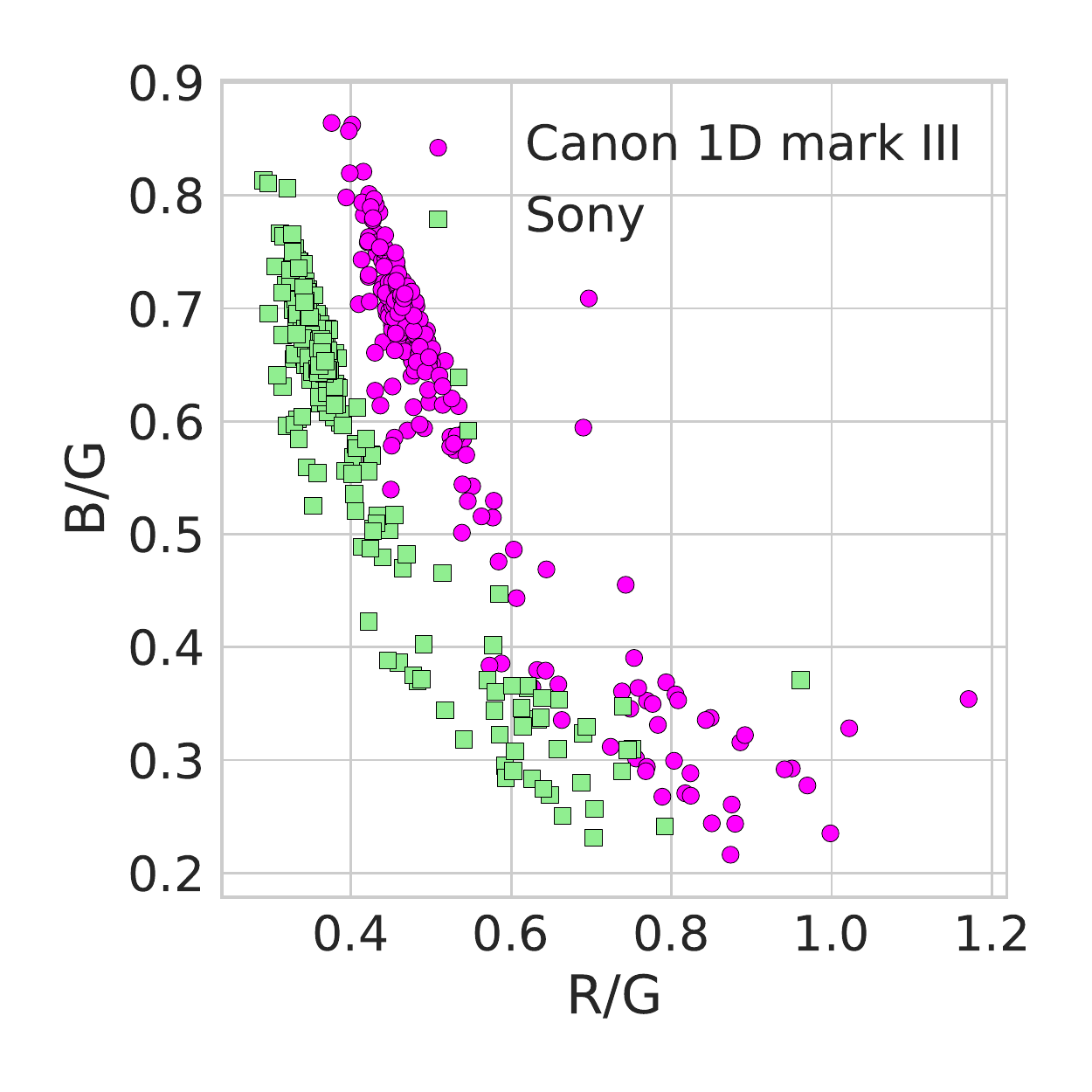}
    \label{fig:sec:3:dist_subfigb}
    }
    \caption{Our meta-task definition is conditioned on both image temperature and camera, motivated by the expected image light source separability and CSS distribution shifts.} 
    \label{fig:sec:3:dist_subfig}
\end{figure}

\subsection{Illuminant and camera-specific tasks}
\label{sec3:illu_cam_tasks}

Device-specific Camera Spectral Sensitivities (CSS) ($R_k(\lambda)$ in Eq. \ref{eq:illuminant_eq}) affect the color domain of captured images and the recording of scene illumination. Images captured by different cameras can therefore exhibit ground-truth illuminant distributions that occupy differing regions of the chromaticity space~\cite{gao2017improving}, as can be observed in Fig.~\ref{fig:sec:3:dist_subfigb}. Intuitively, this means that two images of the same scene and illuminant will have different illuminant corrections if taken by different cameras. In this context, a standard approach is to treat each camera dataset as an independent regression task. 
However, we expect to observe large variability in illuminant correction within one camera dataset, due to both scene and light source diversity. Achieving good performance and efficient generalisation to unseen cameras using tasks that contain too wide \emph{intra-task} diversity may be difficult in a setting where camera specific data is sparse. Gamut based color constancy methods~\cite{forsyth1990novel,finlayson1996color,barnard2000improvements} assume that the color of the illuminant is constrained by the colors observed in the image. We make a similar hypothesis and aim to regroup images with similar dominant colors in the same task.

As a result, we decompose the inter-camera color constancy problem in a set of regression problems that comprise images acquired \emph{with the same camera and with similar CCT} (i.e. similar dominant color). Our intuition is that, for each of these problems, illuminant corrections are clustered such that good performance can be obtained quickly with only a handful of training samples. 

We propose two strategies to separate images based on color temperature values. 
Our first approach is to compute a histogram $H_s$ for camera $s$ containing $M$ bins of CCT values and define each task as containing the set of images in each histogram bin. As a result, we define a task $T(\mathcal{D}_s, m) \in \mathcal{T}$ as:  
    $T(\mathcal{D}_s, m) = \{\ I\ \vert\  a^m_{s} \leq CCT(I) \leq b^m_{s},\ Cam(I) = C_s\}$  
where $Cam(I)$ is the camera used to acquire image $I$, and $a^m_{s}$, $b^m_{s}$ are the edges of bin $m$ in histogram $H_s$.   
Intuitively, images within the same temperature bin will have a similar dominant color, and therefore one could expect them to have similar illuminant corrections. Figure~\ref{fig:scale} highlights that a large variety of light sources are defined by relatively low temperatures. Accounting for this non-uniform distribution, we define bin edges of $H_s$ as a partition of temperature values on a logarithmic scale. In particular, when setting $M=2$, we expect to separate images under a \emph{warm} light source from images under a \emph{cold} light source (\emph{eg.} indoor images vs. outdoor images). This effect is illustrated in Fig.~\ref{fig:sec:3:dist_subfiga} where images are plotted marking their respective ground-truth gain correction in $\lbrack\frac{r}{g},\frac{b}{g}\rbrack$ space and image frame border colors indicate temperature bin membership. This low granularity decomposition yields a few-shot scenario such that only $10$ to $20$ training images will be required to adapt to a previously unseen camera.
A second, more granular approach consists of sampling K-nearest neighbour images in terms of temperatures, where K is the number of images comprising the regression task. Such a setting allows to separate the types of illuminants more precisely, but conversely requires more illuminant specific training images at test time. 

\subsection{Meta-learning formulation}
\label{sec3:meta}

Using the task formulation of Section~\ref{sec3:illu_cam_tasks}, we can frame camera-adaptive illuminant estimation as a few-shot learning problem where tasks are used to define learning episodes. One way of approaching this type of problem is to use meta-learning techniques such as the popular MAML algorithm \cite{finn2017model}, where the strategy is to learn an optimal neural network \emph{initialisation} capable of achieving strong performance on a new unseen task in only a few gradient updates, using only a small number of training samples.




Each regression task instance $\mathcal{T}:\boldsymbol{\hat{\rho}}_{\theta} = f_{\theta}(I)$ aims to estimate a global illuminant correction vector $\boldsymbol{\rho} = [r, g, b]$ for an input image $I$, where $f_{\theta}$ is a nonlinear function described by a neural network model. The model's parameters $\theta$ are learned by minimising the angular error loss:
\begin{equation}
\label{eq:loss}
    \mathcal{L}_{\mathcal{T}}(\hat{\theta}) = \arccos(\frac{\boldsymbol{\hat{\rho}}_{\theta}}{\parallel \boldsymbol{\hat{\rho}}_{\theta} \parallel} \cdot \frac{\boldsymbol{\rho}}{\parallel \boldsymbol{\rho} \parallel}).
\end{equation}
Angular error provides a standard metric sensitive to the inferred orientation of the $\boldsymbol{\hat{\rho}}_{\theta}$ vector, with respect to the ground-truth $\boldsymbol{\rho}$ yet agnostic to its magnitude, providing independence to the brightness of the illuminant.

MAML is an iterative algorithm that learns a global set of parameters $\theta^*$ across tasks by optimising fine-tuning performance on each training task. Each iteration comprises an \emph{inner update} which consists of fine-tuning global parameters $\theta$ to be task specific on a set of training images via $n$ gradient descent steps with learning rate $\alpha$. The second step, the \emph{outer update}, updates $\theta$ as:        
\begin{equation}
\label{eq:outer}
    \theta^* = \theta - \beta \nabla_{\theta} \sum_i \mathcal{L}_{\mathcal{T}_i}(f_{\theta_i}),
\end{equation}
where $\beta$ is the meta-learning rate parameter and $\mathcal{L}_{\mathcal{T}_i}(f_{\theta})$ is the regression loss function as described in Eq.~\ref{eq:loss}, computed using task specific fine-tuned parameters on a new set of (previously unseen) meta-test images.
At test time, parameters are fine-tuned for a new unseen task for $n$ gradient updates and $K$ training samples. We finally compute the illuminant correction for each test image $I$ as $\boldsymbol{\rho}_{\theta_i} = f_{\theta_i}(I)$. 



\section{Results}
\label{sec:results}


\noindent\textbf{Datasets and preprocessing.}
\label{sec3:data}
Three public color constancy datasets: Gehler-Shi~\cite{shi2000re,gehler2008bayesian}, NUS-9~\cite{cheng2014illuminant}, and Cube~\cite{banic2018unsupervised} 
are combined to investigate the capabilities of the proposed methodology. We use a total of $4128$ images captured by $12$ different cameras. For each dataset, we make use of the provided `almost-raw' PNG images for all experimental work that follows in Section \ref{sec:results}. Ground-truth illumination is measured by Macbeth Color Checker (MCC) in each dataset except for the `Cube' database that alternatively uses a SpyderCube~\cite{spydercube2018} calibration object. Illuminant ground-truth information (respective calibration objects) are masked in all images (using provided MCC coordinates and Cube+ using the fixed SpyderCube image location with mask value $\text{RGB}=[0,0,0]$), during both learning and inference. 
The \textbf{Gehler-Shi} dataset~\cite{shi2000re,gehler2008bayesian} contains $568$ images of indoor and outdoor scenes. Images were captured using Canon 1D and Canon 5D cameras.
The \textbf{NUS-9}-Camera dataset~\cite{cheng2014illuminant} consists of 9 subsets of $\sim$210 images per camera providing a total of $(1736+117$\footnote{The NUS dataset has recently been updated to include 117 additional images from a ninth camera. During training we use all nine cameras.}$)=1853$ images. All subsets comprise images representing the same scene, highlighting the influence of the camera sensor. 
The \textbf{Cube} dataset~\cite{banic2018unsupervised} contains $1365$ images and consists of predominantly outdoor
imagery (Canon EOS 550D camera). It was recently updated to include an additional $342$ indoor images (renamed \textbf{Cube+} dataset). We use all Cube+ data at train time. At inference time, we limit evaluation to the Cube dataset, in order to be directly comparable to previous work.




Camera specific black-level corrections are applied in keeping with offsets specified in the dataset descriptions. We apply a standard gamma correction ($\gamma = 2.2$) and normalize network input to $[0,1]$. Input images are converted to 8-bit where required, providing bit depth consistency across all datasets. Impoverished input (low bit-depth) has been shown to make the color constancy problem more difficult~\cite{barron2017fast} yet also provides a good real-world test bed as specialized camera hardware typically performs illuminant estimation using small, low bit-depth images.


\noindent\textbf{Implementation.}
For each camera, we train a model using random image crops of variable size ($128\times128$ to original image size) from the remaining $11$ cameras, spatially resized to $128\times128$. Due to the discussed limits on typical dataset size per camera and assumed simplicity of our regression tasks, we adopt a simple architecture comprising four convolutional layers (all sized $3\times3\times64$), an average pooling layer and two fully connected layers (sizes $64\times64$, $64\times3$), with ReLU activations on all hidden layers. Due to the makeup of our proof-of-concept amalgamated dataset, the majority of cameras capture similar scene content (NUS-9 imagery). Since we aim to optimise generalisation between cameras without overfitting to particular scenes, we choose the Gehler-Shi (Canon 5D) images as a validation set. This allows for optimisation of model hyperparameters using imagery containing unique scene content. 

We evaluate three variants of MAML, characterised by different definitions of the learning rate $\alpha$. \emph{MAML}~\cite{finn2017model}, uses a constant $\alpha$; \emph{metaSGD}~\cite{li2017meta} learns an $\alpha$ value per parameter in the network, allowing the direction and magnitude of the gradient descent to be learned; \emph{LSLR}~\cite{antoniou2018train} substantially reduces the number of trainable parameters, learning a $\alpha$ single parameter per \emph{layer} in the network for each inner gradient update.
We train our CNN model for $25k$ iterations, using a meta-batch size of $10$, number of training images per batch $K_{train}=10$, learning rate $\beta=0.001$ (with exponential decay). The inner-update learning rate $\alpha$ is set (or initialised for \emph{metaSGD} and \emph{LSLR}) to $0.001$. We use layer normalisation on all convolutional layers. 
At inference time, for each test image, we randomly select $K_{test}$ training samples (from the test image task) and fine-tune the model for $10$ iterations. To evaluate the statistical robustness of our method to variation in the selection of the $K_{test}$ images, we independently repeat $10$ draws for each test image. We report the median angular error over all images, averaged over all draws.
As a baseline, we train a model with standard back-prop and leave-one-out cross validation on camera using network architecture matching our introduced base-learner.
At test time we report both with (\emph{Baseline - fine tune}) and without (\emph{Baseline - no fine tune}) $K$-shot fine tuning. Baselines are trained for $25k$ iterations using the same parameters as our Meta-AWB model.

\begin{figure}{
    \centering
    \subfigure[Cube]{
    \label{fig:results:taskdef:subfiga}
    \includegraphics[width=0.2\linewidth]{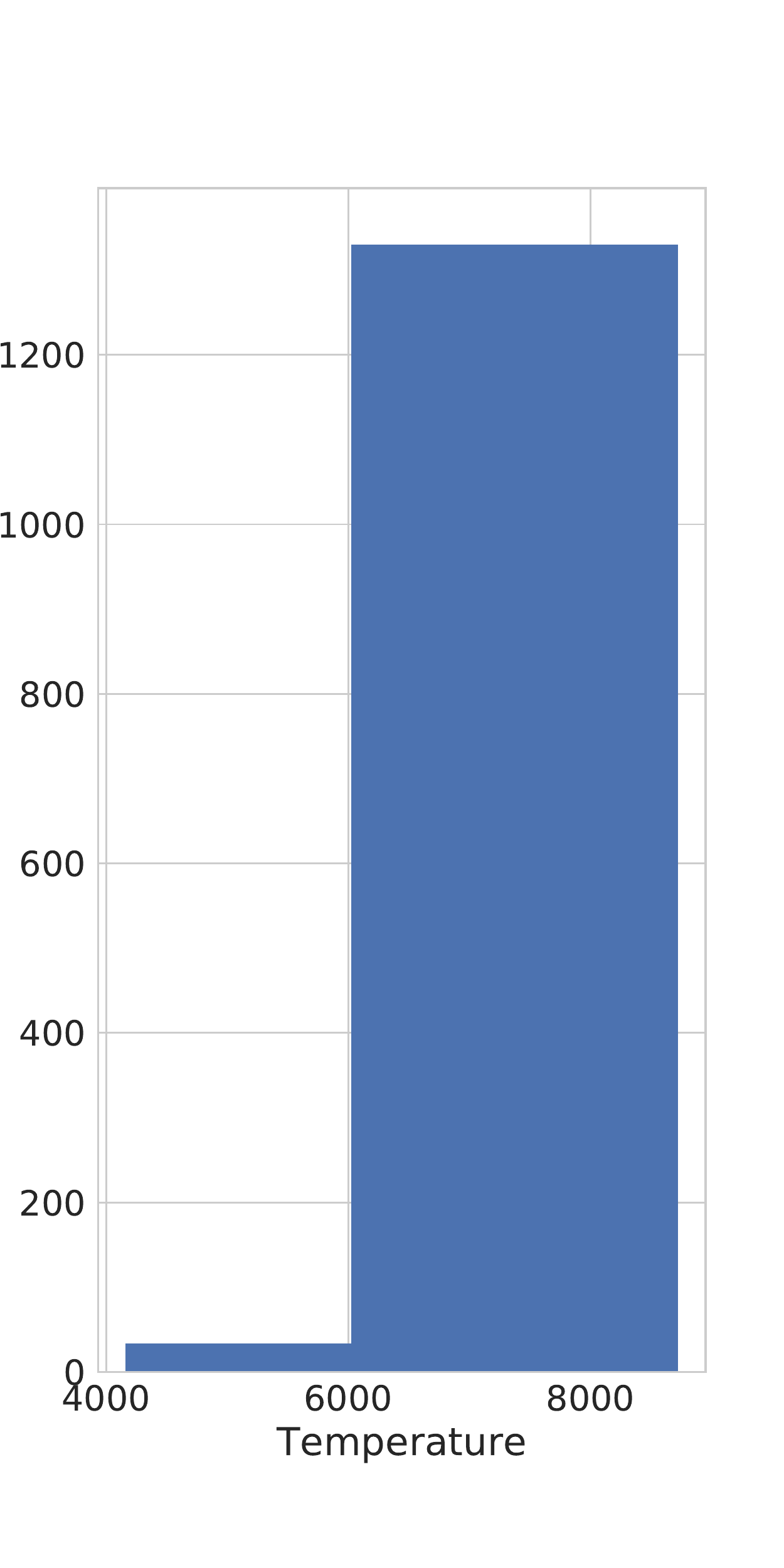}}~
    \subfigure[Canon600D (NUS)]{
    \label{fig:results:taskdef:subfigb}
    \includegraphics[width=0.2\linewidth,trim={0.2cm 0 0cm 0},clip]{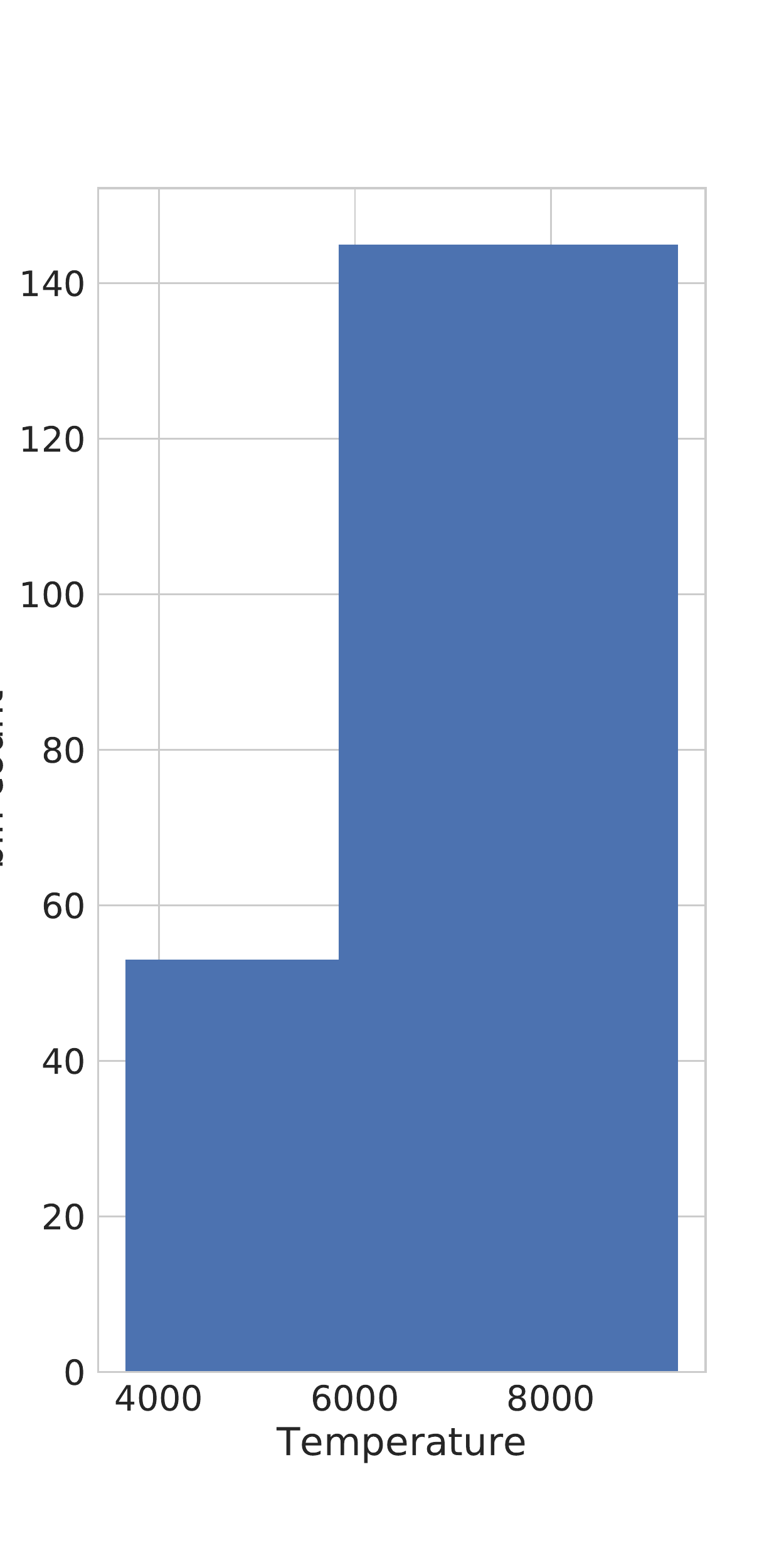}
}~\subfigure[Canon 600D (NUS) ground-truth illuminants in RGB space]{
        \label{fig:results:taskdef:subfigc}
        \includegraphics[width=0.52\linewidth,trim={0.8cm 0 1.8cm 0},clip]{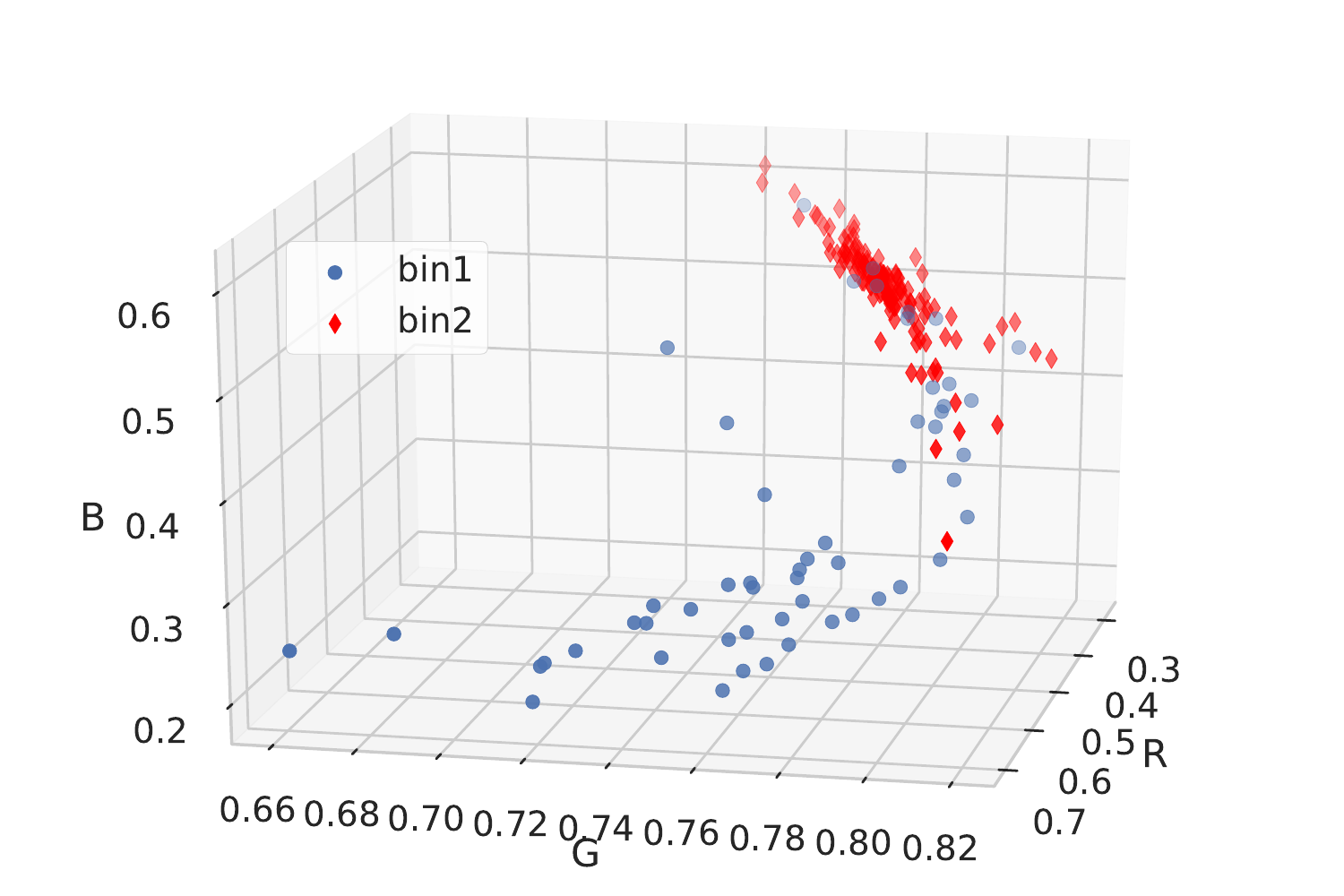}
    }
    \caption{Task definition: Temperature histograms and corresponding separation in RGB space.}
    \label{fig:results:taskdef}
}
\end{figure}

\textbf{Histogram-based task definition.}
To explore the validity of our learning task formulation, we plot CCT histograms per camera and ground-truth $[r,g,b]$ gain corrections per image in RGB space, with CCT bin assignment indicated. Figure~\ref{fig:results:taskdef:subfigb} and~\ref{fig:results:taskdef:subfigc} provide examples of these respectively for the NUS Canon 600D image set ($M{}={}2$ bins). Correlating bin edges (as defined in Section~\ref{sec3:illu_cam_tasks}) to the temperature chart found in Figure~\ref{fig:scale}, we see that images can be separated based on light type, and more specifically, indoor vs outdoor light sources. This observation is confirmed in Fig.~\ref{fig:results:taskdef:subfiga} where the CCT histogram, obtained using the original subset of Cube images (all images contain outdoor scenes), essentially contains all images in one bin. Figure~\ref{fig:results:taskdef} also illustrates that our CCT-histogram strategy generates homogeneous learning tasks since ground-truth illuminant corrections belonging to the same bin are well clustered in RGB space. 
This setup assigns images to a task, conditioned on both camera sensor and CCT bin, resulting in $M\cdot|\{ C_j \in \Delta\}|$ valid learning tasks assuming that each CCT bin is non-empty. In the remaining experiments, we set $M=2$ as discussed in Sec. \ref{sec3:illu_cam_tasks}. 

\begin{figure}
    \centering
    \subfigure{
    \addtocounter{subfigure}{3}
    \label{fig:finetune_vs_kshot:subfig}
    \includegraphics[width=0.9\linewidth]{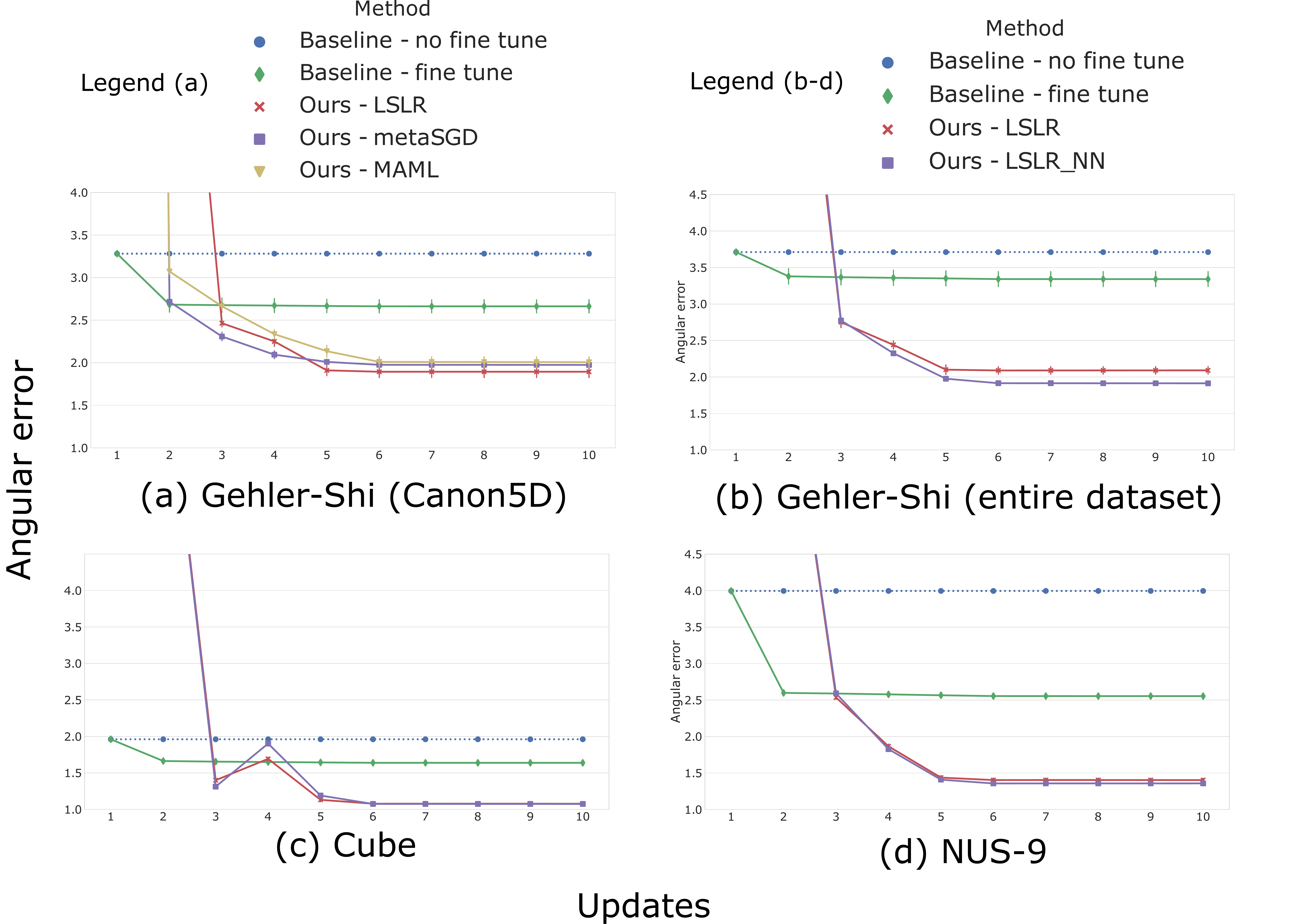}
    }
    
    
    \subfigure[Gehler-Shi (Canon 5D)]{
    \label{fig:finetune_vs_kshot:subfige}
    \includegraphics[width=0.46\linewidth]{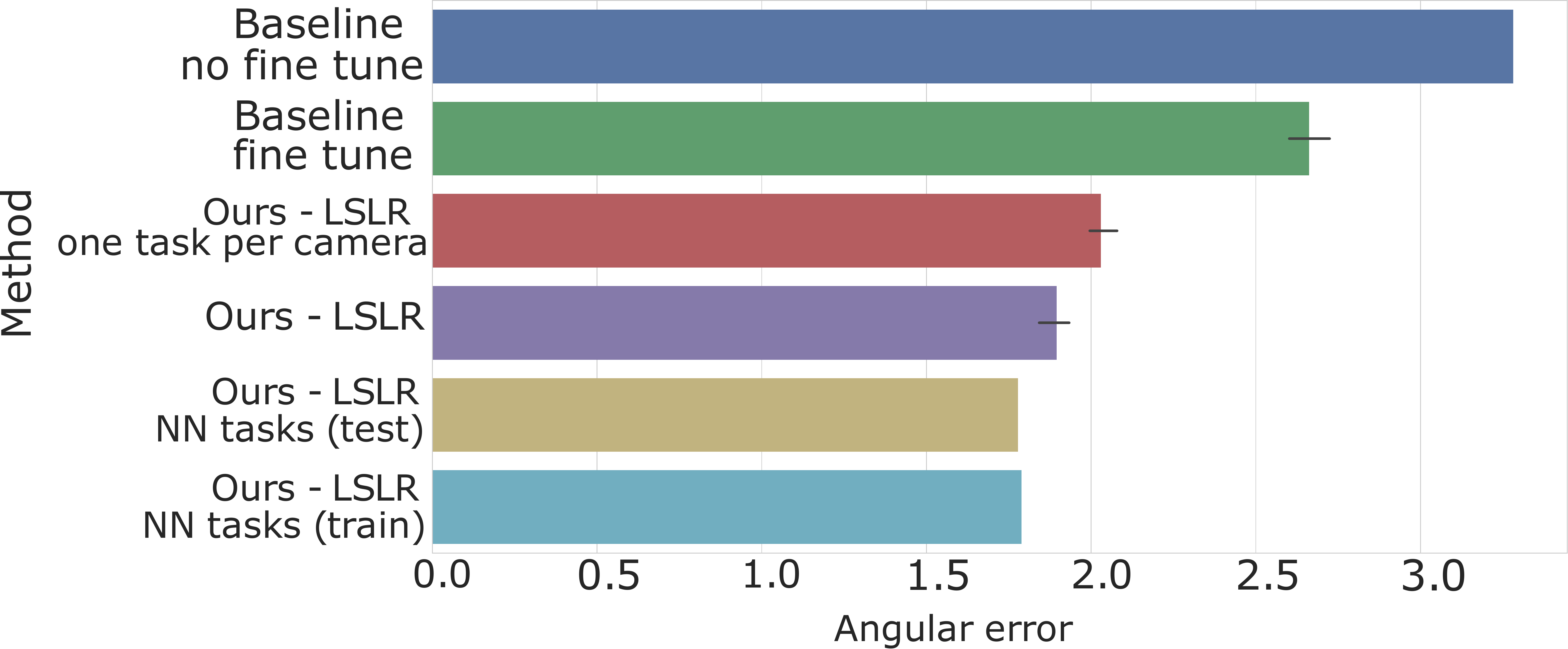}
    }
    \subfigure[NUS-9, one random draw]{
    \label{fig:finetune_vs_kshot:subfigf}
    \includegraphics[width=0.43\linewidth]{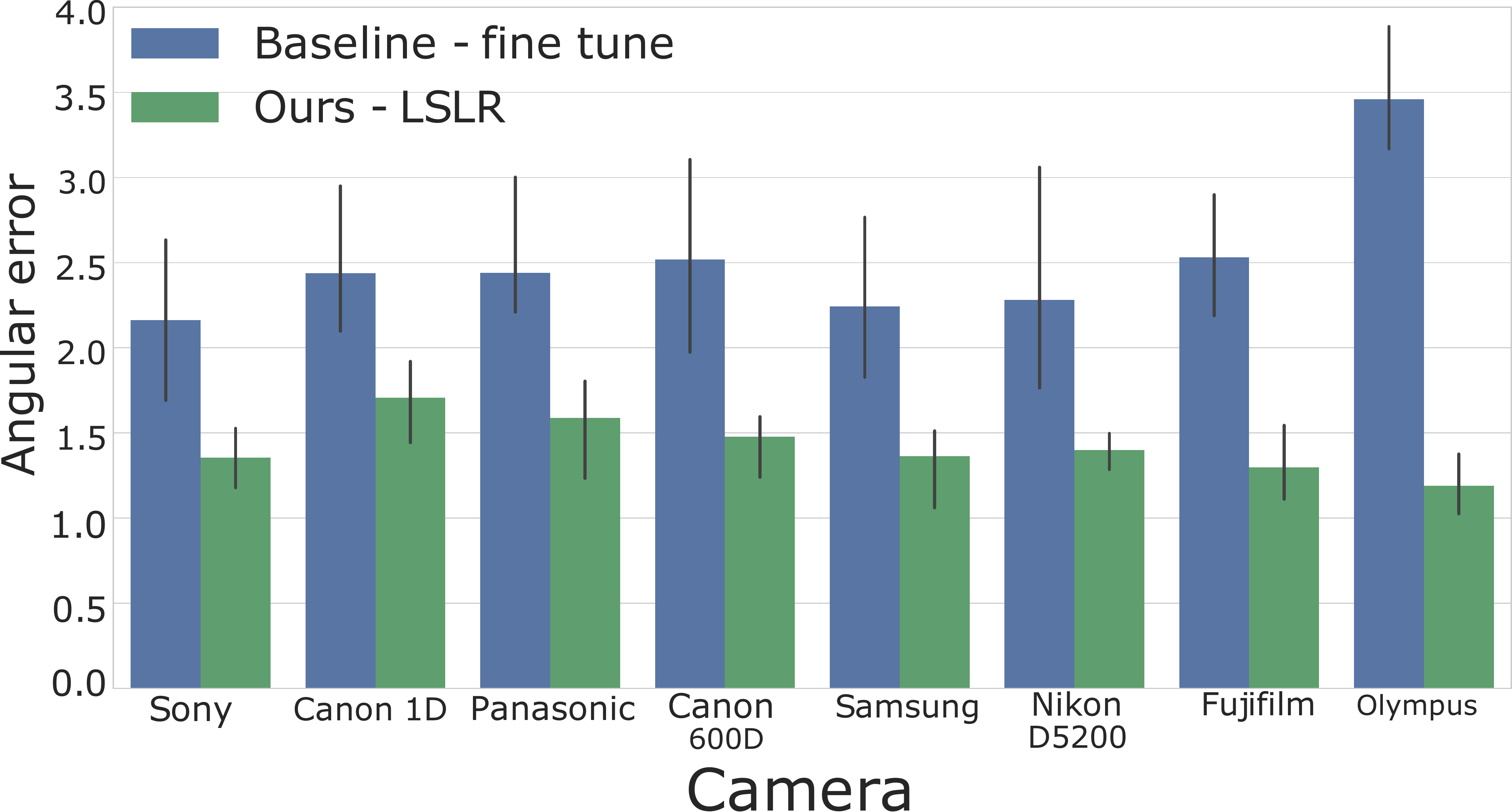}
    }
    \caption{Median angular-error with respect to the number of fine-tuning updates. (a,e) Parameter study results, (b-d) Dataset specific results compared to the baselines, (f) Per camera angular-error after 10 updates, all NUS-9 cameras. Error bars in (a-e) report inter-draws variance.}
    \label{fig:finetune_vs_kshot}
\end{figure}

\noindent\textbf{Parameter and method analysis.}
Using our validation camera we evaluate the influence of key parameters and meta-learning strategy. 
For each method considered (\emph{MAML}, \emph{metaSGD} and \emph{LSLR}),
we train models for variable numbers of inner gradient updates $n_{train} \in \{1,5,10\}$. While $K_{train}=10$ images is fixed during meta-model training, we evaluate the influence of available $k$-shot image count at test time, computing performance for $K_{test} \in \{5, 10, 20\}$. We also report results for inner updates $n_{test} \in \{1, 5, 10\}$. Since \emph{LSLR} learns a different learning rate per inner update, we set $\alpha_i = \alpha_n$, $\forall i \geq n$ when $n_{test}$ is set to a value larger than that used during training.

We report results in Fig. \ref{fig:finetune_vs_kshot}(a) and Table \ref{tab:param}, with the best results for each $K_{test}$ reported in bold. We observe a substantial improvement in performance when increasing the number of inner updates from $1$ to $5$, but note that $10$ updates do not improve performance.  
\emph{LSLR} appears to offers a compromise between simplicity and flexibility, yielding the best results when $n_{train}=5,10$. However, interestingly, \emph{LSLR} and \emph{metaSGD} perform poorly compared to \emph{MAML} when training with only $1$ gradient update. 
Predictably, we observe an increase in performance as we increase $K_{test}$ reaching our overall best performance with a median angular error of $1.81$ degrees and $K_{test}=20$ samples. Finally, we can see that all methods benefit from $n_{test}>1$ updates, but tend to plateau when $n_{test}>5$. 
Considering our experimental observations, we use $n_{train}=5$ and $n_{test}=10$ with our \emph{LSLR} variant for the remainder of our experimental work. We set $K_{test}=10$ unless otherwise specified. 

\noindent\textbf{Influence of task definition strategy.}
Using our validation camera, we further evaluate the influence of different task definition approaches. As shown in Fig. \ref{fig:finetune_vs_kshot:subfige}, we compare our $M = 2$ bins histogram based approach (\emph{Ours - LSLR}) to 1) $M = 1$, corresponding to the naive approach of setting one camera dataset $D_s$ as a task (\emph{Ours - LSLR - one task per camera}), 2) defining tasks as temperature nearest neighbours at test time (\emph{Ours - LSLR - NN tasks (test)}) and both and train and test time (\emph{Ours - LSLR - NN tasks (train)}). Results are compared to the baseline method for context. We observe that results improve as task definition methods get more granular (i.e. provide better illuminant separation), while our two experiments on nearest neighbour tasks show that testing on more granular tasks (with respect to train tasks) can equally improve performance and allows to make use of potential larger datasets at test time.

\begin{table}[t]
\begin{center}
\centering
\renewcommand\arraystretch{1.2}
\setlength{\tabcolsep}{3pt}
\scriptsize
\begin{tabular}{ | l  l | c  c  c  c  c  c  c  c  c|}
   \cline{3-11}
   \multicolumn{2}{c}{}& \multicolumn{9}{|c|}{Training updates $u=n_{train}$}  \\ 
   \multicolumn{2}{c|}{} 
   & \multicolumn{3}{c}{ MAML} 
   & \multicolumn{3}{c}{ metaSGD} 
   & \multicolumn{3}{c|}{ LSLR}\\ 
   \multicolumn{2}{c|}{} & \hbox{\strut\tiny{$u{=}1$}} & \hbox{\strut\tiny{$u{=}5$}} & \hbox{\strut\tiny{$u{=}10$}} & \hbox{\strut\tiny{$u{=}1$}} & \hbox{\strut\tiny{$u{=}5$}} & \hbox{\strut\tiny{$u{=}10$}} & \hbox{\strut\tiny{$u{=}1$}} & \hbox{\strut\tiny{$u{=}5$}} & \hbox{\strut\tiny{$u{=}10$}}\\
   \hline
   \multirow{9}{*}{\rotatebox{90}{\small Testing parameters\,}} 
   & \tiny{\textbf{$K_{test}$ = 5}}   &&&&&&  &&& \\ 
   & \quad 1 update   & $2.18$ & $3.08$ & $3.11$ & $2.12$ & $2.80$ & $3.10$ & $2.42$ & $9.07$ & $4.68$ \\ 
   & \quad 5 updates  & $2.06$ & $2.07$ & $2.08$ & $2.27$ & $2.11$ & $2.07$ & $2.31$ & $\mathbf{2.00}$ & $2.76$ \\ 
   & \quad 10 updates & $2.06$ & $2.07$ & $2.08$ & $2.28$ & $2.11$ & $2.06$ & $2.31$ & $\mathbf{2.00}$ & $2.05$ \\ 
   & \tiny{\textbf{$K_{test}$ = 10}}  &&&&&& &&&   \\ 
   & \quad 1 update   & $2.13$ & $3.05$ & $3.12$ & $2.00$ & $2.71$ & $3.09$ & $2.37$ & $9.05$ & $4.52$ \\ 
   & \quad 5 updates  & $2.02$ & $2.01$ & $2.00$ & $2.09$ & $1.98$ & $1.94$ & $2.23$ & $\mathbf{1.87}$ & $2.50$ \\ 
   & \quad 10 updates & $2.02$  & $2.00$ & $2.00$ & $2.09$ & $1.97$ & $1.93$ & $2.23$ & $\mathbf{1.87}$ & $1.92$ \\ 
   & \tiny{\textbf{$K_{test}$ = 20}}  &&&&&& &&&  \\ 
   & \quad 1 update   & $2.10$ & $3.05$ & $3.09$ & $1.99$ & $2.70$ & $3.05$ & $2.31$ & $9.01$ & $4.40$ \\ 
   & \quad 5 updates  & $1.98$ & $1.94$ & $2.00$ & $2.06$ & $1.89$ & $1.91$ & $2.18$ & $\mathbf{1.81}$ & $2.41$ \\ 
   & \quad 10 updates & $1.98$ & $1.94$ & $1.95$ & $2.06$ & $1.89$ & $1.91$ & $2.18$ & $\mathbf{1.81}$ & $1.84$ \\ \hline
\end{tabular}
\captionof{table}{Our meta-learning hyper-parameter investigation and method analysis. Median angular-error. Best results for each $K$-shot configuration are reported in bold. } 
\label{tab:param}
\end{center}
\end{table}

\begin{figure}[t] 
\centering
    \subfigure[\small{input image}]{
    \label{fig:results:example:subfig1}
    \includegraphics[width=0.42\linewidth,trim={0cm 12cm 0cm 0cm},clip]{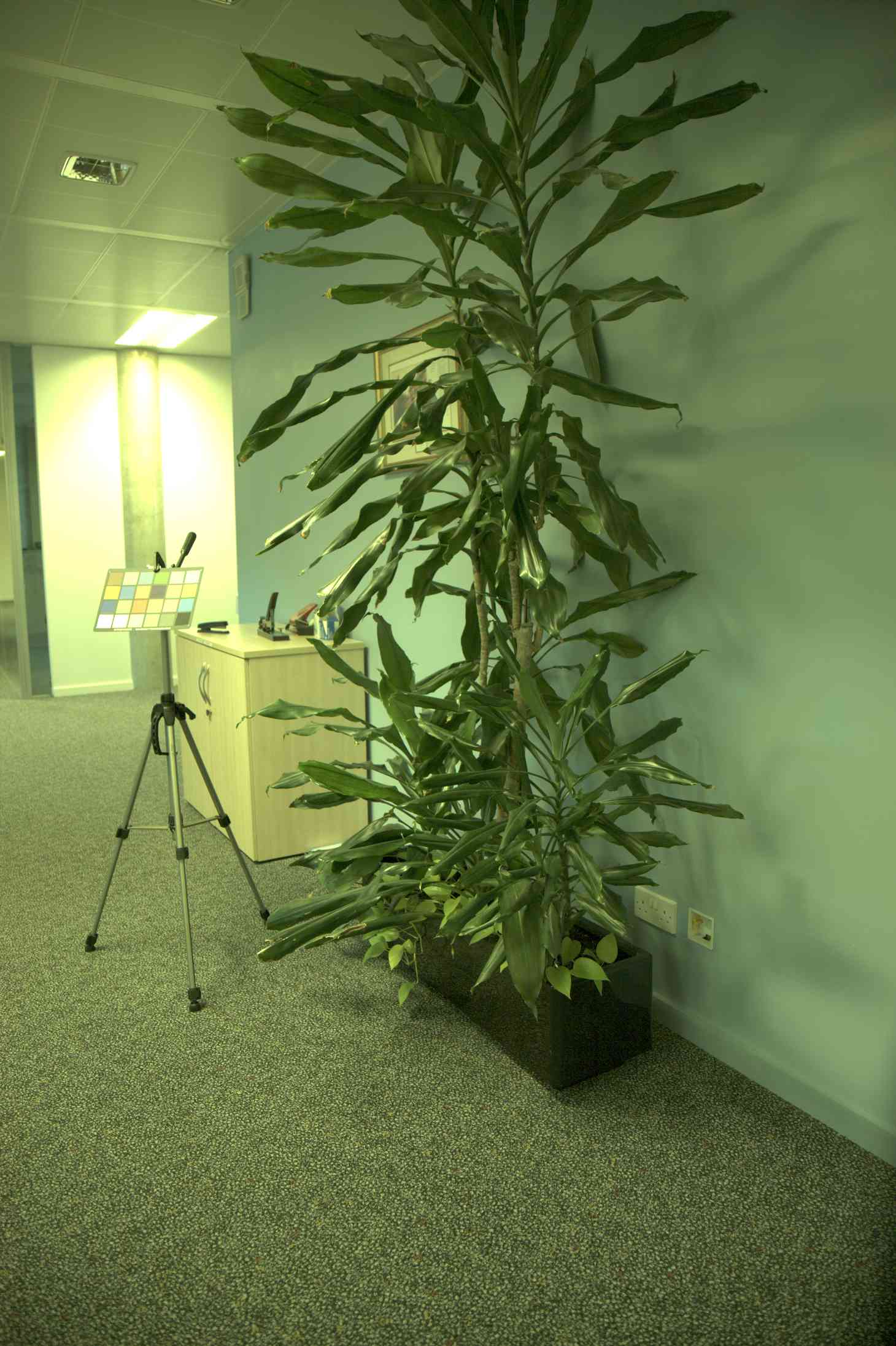}}
    \subfigure[\small{Ground-truth solution}]{
    \label{fig:results:example:subfig2}
    \includegraphics[width=0.42\linewidth,trim={0cm 12cm 0cm 0cm},clip]{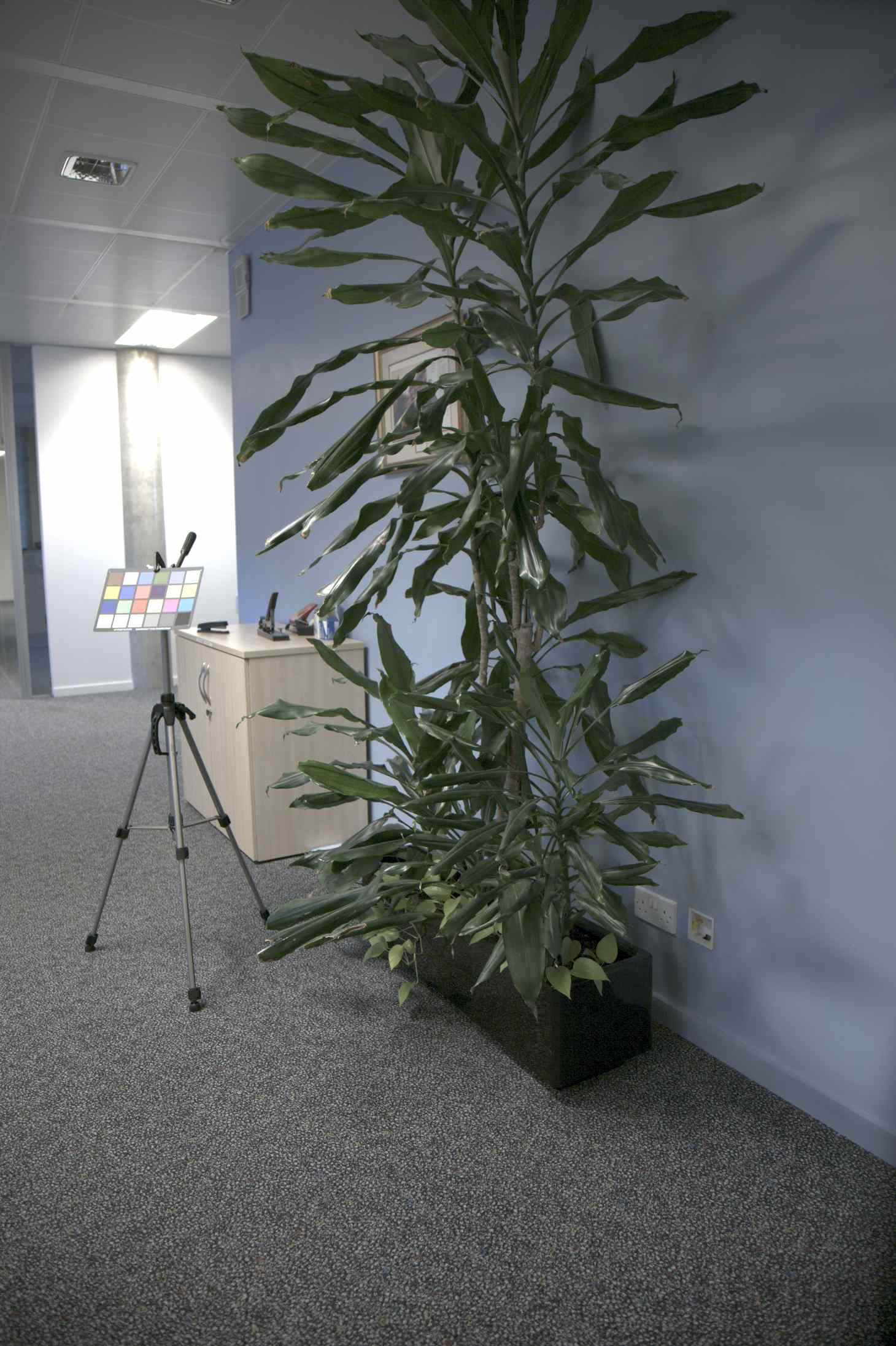}}\\
    \subfigure[\small{Meta-AWB, angular error: $5.42\degree$}]{
    \label{fig:results:example:subfig3}
    \includegraphics[width=0.42\linewidth,trim={0cm 12cm 0cm 0cm},clip]{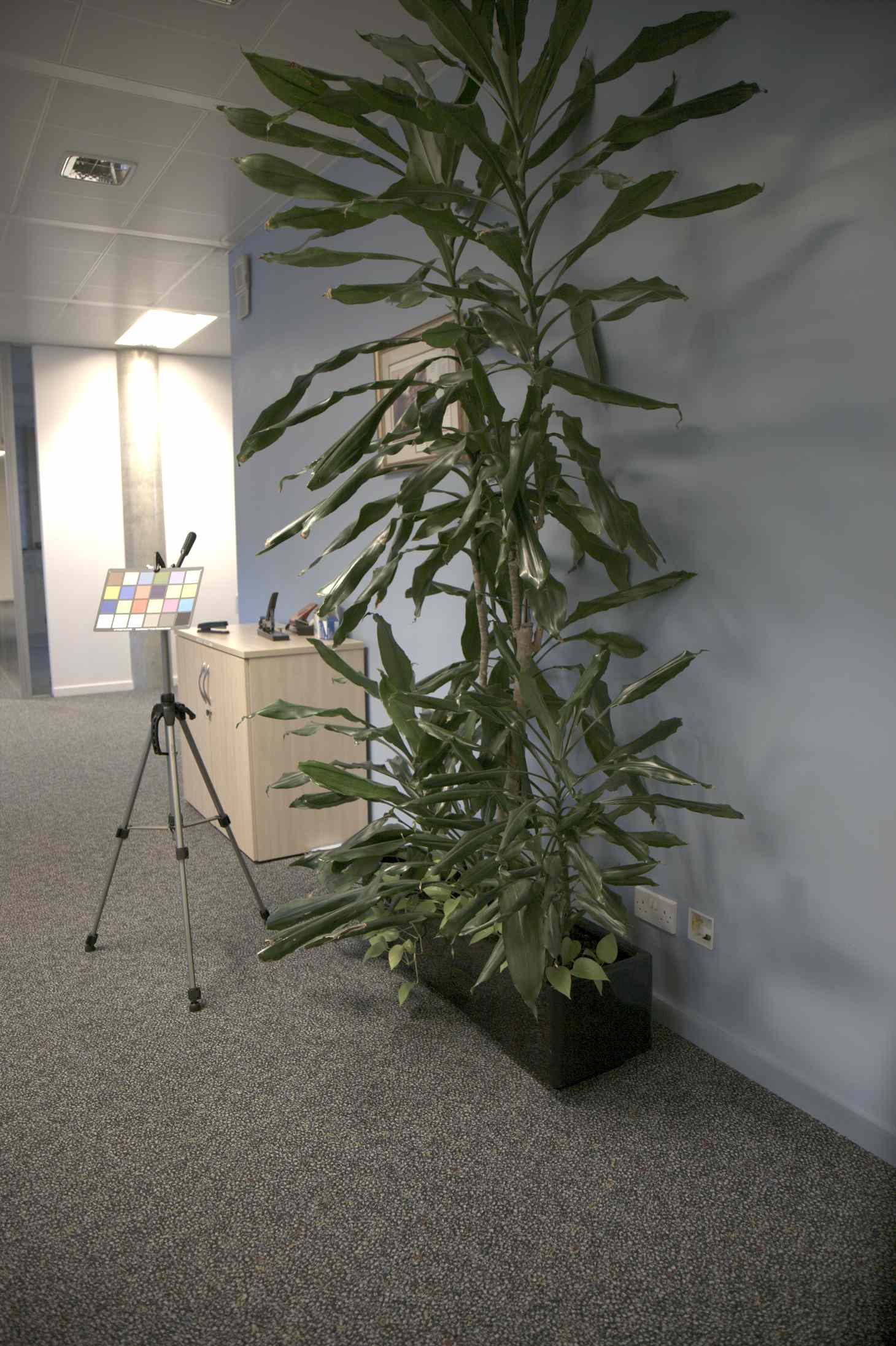}}
    \subfigure[\small{Baseline fine-tuned, angular error: $18.76\degree$}]{
    \label{fig:results:example:subfig4}
    \includegraphics[width=0.42\linewidth,trim={0cm 12cm 0cm 0cm},clip]{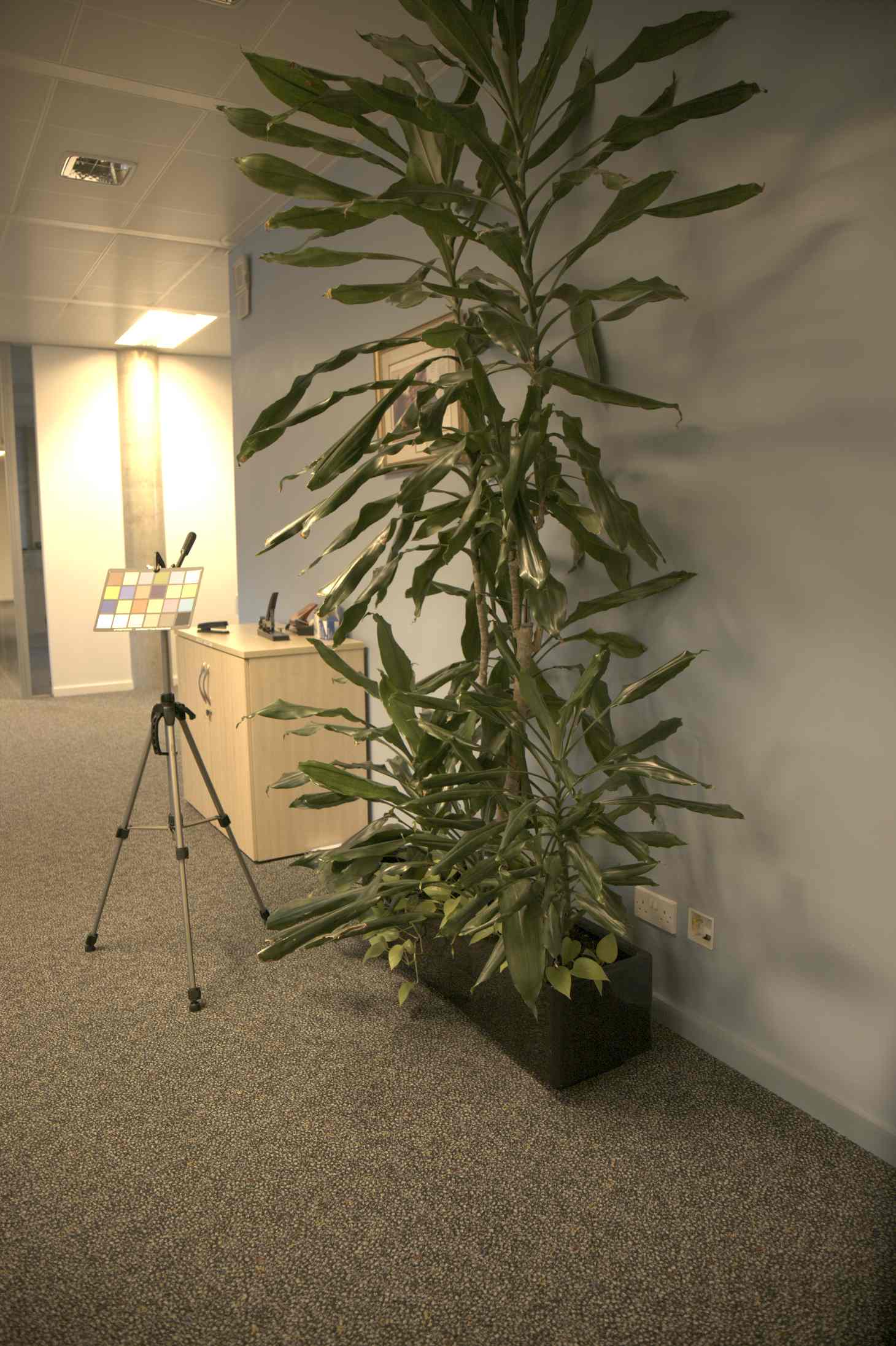}}
\caption{A challenging  Gehler-Shi~\cite{gehler2008bayesian,shi2000re} test image from our worst 25$\%$. 
Images are shown in sRGB space and clipped at the $97.5$ percentile.
See supplementary material for additional qualitative results.
}
\label{fig:results:example}
\end{figure}

\textbf{Comparisons to the state of the art}
Figure~\ref{fig:finetune_vs_kshot} shows the evolution of the median angular error, with respect to the number of gradient updates, for all datasets under both baselines and our approach (with and without nearest neighbour tasks at test time). 
We observe 
a significant gap in performance for all datasets. 
Fig.~\ref{fig:finetune_vs_kshot:subfigf} shows per camera performance and highlights that, unlike the baseline which particularly struggles for some cameras (\emph{eg.} Olympus), 
our method provides a consistent and better performance on each NUS-9 test camera individually, in addition to average performance. We provide a visual example in Fig.~\ref{fig:results:example}.

Finally, we compare our results on all datasets with recent state of the art approaches.
We report results on NUS-8 (without the recently added Nikon D40 camera) to provide a fair and accurate comparison. Quantitative results are shown in Tables \ref{tab:final_meta_results_nus} (NUS-8), \ref{tab:final_meta_results_shigehler} (Gehler-Shi) and \ref{tab:final_meta_results_cube} (Cube) where we report standard angular-error statistics (Tri.\ is trimean and G.M.\ geometric mean). We obtain results that are competitive with the state of the art and fully supervised methods, despite using an order of magnitude less camera specific training data. We achieve good generalisation on all datasets, in particular with the NUS-8 and Cube datasets, where we outperform most state of the art methods. 
The superior performance on NUS-8 can be linked to the fact that the NUS scene content is repeatedly seen during training.

\begin{table}[t]
\small
\centering
\scriptsize
\begin{tabular}{ p{28mm}|p{5mm} p{5mm} p{5mm} p{9.5mm} p{10.825mm} p{5mm}  }
\multicolumn{7}{c}{} \\
Algorithm & Mean & Median & Tri. & {Best 25{\%}} & {Worst 25{\%}} & G.M. \\
\hline
\tiny{Low-level statistics-based methods}                  & & & & & & \\
\hline
\rowcolor{Gray}
White-Patch~\cite{brainard1986analysis}                      & ${9.91}$ & ${7.44}$ & ${8.78}$ & ${1.44}$ & ${21.27}$ & ${7.24}$ \\ 
\rowcolor{Gray}
Gray-world~\cite{buchsbaum1980spatial}                       & ${4.59}$ & ${3.46}$ & ${3.81}$ & ${1.16}$ & ${9.85}$  & ${3.70}$ \\ 
\rowcolor{Gray}
Edge-based Gamut~\cite{gijsenij2010generalized}              & ${4.40}$ & ${3.30}$ & ${3.45}$ & ${0.99}$ & ${9.83}$  & ${3.45}$ \\ 
\rowcolor{Gray}
Natural Image Statistics~\cite{gijsenij2011color}            & ${3.45}$ & ${2.88}$ & ${2.95}$ & ${0.83}$ & ${7.18}$  & ${2.81}$ \\ 
\hline
\tiny{Fully-supervised learning}                  & & & & & & \\
\hline
\rowcolor{Gray}
Bayesian~\cite{gehler2008bayesian}                           & ${3.50}$ & ${2.36}$ & ${2.57}$ & ${0.78}$ & ${8.02}$  & ${2.66}$ \\ 
\rowcolor{Gray}Cheng et al. 2014~\cite{cheng2014illuminant}  & ${2.93}$ & ${2.33}$ & ${2.42}$ & ${0.78}$ & ${6.13}$  & ${2.40}$ \\ 
\rowcolor{Gray} SqueezeNet-FC4~\cite{hu2017fc4}              & ${2.23}$ & ${1.57}$ & ${1.72}$ & ${0.47}$ & ${5.15}$  & ${1.71}$ \\
\rowcolor{Gray}CCC~\cite{barron2015convolutional}            & ${2.38}$ & ${1.48}$ & ${1.69}$ & ${0.45}$ & ${5.85}$  & ${1.74}$ \\ 
\rowcolor{Gray}Cheng et al. 2015~\cite{cheng2015effective}   & ${2.18}$ & ${1.48}$ & ${1.64}$ & ${0.46}$ & ${5.03}$  & ${1.65}$ \\ 
\rowcolor{Gray}Shi et al. 2016~\cite{shi2016deep}            & ${2.24}$ & ${1.46}$ & ${1.68}$ & ${0.48}$ & ${6.08}$  & ${1.74}$ \\
FFCC~\cite{barron2017fast}  \tiny{(thumb, 8bit input)}       & \cellcolor{Gray} ${2.06}$ & \cellcolor{Gray} ${1.39}$ & \cellcolor{Gray} ${1.53}$ & ${0.39}$ & \cellcolor{Gray} ${4.80}$ & \cellcolor{Gray} ${1.53}$ \\
FFCC~\cite{barron2017fast}                                   & \cellcolor{Gray} ${1.99}$ & ${1.31}$ & ${1.43}$ & ${0.35}$ & \cellcolor{Gray}${4.75}$  & ${1.44}$ \\
\hline
\tiny{\{Unsupervised, Few-shot\} learning}                  & & & & & & \\
\hline
\rowcolor{Gray}
Color Tiger~\cite{banic2018unsupervised}     & ${2.96}$ & ${1.70}$ & ${1.97}$ & ${0.53}$ & ${7.50}$  & ${2.09}$ \\ 
Meta-AWB $K=10$                              & ${1.93}$ & ${1.38}$ & ${1.49}$ & ${0.47}$ & ${4.37}$ & ${1.52}$  \\ 
Meta-AWB $K=20$                              & ${1.89}$ & ${1.34}$ & ${1.44}$ & ${0.45}$ & ${4.28}$ & ${1.47}$  \\ 
\hline
\end{tabular}
\caption{Performance on the NUS-8 dataset~\cite{cheng2014illuminant}. We follow the same format as~\cite{barron2017fast}, reporting average performance (geometric mean) over the $8$ original NUS cameras.  Results outperformed by ours are marked in gray. }
\label{tab:final_meta_results_nus}
\end{table}

\begin{table}[t]
\small
\centering
\scriptsize
\begin{tabular}{ p{25mm}|p{5mm} p{5mm} p{5mm} p{10mm} p{11mm} p{5mm}  }
\multicolumn{7}{c}{} \\
Algorithm & Mean & Median & Tri. & Best 25\% & Worst 25\% & G.M. \\
\hline
\tiny{Low-level statistics-based methods }                  & & & & & & \\
\hline
\rowcolor{Gray}Gray-world~\cite{buchsbaum1980spatial}         & ${6.36}$ & ${6.28}$ & ${6.28}$ & ${2.33}$ & ${10.58}$ & ${5.73}$ \\ 
\rowcolor{Gray}White-Patch~\cite{brainard1986analysis}        & ${7.55}$ & ${5.86}$ & ${6.35}$ & ${1.45}$ & ${16.12}$ & ${5.76}$ \\ 
\rowcolor{Gray}Edge-based Gamut~\cite{gijsenij2010generalized}& ${6.52}$ & ${5.04}$ & ${5.43}$ & ${1.90}$ & ${13.58}$ & ${5.40}$ \\ 
\hline
\tiny{Fully-supervised learning }                  & & & & & & \\
\hline
\rowcolor{Gray} Bayesian~\cite{gehler2008bayesian}           & ${4.82}$ & ${3.46}$ & ${3.88}$ & ${1.26}$ & ${10.49}$ & ${3.86}$ \\ 
\rowcolor{Gray} Cheng et al. 2014~\cite{cheng2014illuminant} & ${3.52}$ & ${2.14}$ & ${2.47}$ & ${0.50}$ & ${8.74}$  & ${2.41}$ \\ 
Bianco CNN~\cite{bianco2015color}                            & \cellcolor{Gray} ${2.63}$ & \cellcolor{Gray} ${1.98}$ &  \cellcolor{Gray} ${2.10}$ & \cellcolor{Gray} ${0.72}$ & ${3.90}$  & \cellcolor{Gray} ${2.04}$ \\ 
Cheng et al. 2015~\cite{cheng2015effective}                  & ${2.42}$ & ${1.65}$ & ${1.75}$ & ${0.38}$ & ${5.87}$  & ${1.73}$ \\ 
CCC~\cite{barron2015convolutional}                           & ${1.95}$ & ${1.22}$ & ${1.38}$ & ${0.35}$ & ${4.76}$  & ${1.40}$ \\ 
SqueezeNet-FC4~\cite{hu2017fc4}                              & ${1.65}$ & ${1.18}$ & ${1.27}$ & ${0.38}$ & ${3.78}$  & ${1.22}$ \\
DS-Net~\cite{shi2016deep}                                    & ${1.90}$ & ${1.12}$ & ${1.33}$ & ${0.31}$ & ${4.84}$  & ${1.34}$ \\
FFCC~\cite{barron2017fast} \tiny{(thumb, 8bit input)}        & ${2.01}$ & ${1.13}$ & ${1.38}$ & ${0.30}$ & ${5.14}$  & ${1.37}$ \\
FFCC~\cite{barron2017fast}                                   & ${1.61}$ & ${0.86}$ & ${1.02}$ & ${0.23}$ & ${4.27}$  & ${1.07}$ \\ 
\hline
\tiny{Few-shot learning }                  & & & & & & \\
\hline
Meta-AWB $K=10$                             & ${3.07}$ & ${2.08}$ & ${2.28}$ & ${0.56}$ & ${7.31}$ & ${2.26}$    \\ 
Meta-AWB $K=20$                             & ${2.99}$ & ${2.02}$ & ${2.18}$ & ${0.55}$ & ${7.19}$ & ${2.20}$    \\
Meta-AWB NN $K=10$                             & ${2.66}$ & ${1.91}$ & ${1.99}$ & ${0.49}$ & ${6.20}$ & ${1.98}$ \\
Meta-AWB NN $K=20$                             & ${2.57}$ & ${1.84}$ & ${1.94}$ & ${0.47}$ & ${6.11}$ & ${1.92}$ \\
\hline
\end{tabular}
\caption{Performance on the Gehler-Shi dataset~\cite{shi2000re,gehler2008bayesian}. Previous methods as reported by~\cite{barron2017fast}. Results outperformed by our best method are marked in gray.} 
\label{tab:final_meta_results_shigehler}
\end{table}

\begin{table}[t]
\small
\centering
\scriptsize
\begin{tabular}{ p{26mm}|p{5mm} p{5mm} p{5mm} p{10mm} p{11mm} p{5mm}  }
\multicolumn{7}{c}{} \\
Algorithm & Mean & Median & Tri. & Best 25\% & Worst 25\% & G.M. \\
\hline
\tiny{Low-level statistics-based methods}                 & & & & & & \\
\hline
\rowcolor{Gray}
White-Patch~\cite{brainard1986analysis}             & ${6.58}$ & ${4.48}$ & ${5.27}$ & ${1.18}$ & ${15.23}$ & ${4.88}$  \\ 
\rowcolor{Gray}
Gray-World~\cite{buchsbaum1980spatial}              & ${3.75}$ & ${2.91}$ & ${3.15}$ & ${0.69}$ & ${8.18}$ & ${2.87}$   \\ 
\rowcolor{Gray}
General Gray-World~\cite{barnard2002comparison}     & ${2.50}$ & ${1.61}$ & ${1.79}$ & ${0.37}$ & ${6.23}$ & ${1.76}$   \\ 
Smart Color Cat~\cite{banic2015using}               & ${1.49}$ & ${0.88}$ & ${1.06}$ & ${0.24}$ & ${3.75}$ & ${1.04}$   \\ 
\hline
\tiny{\{Unsupervised, Few-shot\} learning }                  & & & & & & \\
\hline
\rowcolor{Gray}
Color Tiger~\cite{banic2018unsupervised}            & ${2.94}$ & ${2.59}$ & ${2.66}$ & ${0.61}$ & ${5.88}$ & ${2.35}$   \\ 
Meta-AWB $K=10$                 & ${1.63}$ & ${1.08}$ & ${1.20}$ & ${0.31}$ & ${3.89}$ & ${1.17}$ \\ 
Meta-AWB $K=20$                 & ${1.59}$ & ${1.02}$ & ${1.15}$ & ${0.30}$ & ${3.85}$ & ${1.16}$ \\ 
\hline
\end{tabular}
\caption{Performance on the Cube dataset. Previous methods as reported by~\cite{banic2018unsupervised}. Results outperformed by ours are marked in gray.}{} 
\label{tab:final_meta_results_cube}
\end{table}

\section{Conclusion}
\label{sec:conclusion}


In this paper, we propose a novel formulation of the computational color constancy problem that adapts and generalises quickly to a large variety of camera sensors. We exploit the concept of color temperature to approximate the type of light source from images, so as to decompose the problem into a set of simpler regression tasks, each associated with a camera sensor and type of light source. The simplified nature of the obtained regression tasks allows us to cast color constancy as a few-shot learning problem that we address using meta-learning. Extensive experiments across three benchmark datasets and $12$ different camera sensors result in performance competitive with the 
fully-supervised state-of-the-art, using only a small fraction of camera specific data at test time. We presented and studied the influence of several variants of our technique, including task definition approaches. We show improved learning ability over standard fine-tuning, resulting in efficient use of only few training samples. Meta-AWB has the ability to generalise quickly and learns to solve the computational color constancy problem in a camera agnostic fashion. This provides the potential for high accuracy performance as new sensors become available yet mitigates arduous and time-consuming calibration of training imagery, required for fully-supervised approaches. 
One would expect to reach better generalisation performance with more imaging content variability per camera. Future work will investigate different base learner components, alternatives to meta-learning to address the few-shot learning problem and more diverse task definition approaches (e.g. scene content).



{\small
\bibliographystyle{ieee}
\bibliography{egbib_cvpr}
}

\clearpage
\newpage
\begin{appendices}

\section{Additional qualitative results}

\begin{figure*}[t]
 \centering
     \subfigure[Input image]{
     \label{fig:supp:results1:log:subfig1}
     \includegraphics[width=0.22\linewidth]{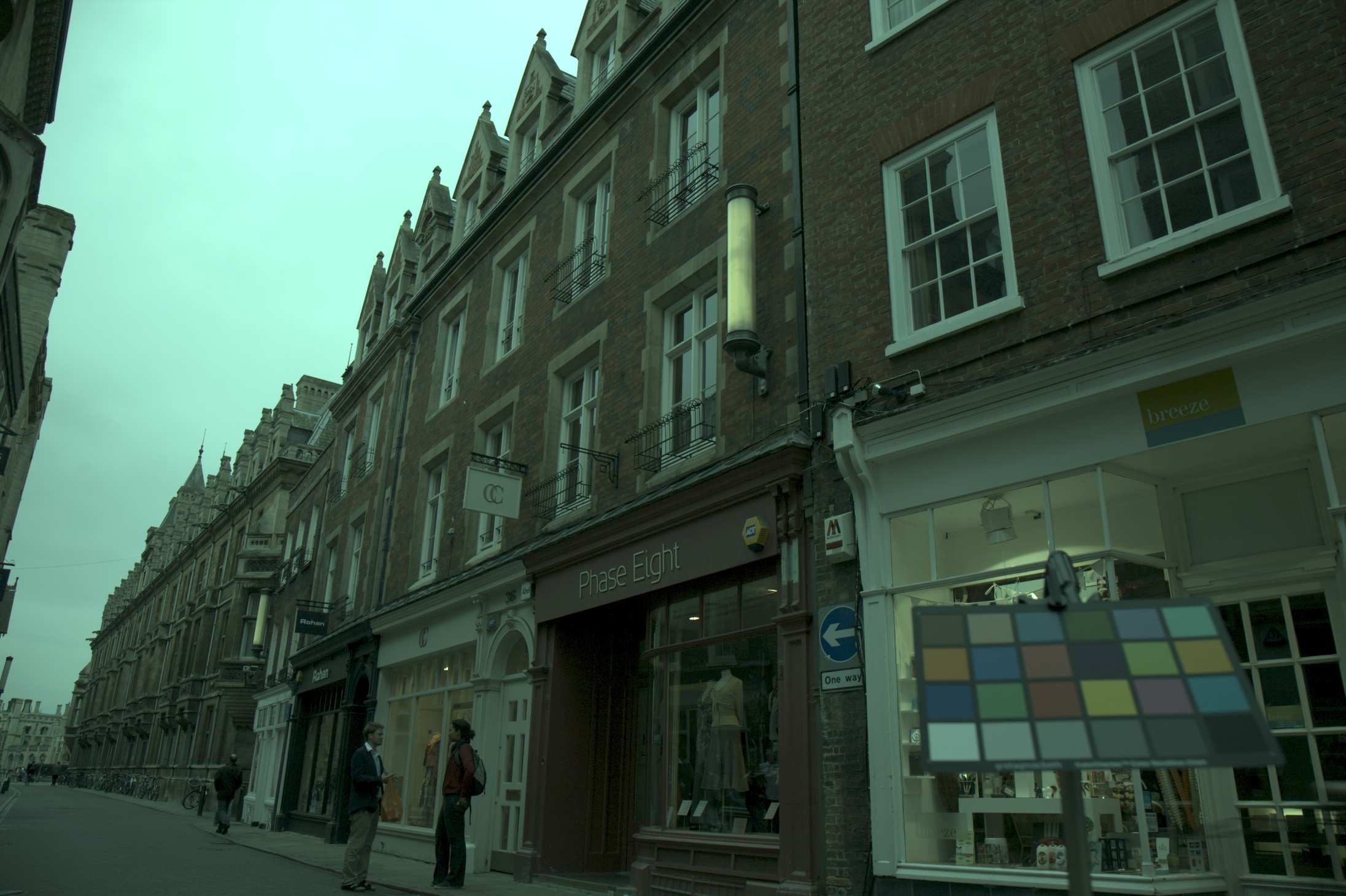}}
     \subfigure[Ground-truth solution]{
     \label{fig:supp:results1:log:subfig2}
     \includegraphics[width=0.22\linewidth]{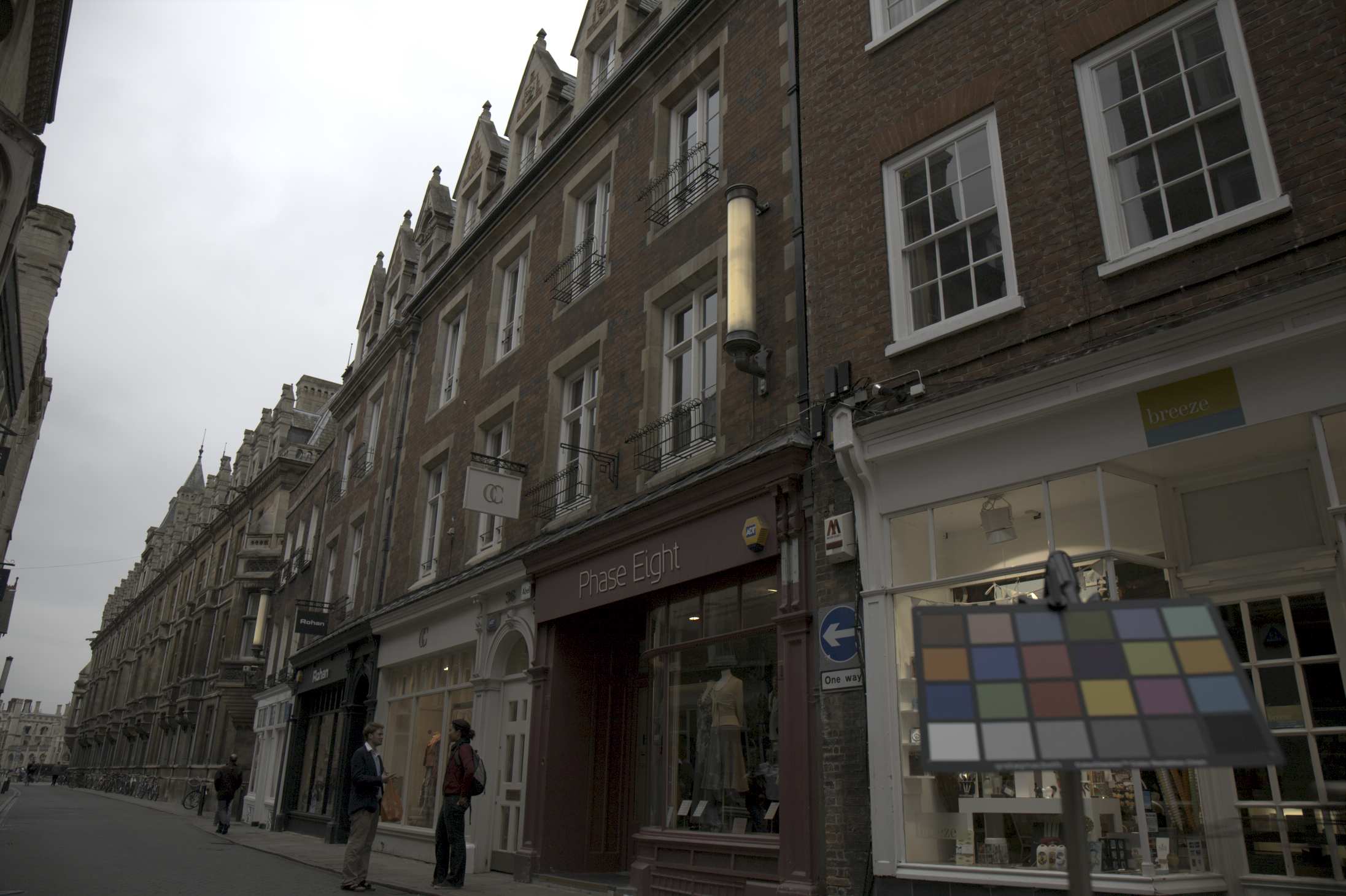}}
     \subfigure[Baseline fine-tuning, ($0.073\degree$)]{
     \label{fig:supp:results1:log:subfig3}
     \includegraphics[width=0.22\linewidth]{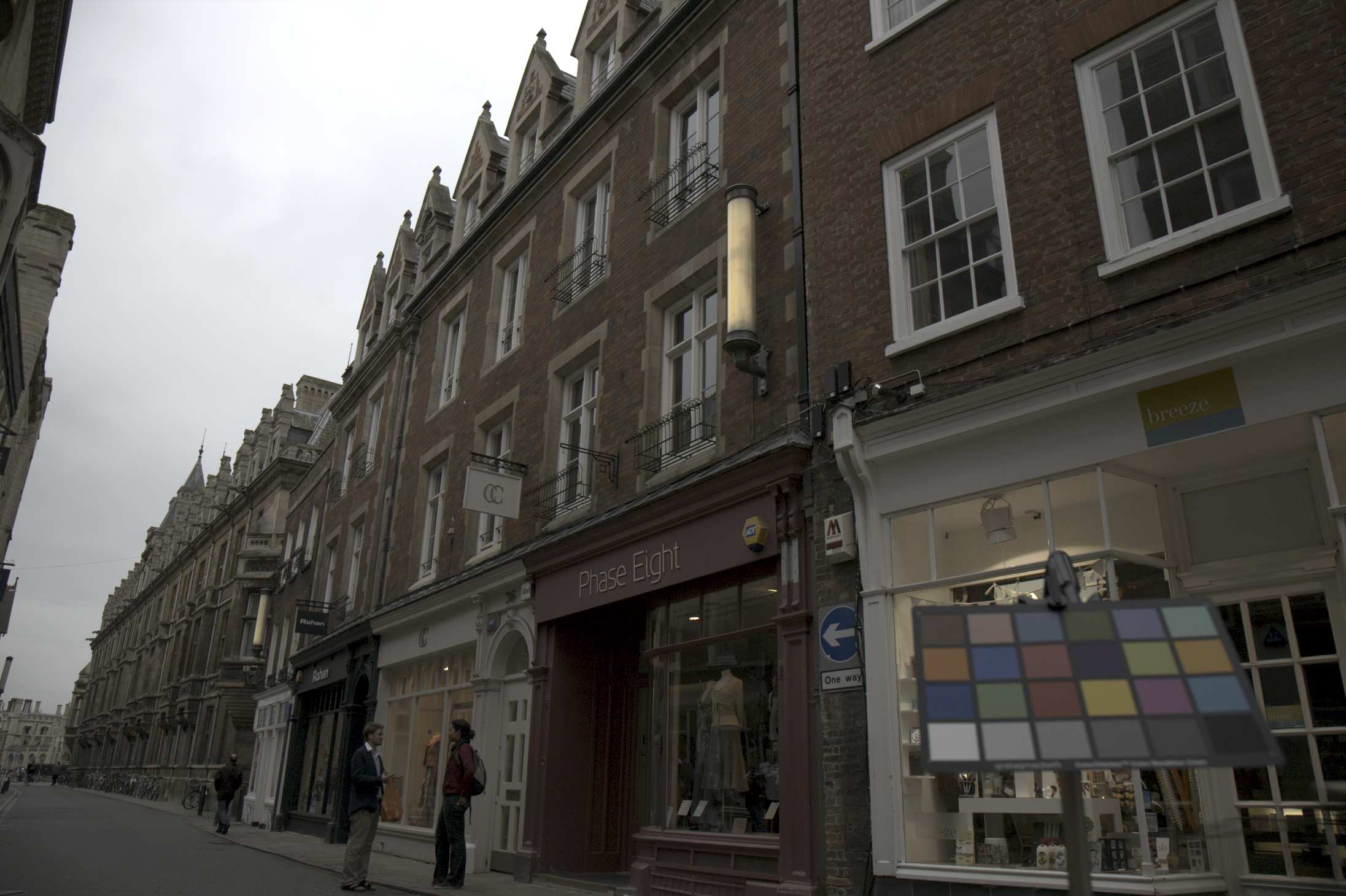}}
     \subfigure[Meta-AWB, ($0.308\degree$)]{
     \label{fig:supp:results1:log:subfig4}
     \includegraphics[width=0.22\linewidth]{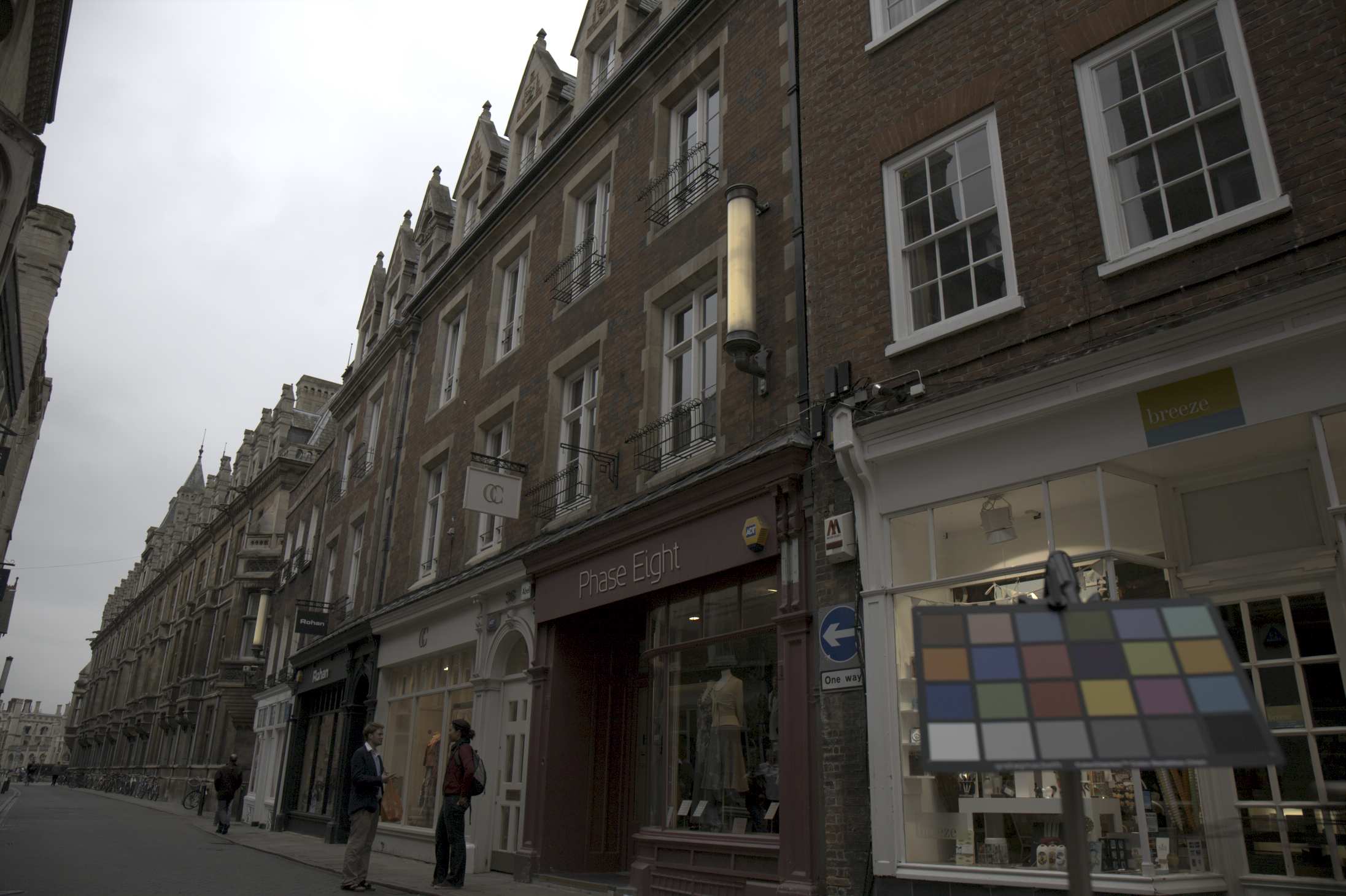}}\\
     \subfigure[Input image]{
     \label{fig:supp:results1:log:subfig5}
     \includegraphics[width=0.22\linewidth]{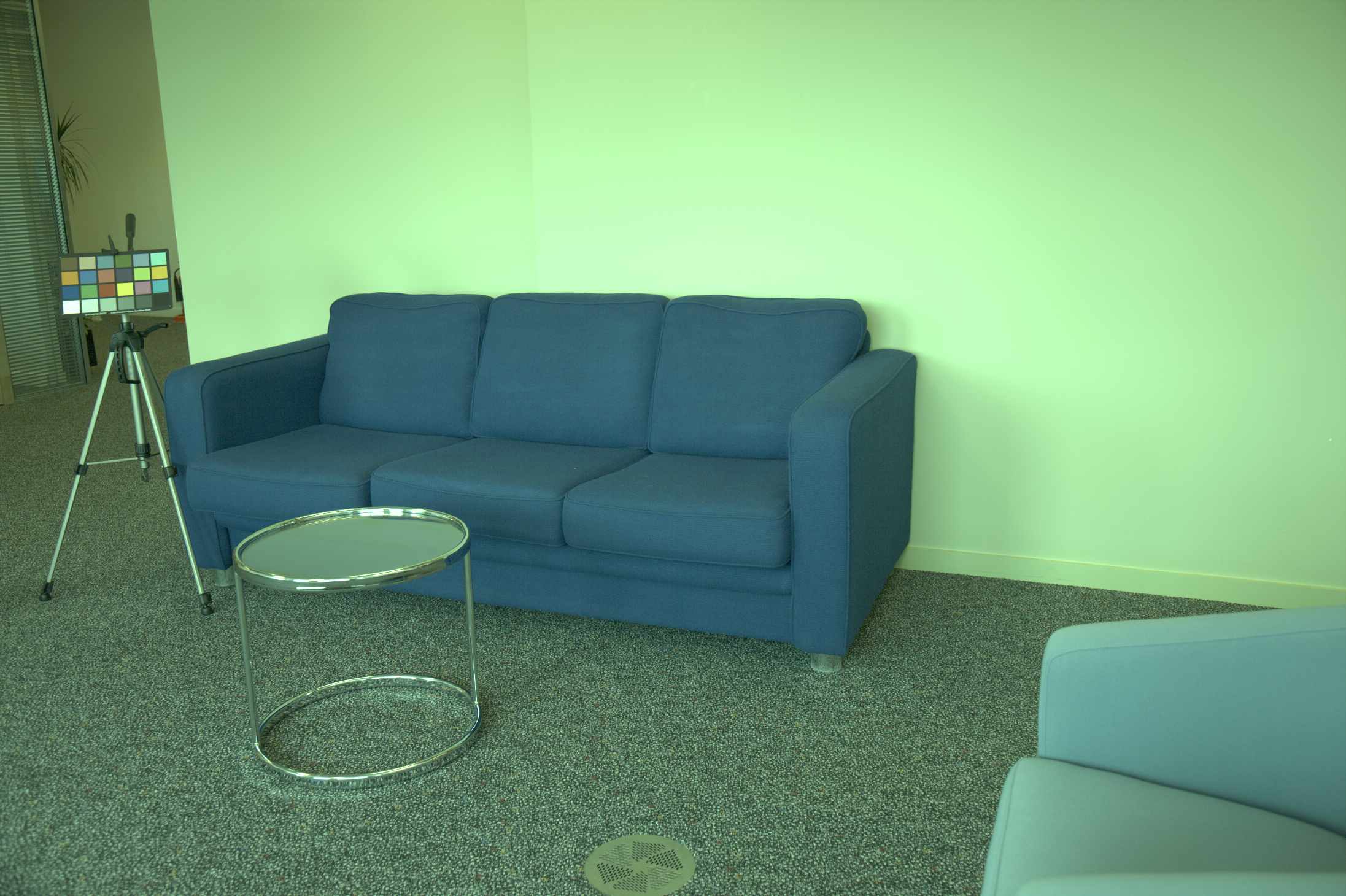}}
     \subfigure[Ground-truth solution]{
     \label{fig:supp:results1:log:subfig6}
     \includegraphics[width=0.22\linewidth]{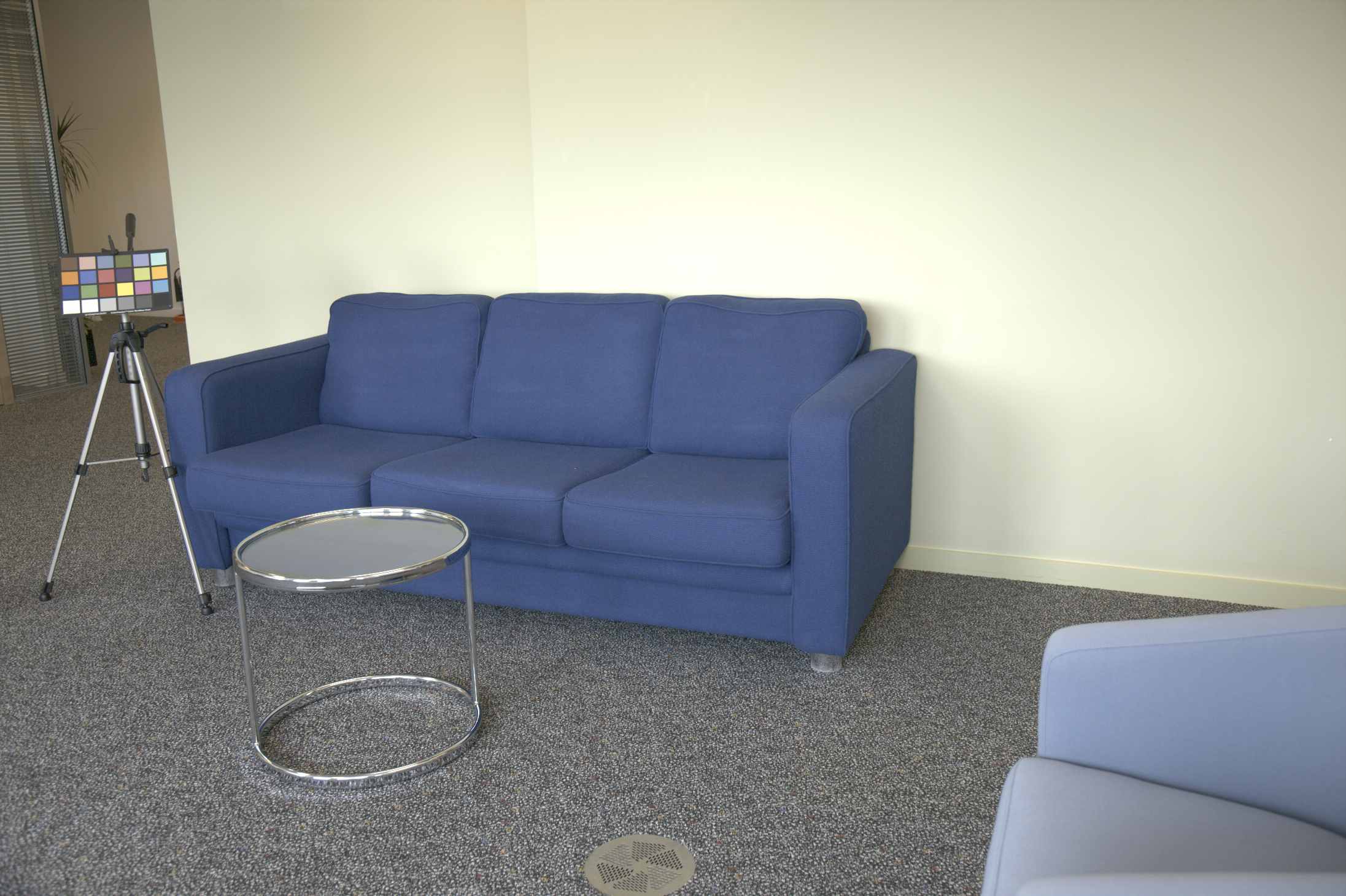}}
     \subfigure[Baseline fine-tuning, ($7.956\degree$)]{
     \label{fig:supp:results1:log:subfig7}
     \includegraphics[width=0.22\linewidth]{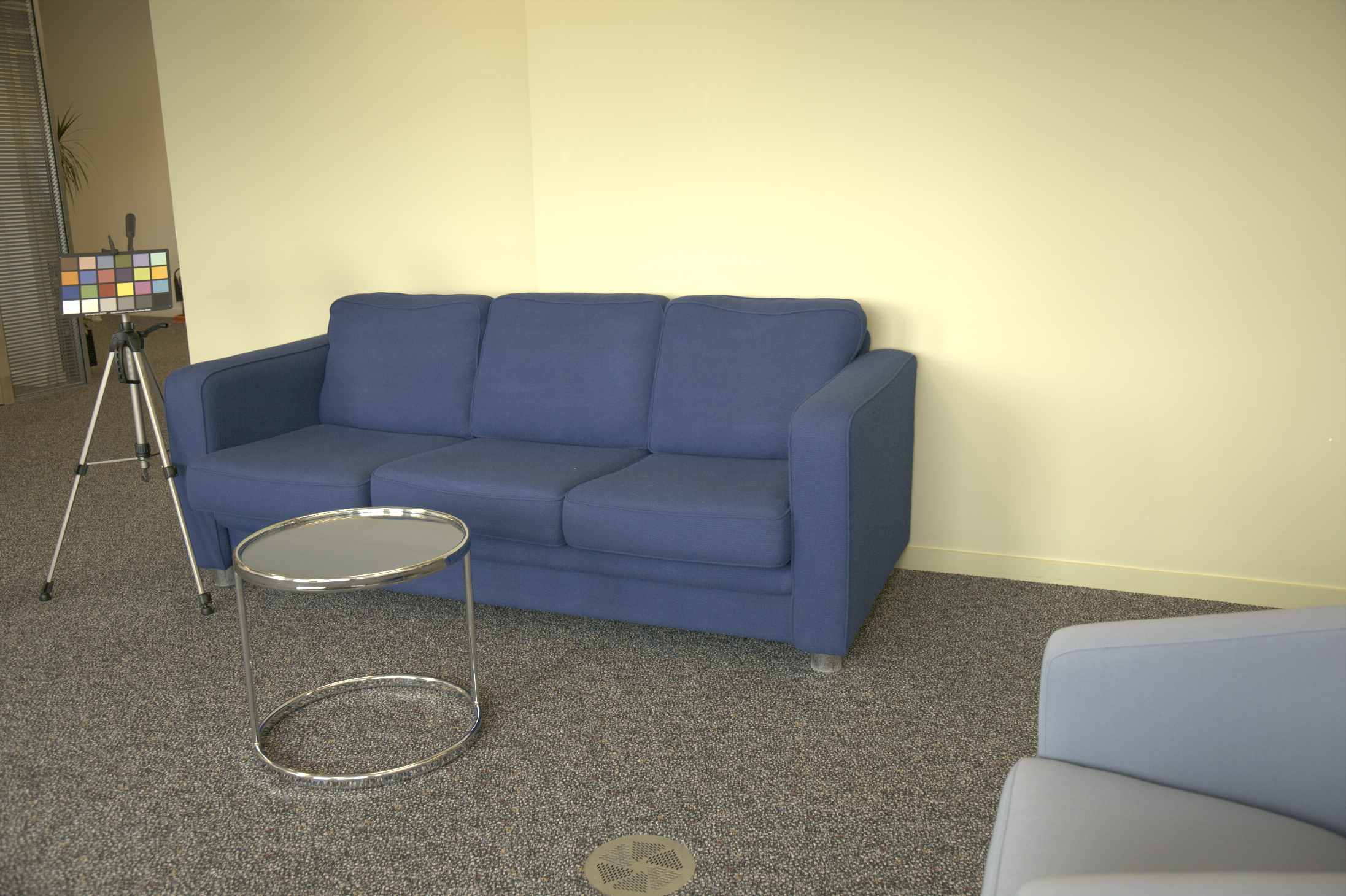}}
     \subfigure[Meta-AWB, ($1.855\degree$)]{
     \label{fig:supp:results1:log:subfig8}
     \includegraphics[width=0.22\linewidth]{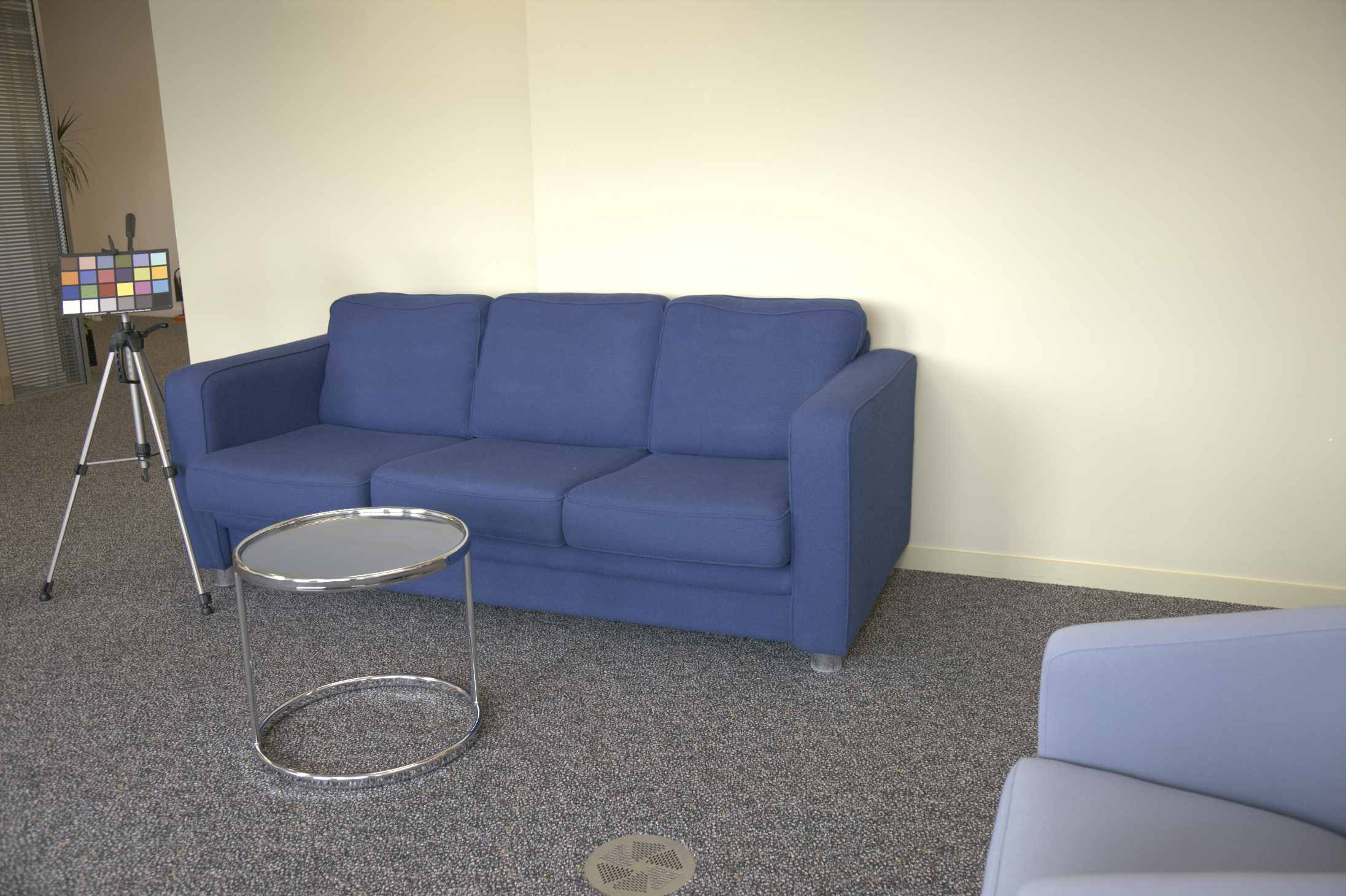}}\\
     \subfigure[Input image]{
     \label{fig:supp:results1:log:subfig9}
     \includegraphics[width=0.22\linewidth]{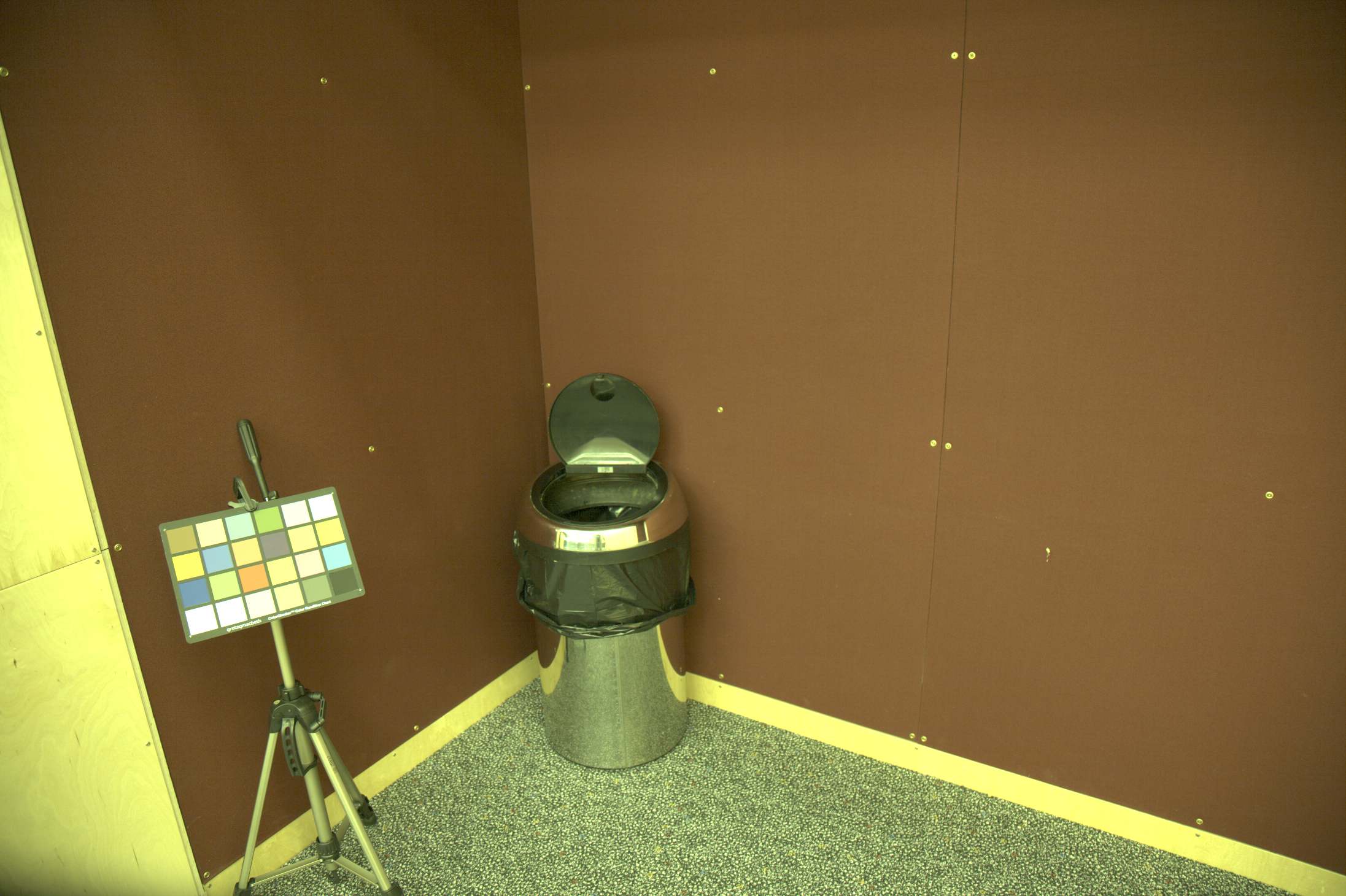}}
     \subfigure[Ground-truth solution]{
     \label{fig:supp:results1:log:subfig10}
     \includegraphics[width=0.22\linewidth]{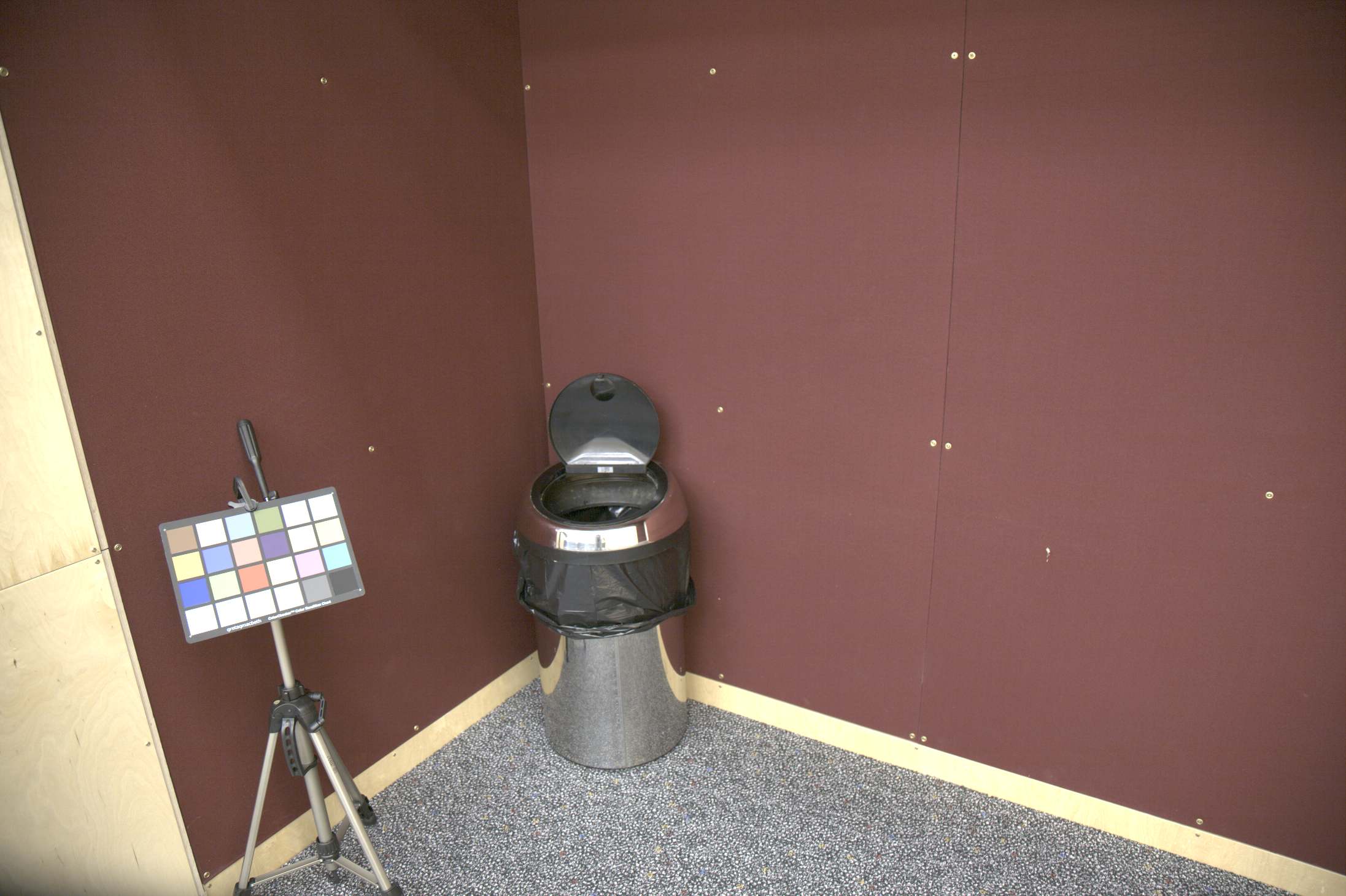}}
     \subfigure[Baseline fine-tuning, ($17.23\degree$)]{
     \label{fig:supp:results1:log:subfig11}
     \includegraphics[width=0.22\linewidth]{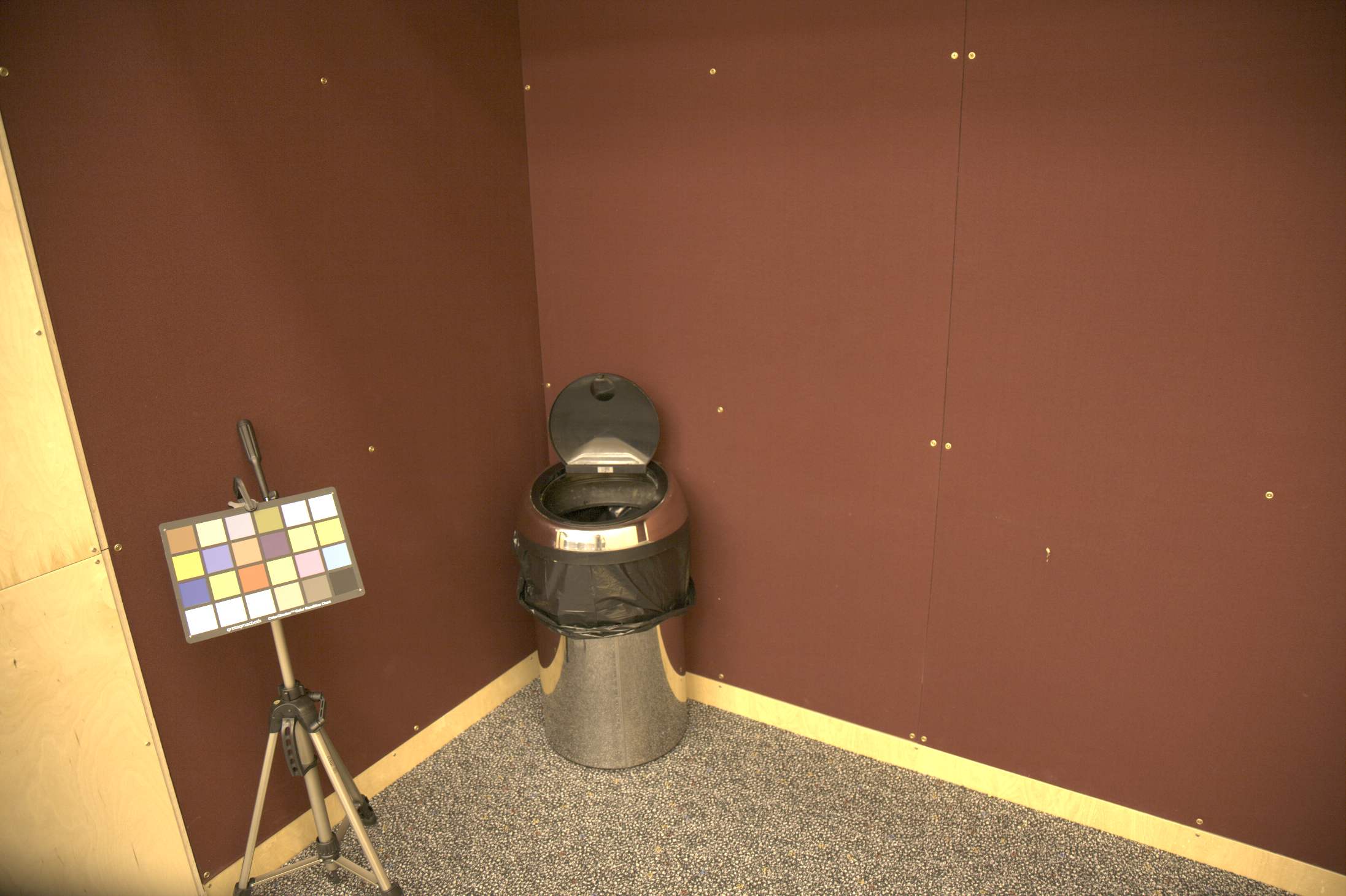}}
     \subfigure[Meta-AWB, ($3.798\degree$)]{
     \label{fig:supp:results1:log:subfig12}
     \includegraphics[width=0.22\linewidth]{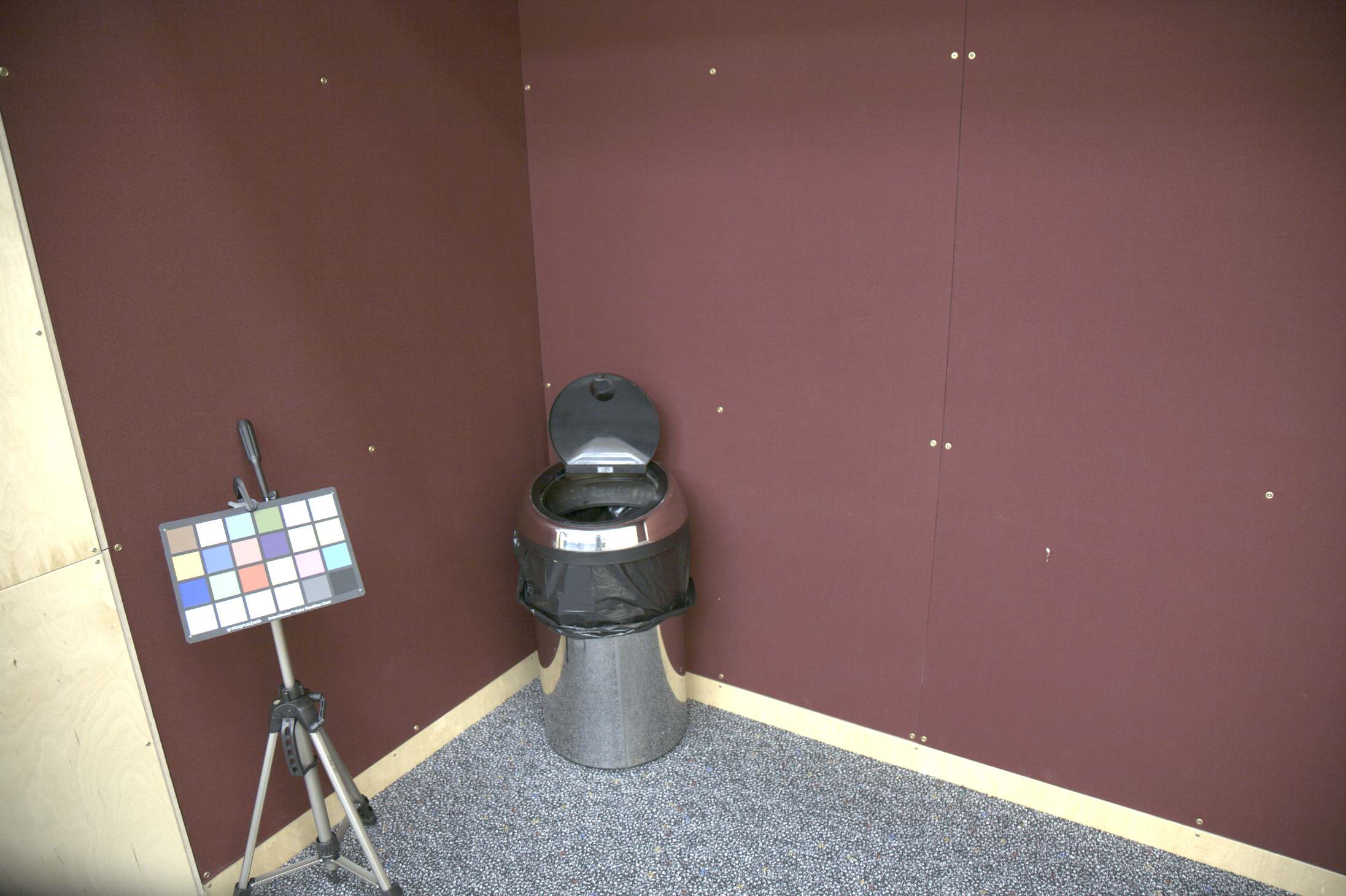}}\\
     \subfigure[Input image]{
     \label{fig:supp:results1:log:subfig13}
     \includegraphics[width=0.22\linewidth]{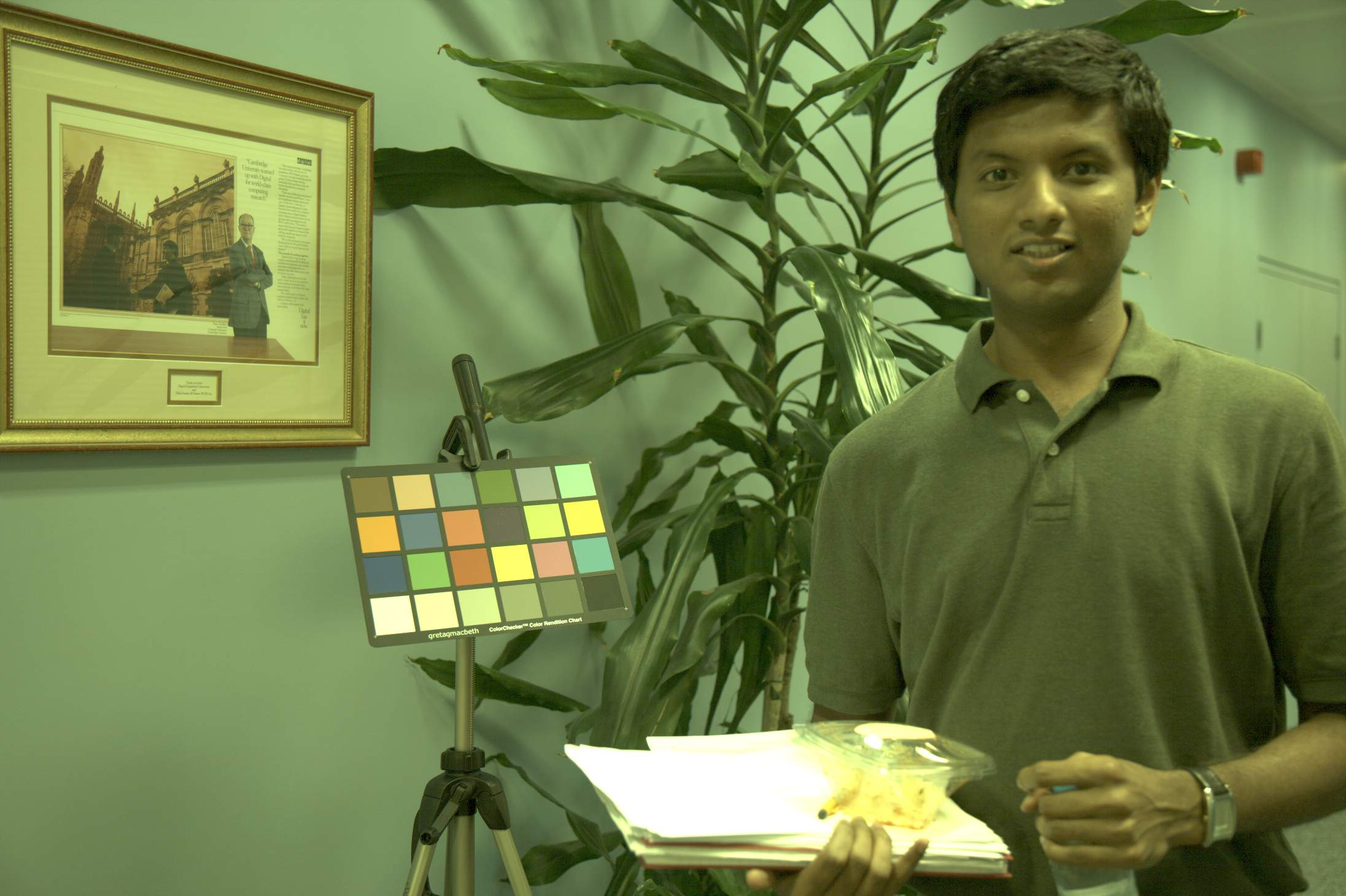}}
     \subfigure[Ground-truth solution]{
     \label{fig:supp:results1:log:subfig14}
     \includegraphics[width=0.22\linewidth]{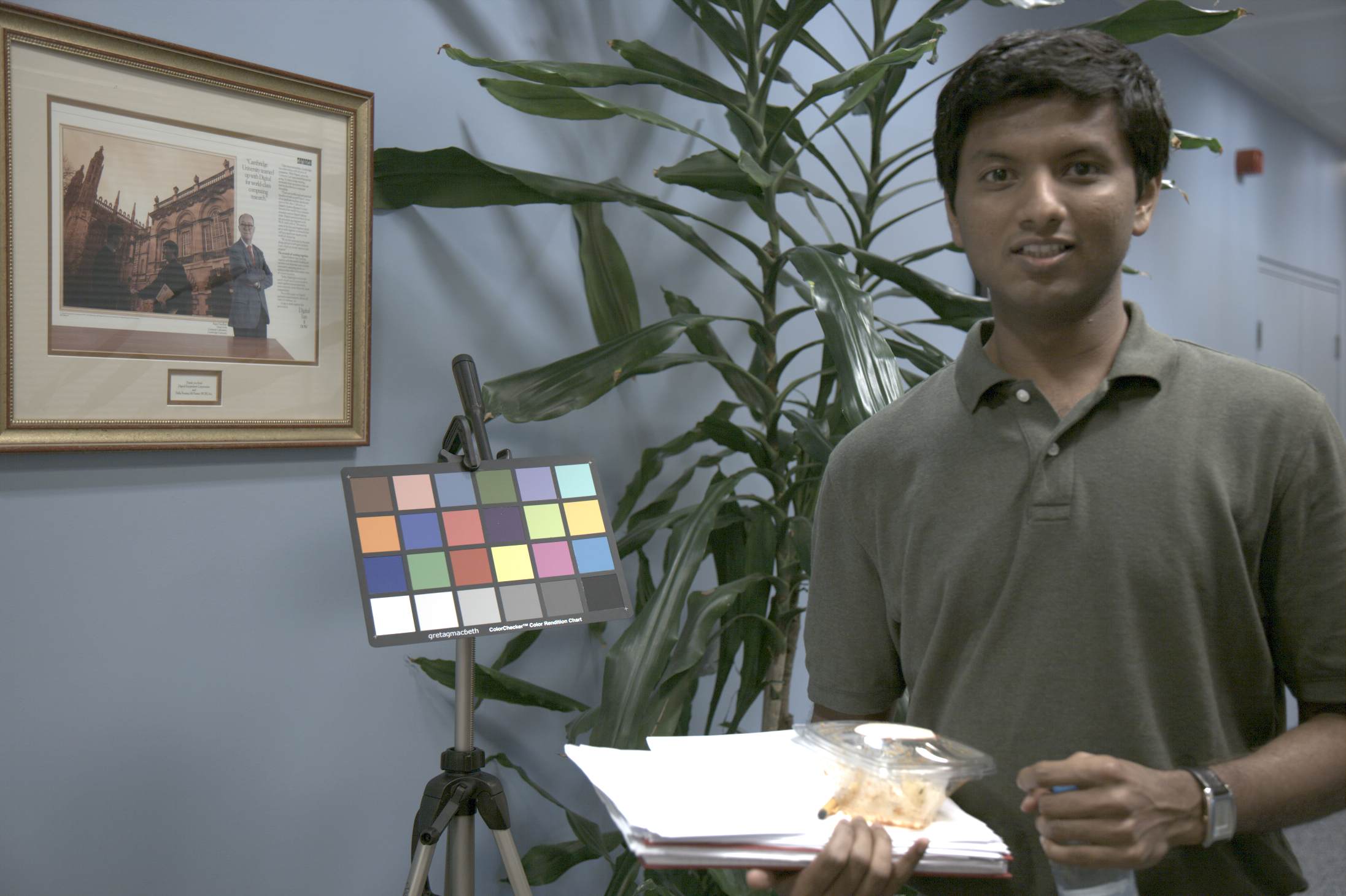}}
     \subfigure[Baseline fine-tuning, ($18.55\degree$)]{
     \label{fig:supp:results1:log:subfig15}
     \includegraphics[width=0.22\linewidth]{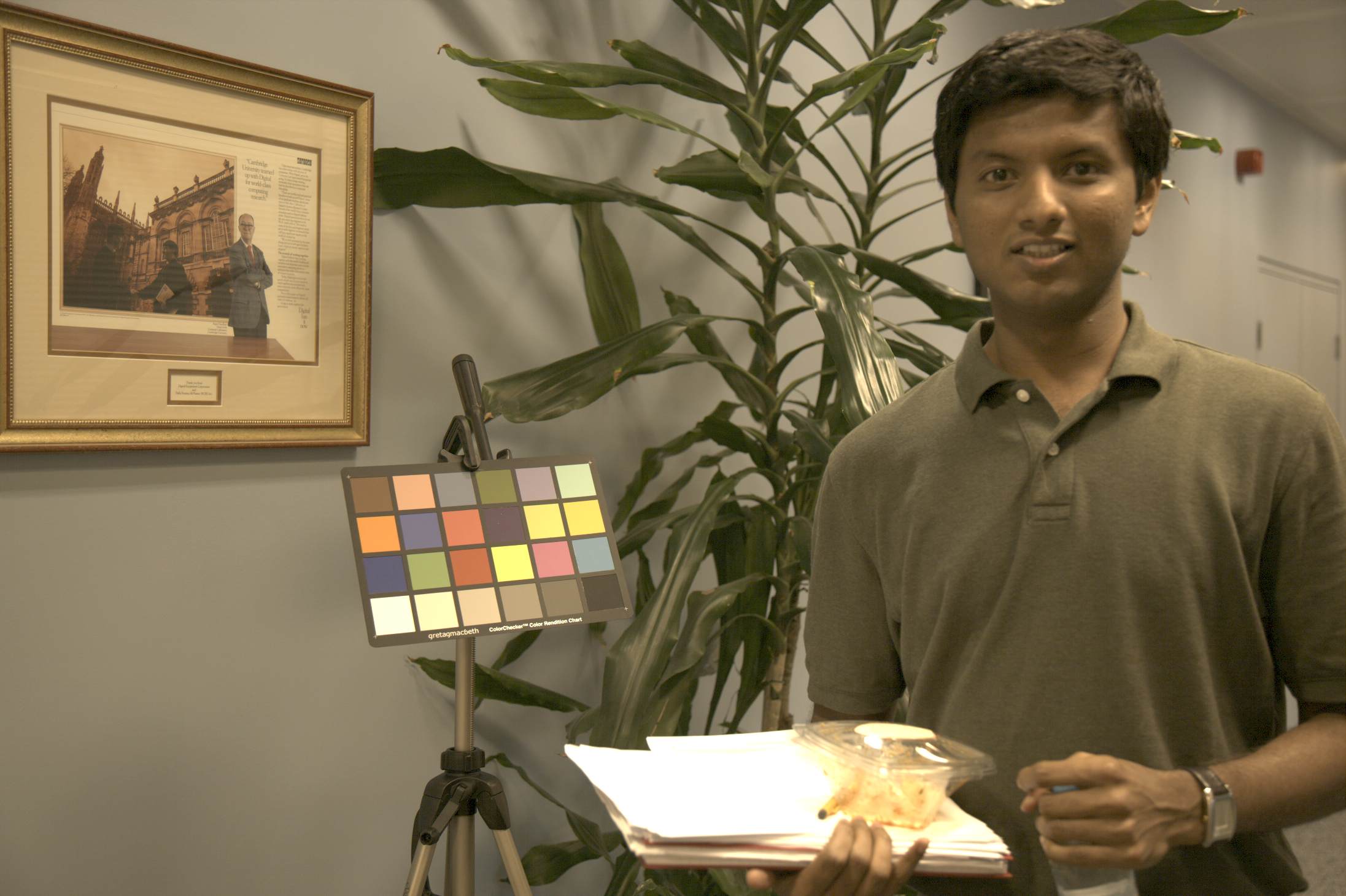}}
     \subfigure[Meta-AWB, ($6.931\degree$)]{
     \label{fig:supp:results1:log:subfig16}
     \includegraphics[width=0.22\linewidth]{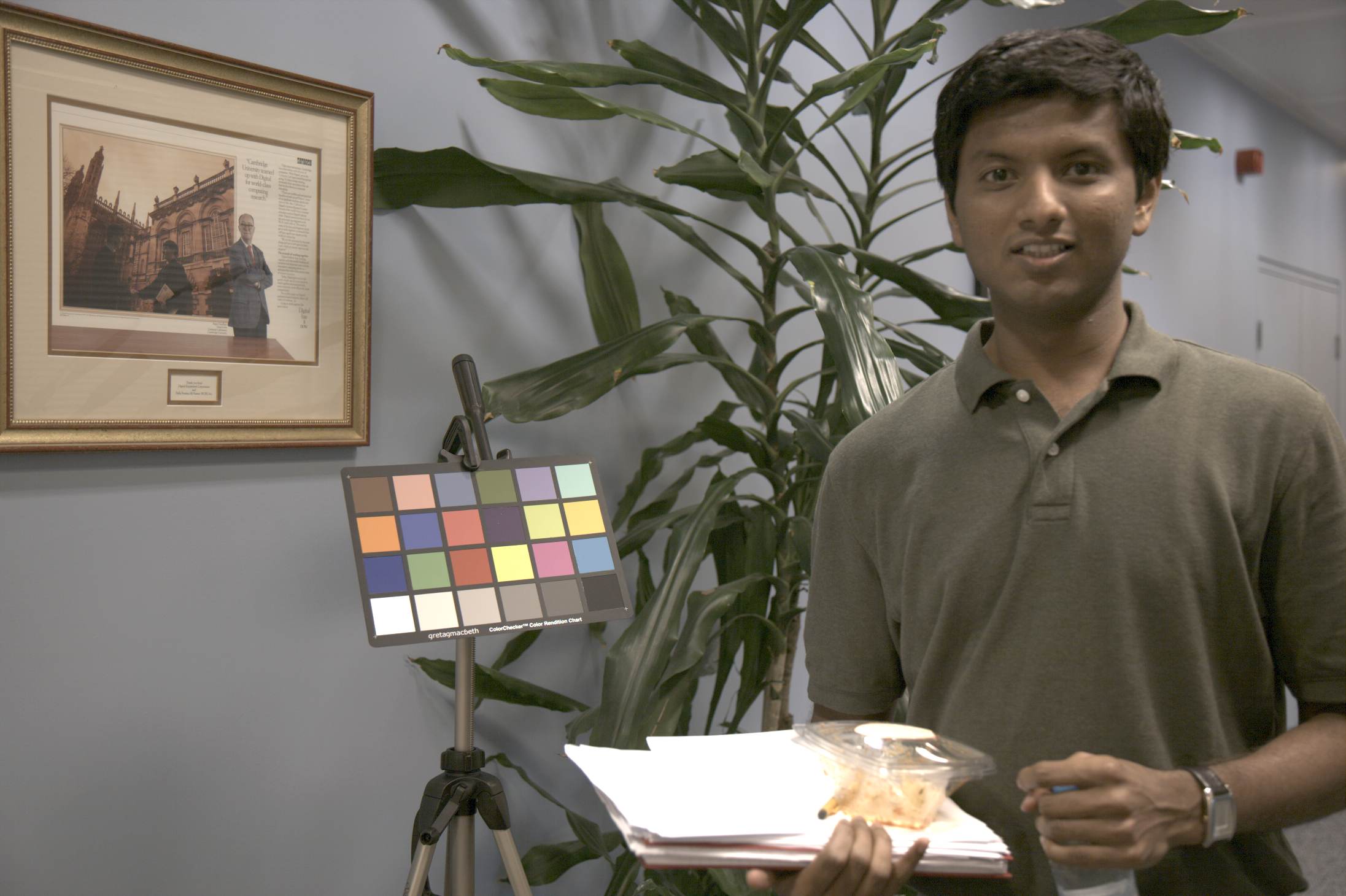}}\\
     \subfigure[Input image]{
     \label{fig:supp:results1:log:subfig17}
     \includegraphics[width=0.22\linewidth]{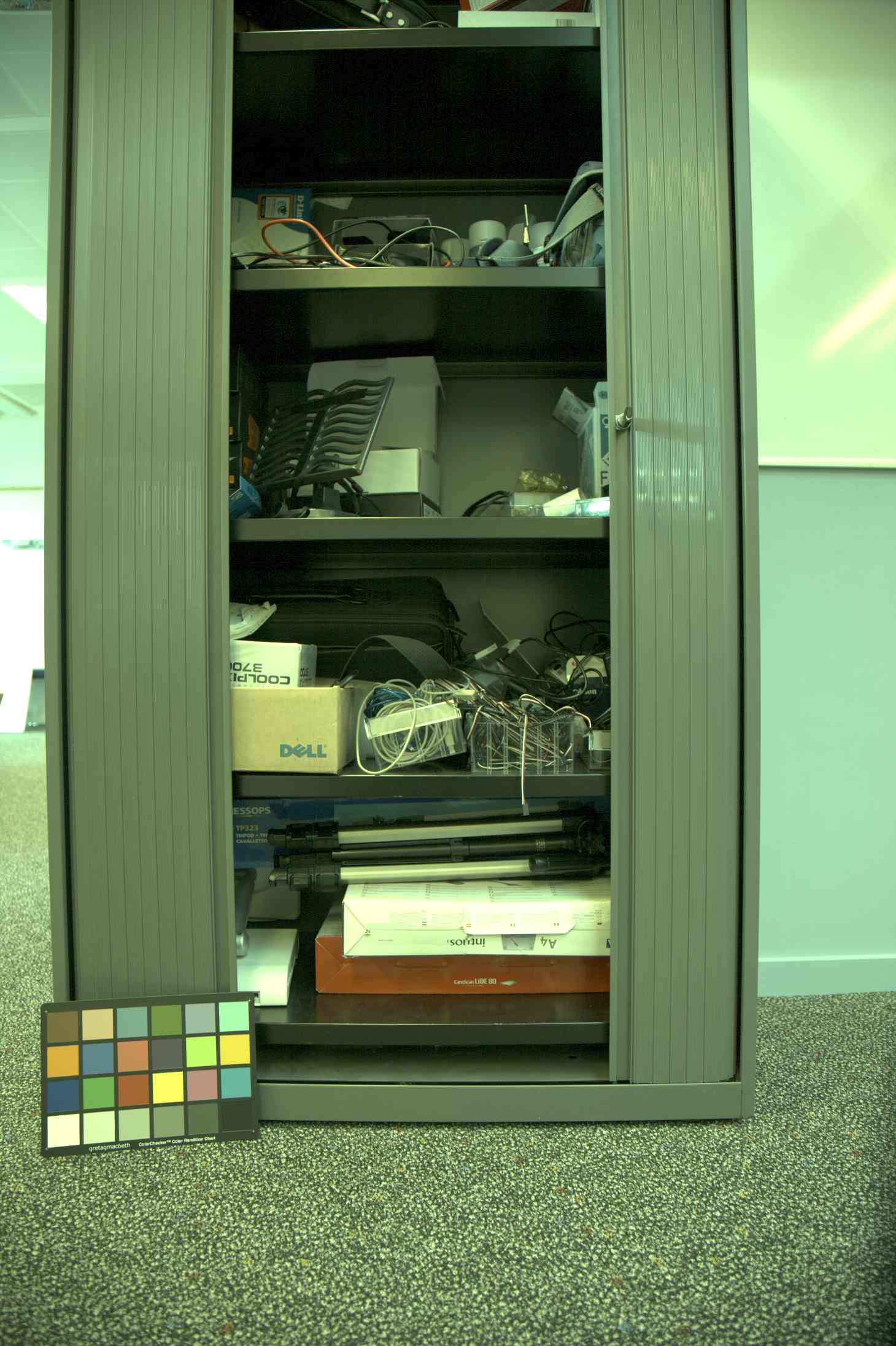}}
     \subfigure[Ground-truth solution]{
     \label{fig:supp:results1:log:subfig18}
     \includegraphics[width=0.22\linewidth]{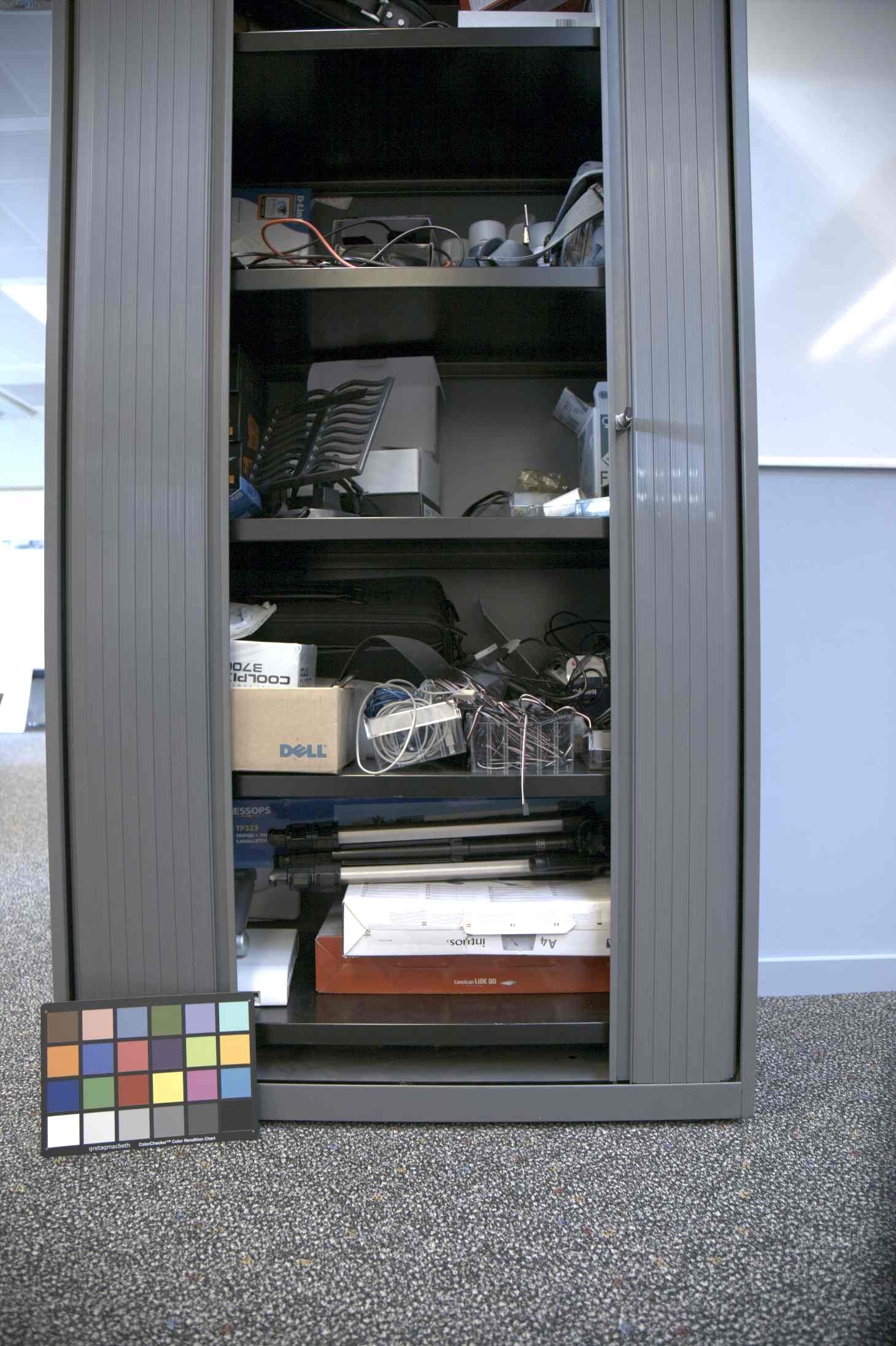}}
     \subfigure[Baseline fine-tuning, ($17.67\degree$)]{
     \label{fig:supp:results1:log:subfig19}
     \includegraphics[width=0.22\linewidth]{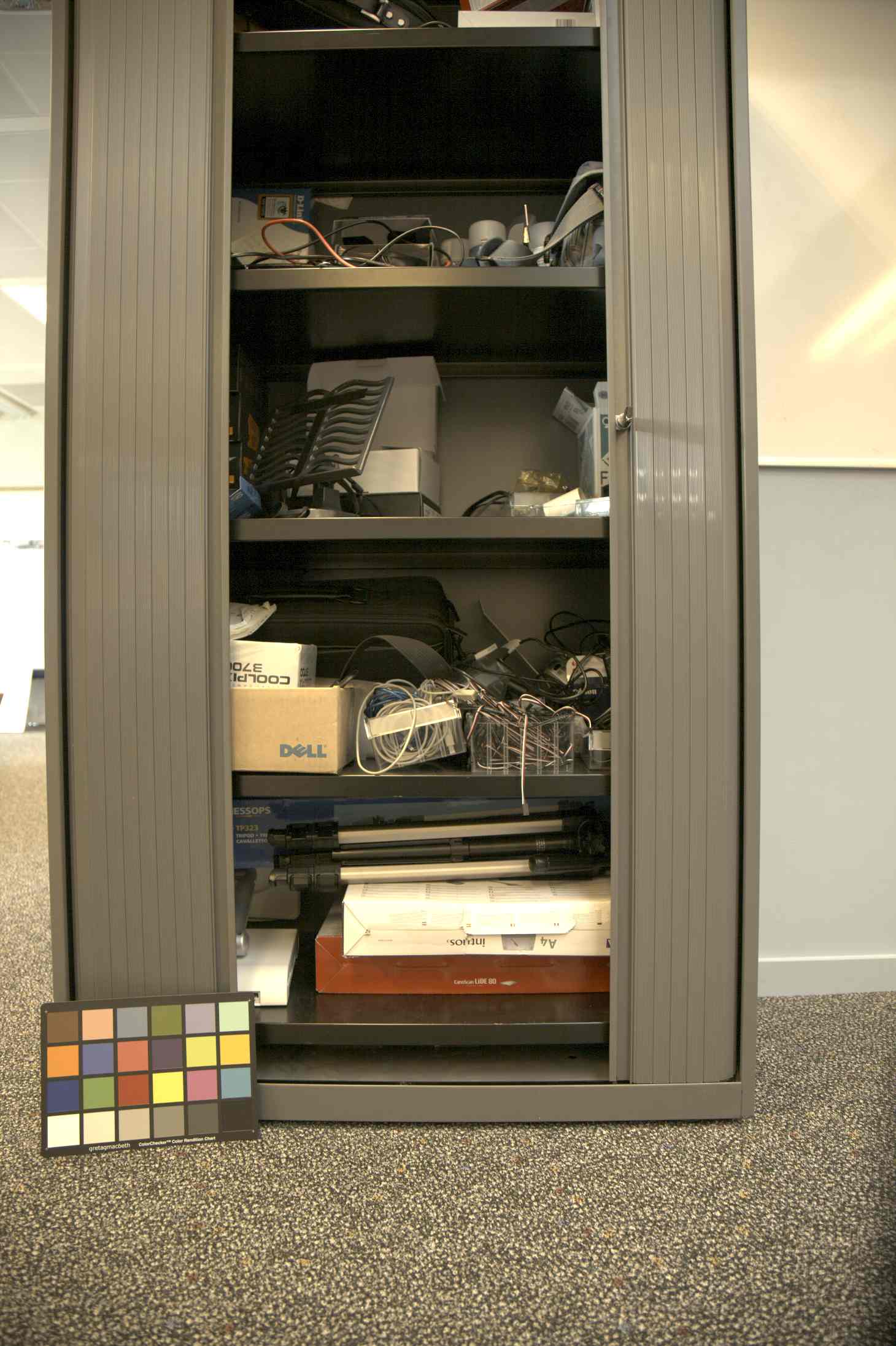}}
     \subfigure[Meta-AWB, ($16.14\degree$)]{
     \label{fig:supp:results1:log:subfig20}
     \includegraphics[width=0.22\linewidth]{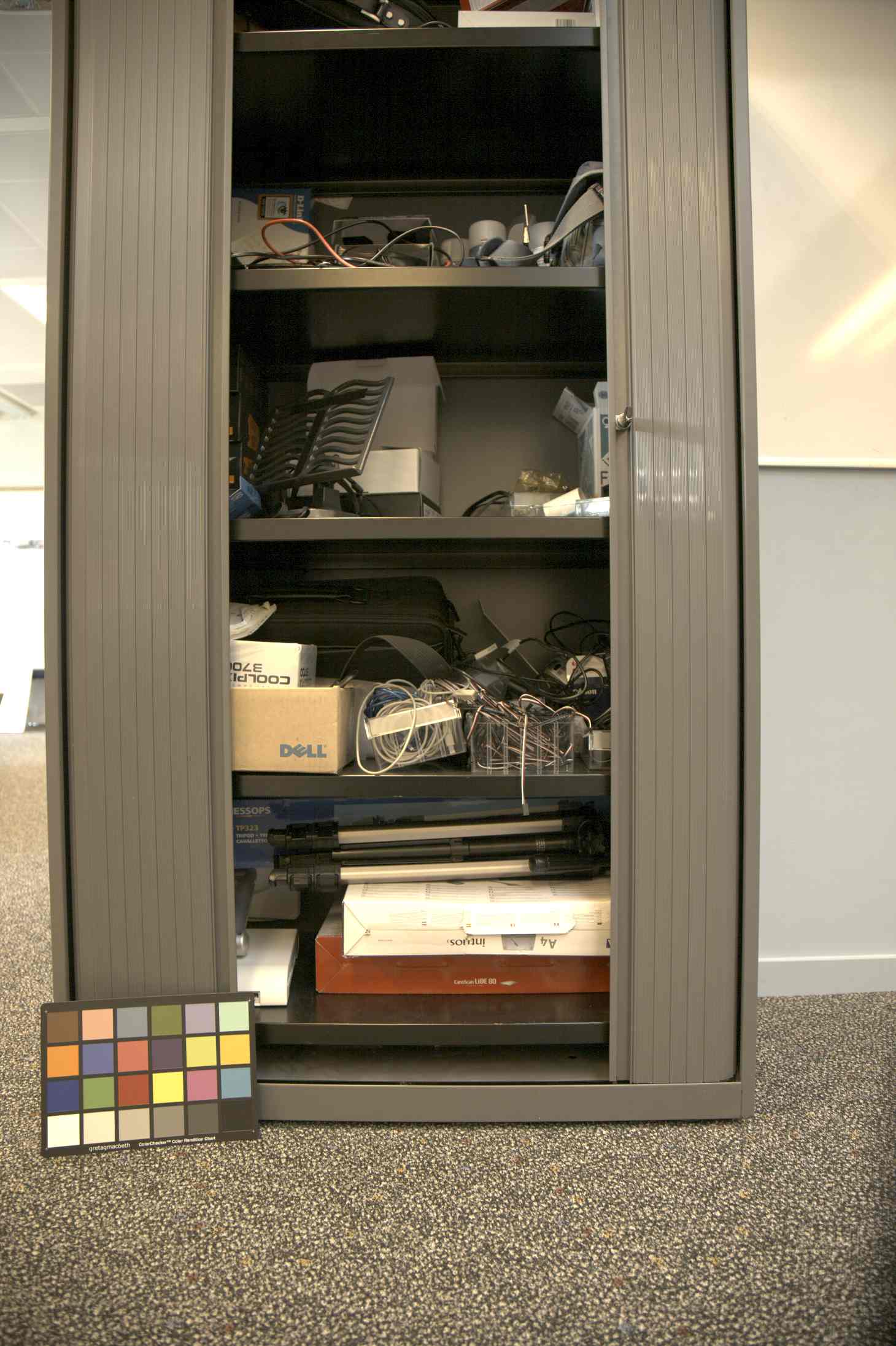}}\\
     \subfigure[Input image]{
     \label{fig:supp:results1:log:subfig21}
     \includegraphics[width=0.22\linewidth]{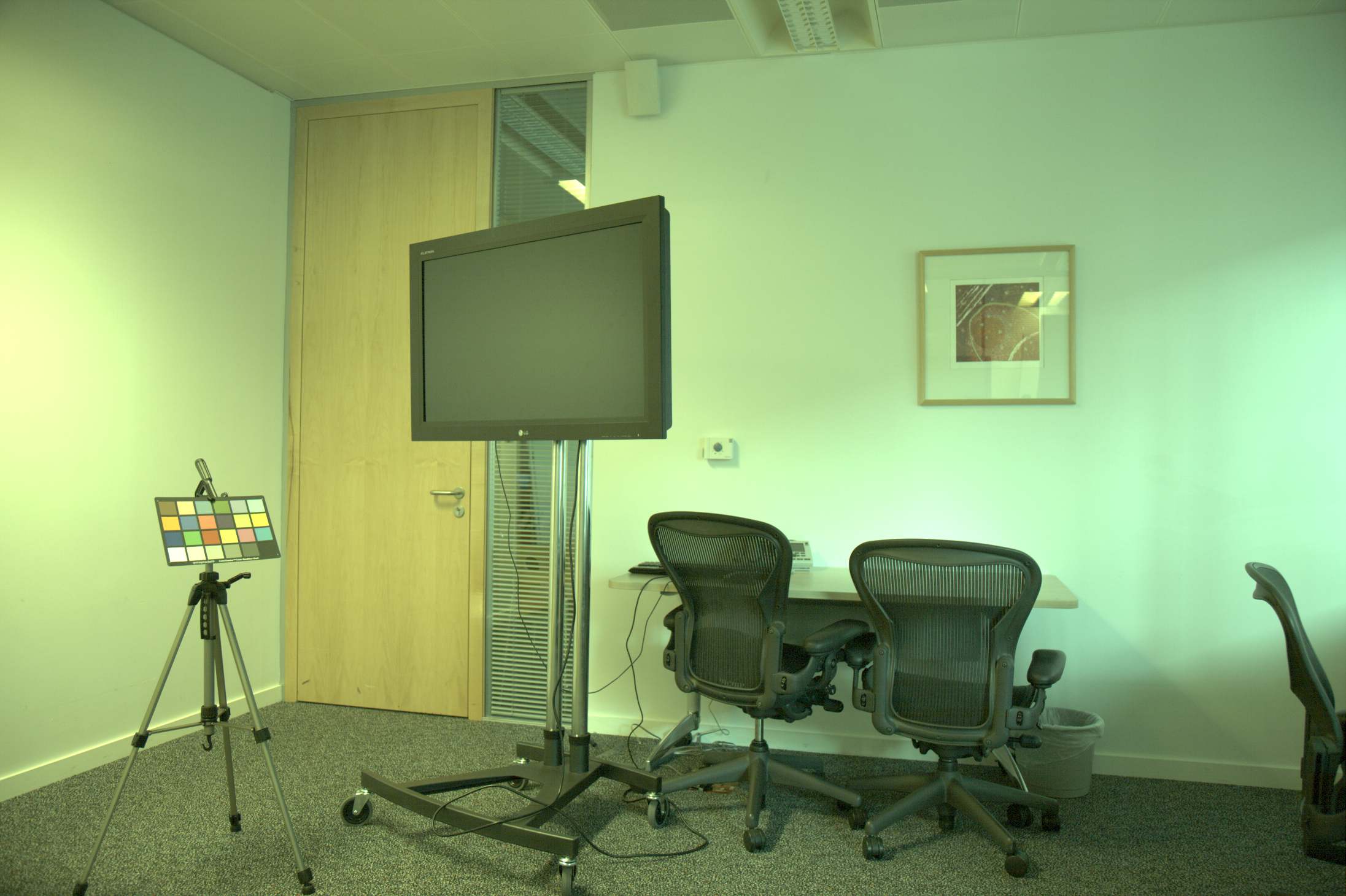}}
     \subfigure[Ground-truth solution]{
     \label{fig:supp:results1:log:subfig22}
     \includegraphics[width=0.22\linewidth]{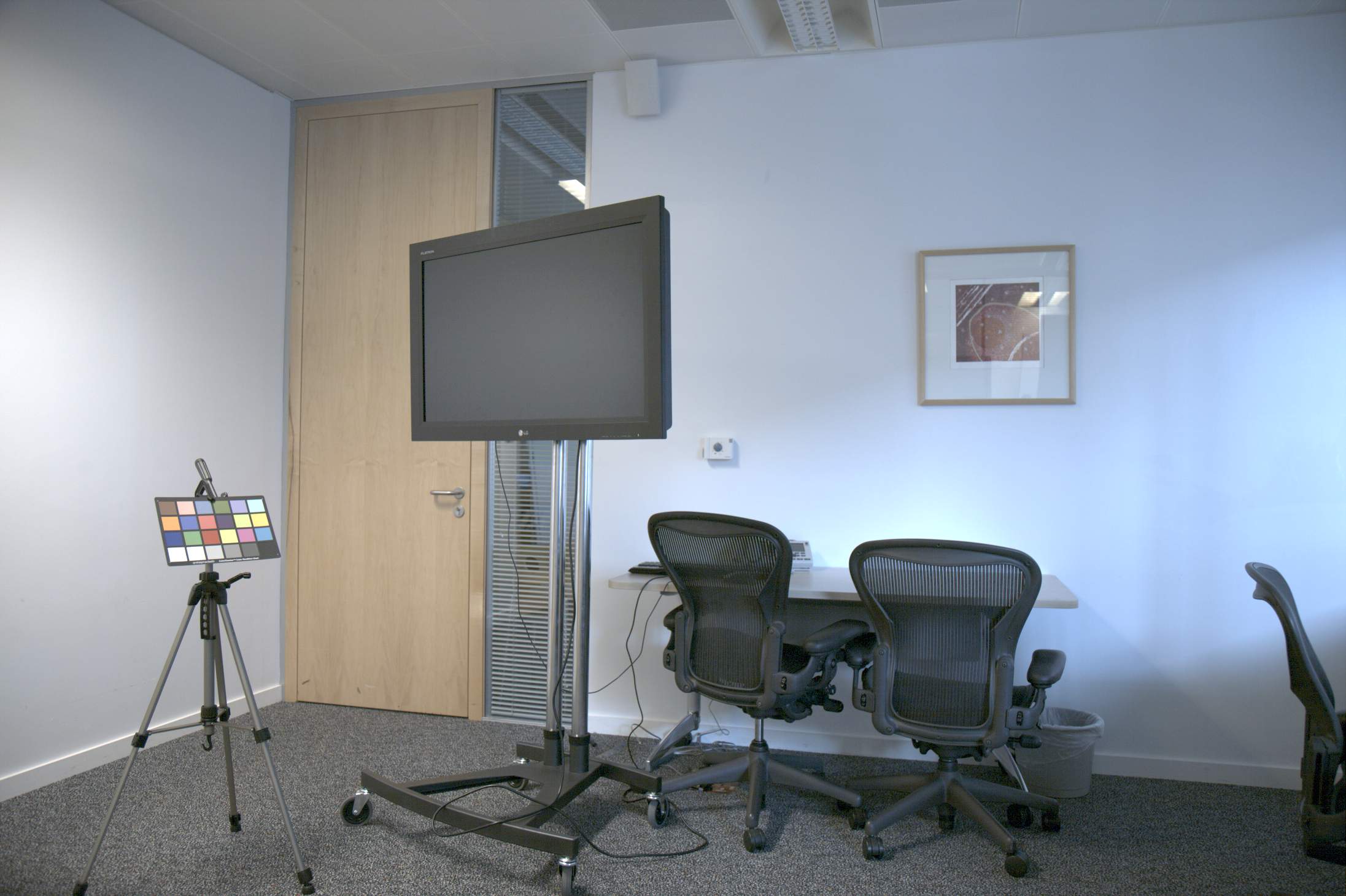}}
     \subfigure[Baseline fine-tuning, ($22.08\degree$)]{
     \label{fig:supp:results1:log:subfig23}
     \includegraphics[width=0.22\linewidth]{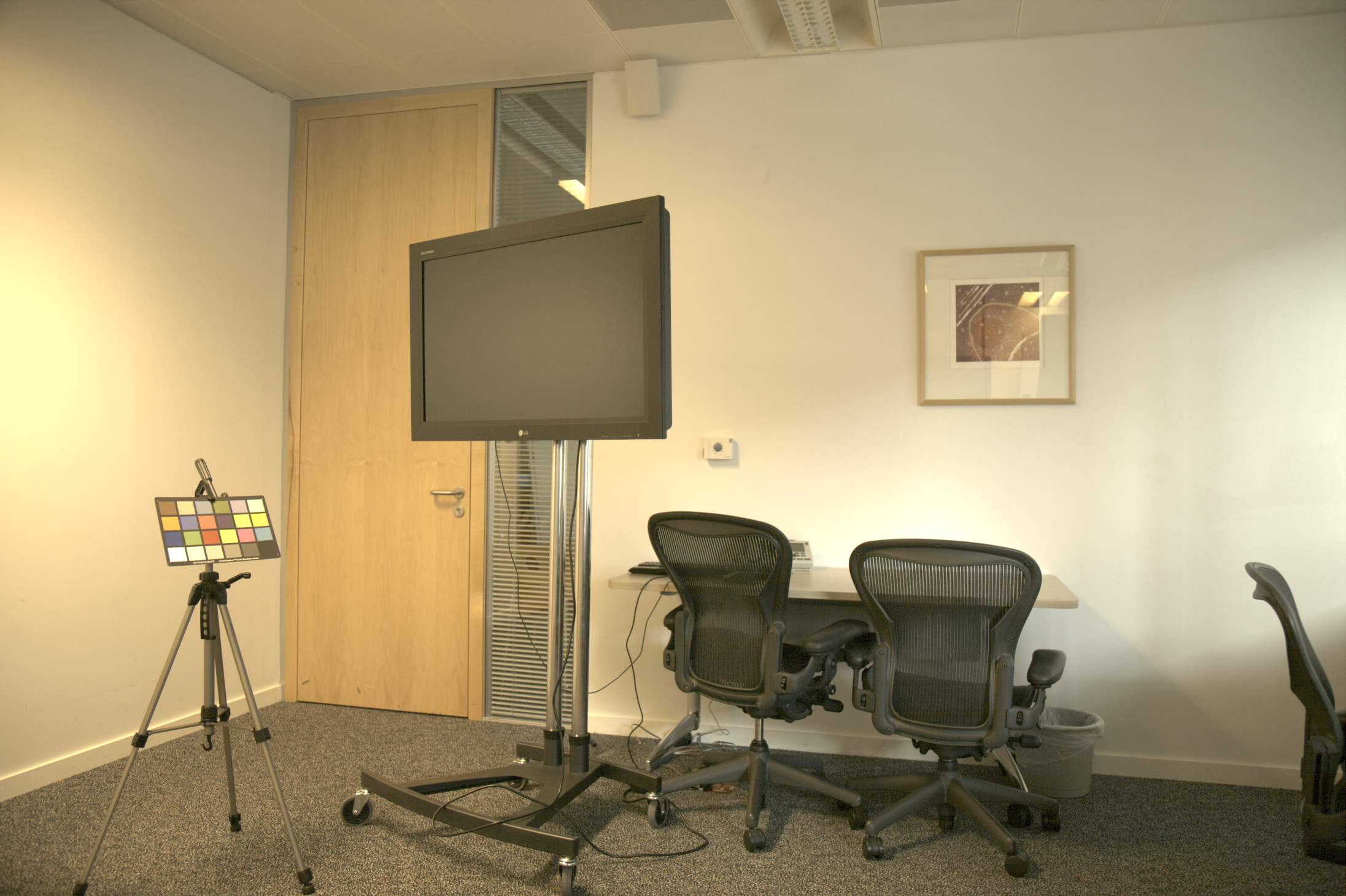}}
     \subfigure[Meta-AWB, ($15.20\degree$)]{
     \label{fig:supp:results1:log:subfig24}
     \includegraphics[width=0.22\linewidth]{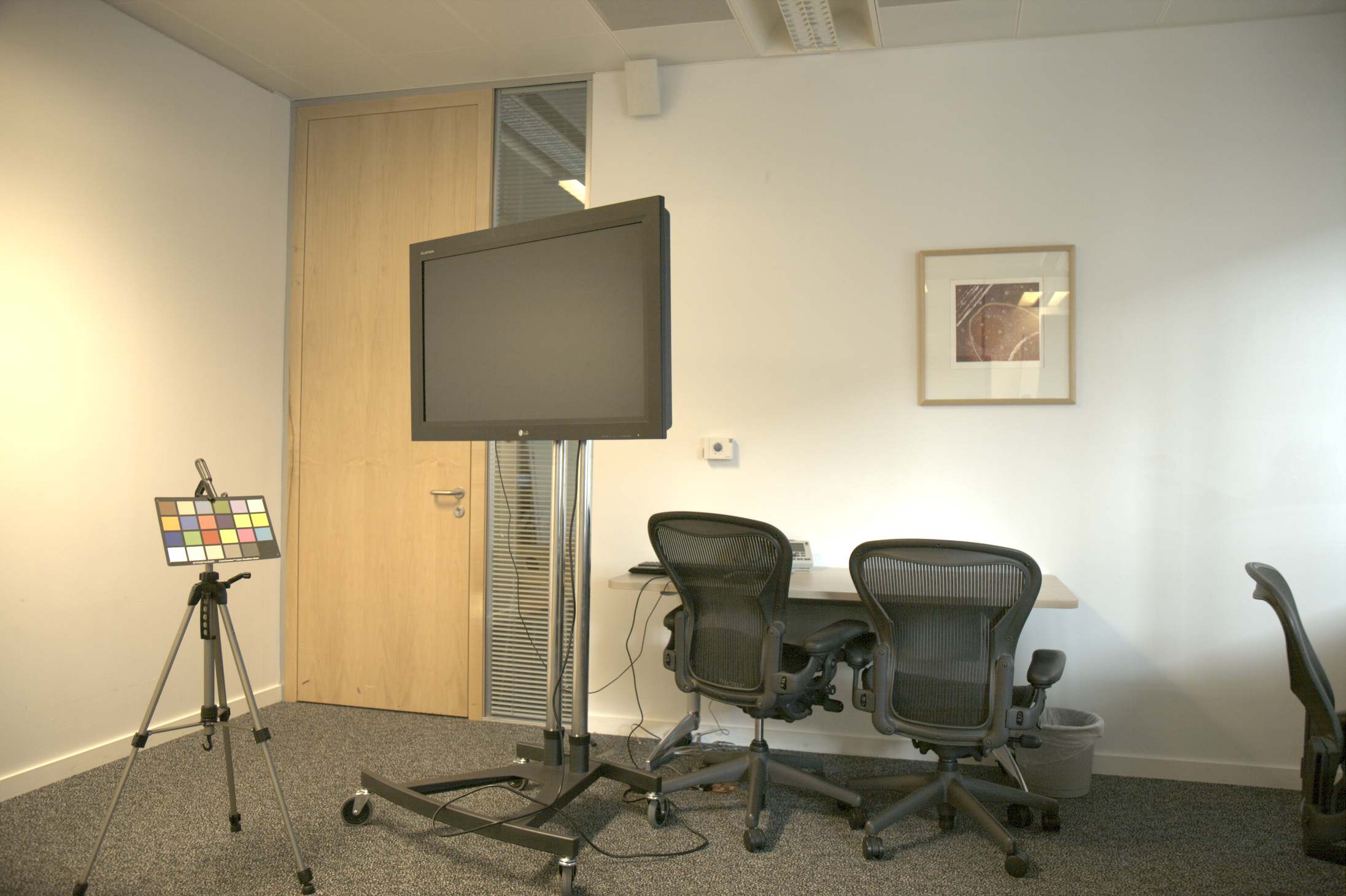}}\\
 \caption{\small{For each scene we present the input image produced by the camera alongside the ground-truth white-balanced image. Images are shown in sRGB space and normalized to the $97.5$th percentile. For both our ``Meta-AWB`` approach and the baseline algorithm we show the white-balanced image, as well as the angular-error of the estimated illumination in degrees, with respect to the ground-truth.}}
 \label{fig:supp:results1:log}
 \end{figure*}

In Figure~\ref{fig:supp:results1:log} we provide additional qualitative results in the form of test images from the Gehler-Shi dataset~\cite{gehler2008bayesian,shi2000re}. For each image we show the input image produced by the camera and a white-balanced image corrected using the ground-truth illumination. We also show the output of our model (``Meta-AWB''), and that of the baseline fine-tuning approach reported in the paper. Color checker boards are visible in the images, however the relevant areas are masked prior to inference. Images are shown in sRGB space and clipped at the $97.5$ percentile. 

In similar fashion to~\cite{barron2015convolutional}, we adopt the strategy of sorting the test images by the combined mean angular-error of the two evaluated methods. We present images of increasing average difficulty however images to report were selected by instead ordering from \emph{``hard''} to \emph{``easy''} and sampling with a logarithmic spacing, providing a greater number of samples that proved challenging, on average.

\section{Color temperature distributions for learning task formulation}

In our paper, we decompose the inter-camera color constancy problem into a set of regression tasks such that each task comprises images acquired from the same camera and 
under similar illumination settings. One approach we propose is to compute Correlated Color Temperature (CCT) histograms and define the images of each task as those that belong to a given histogram bin. 

\begin{figure}[!t]
\begin{center}
\subfigure[Canon 600D histogram]{
    \label{fig:supp:CCT:Canon600D_hist}
    \includegraphics[width=0.22\linewidth]{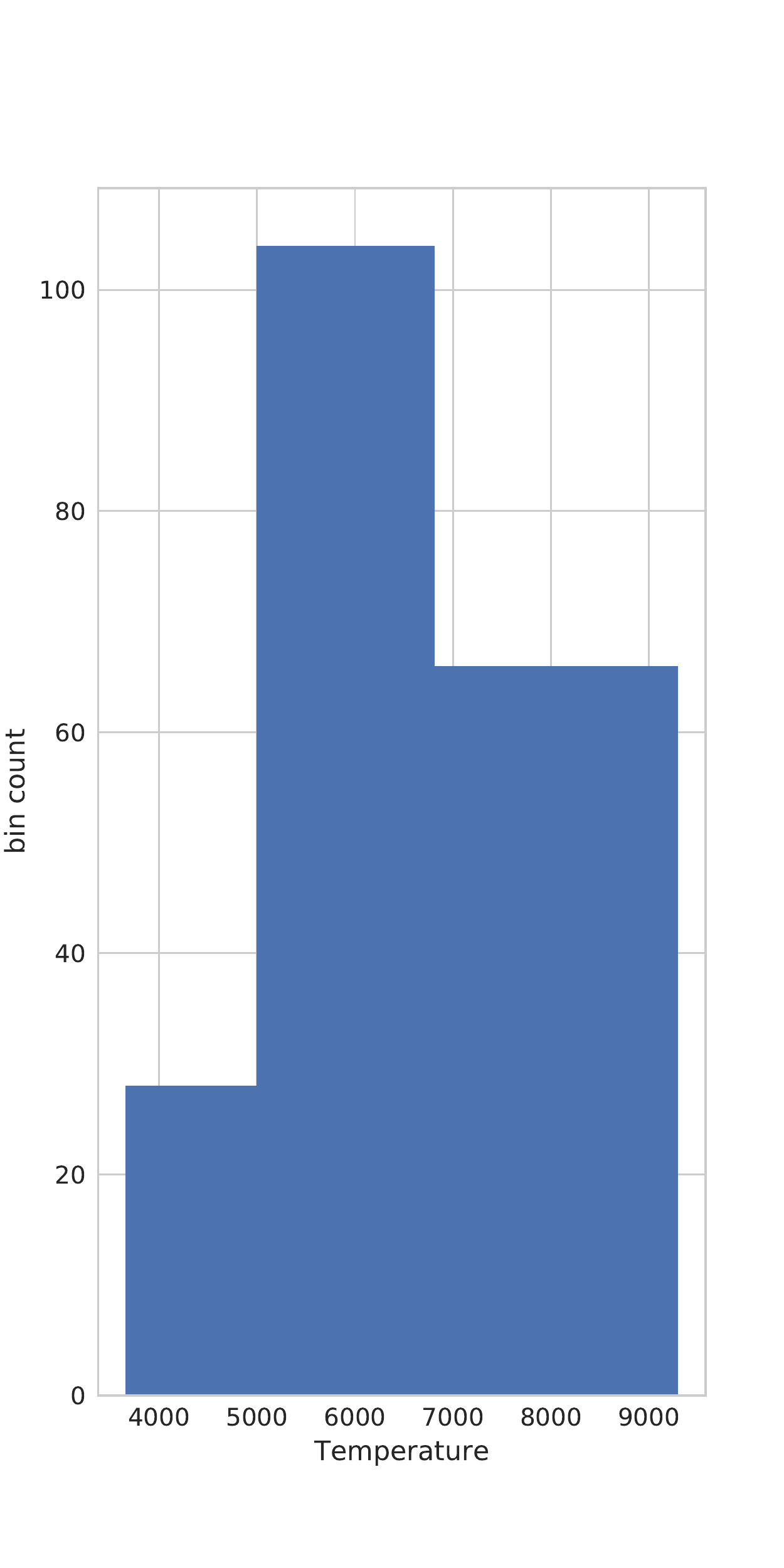}
    }
    \subfigure[Canon 600D GT illuminants in RGB space]{
    \label{fig:supp:CCT:Canon600D_GT}
    \includegraphics[width=0.70\linewidth]{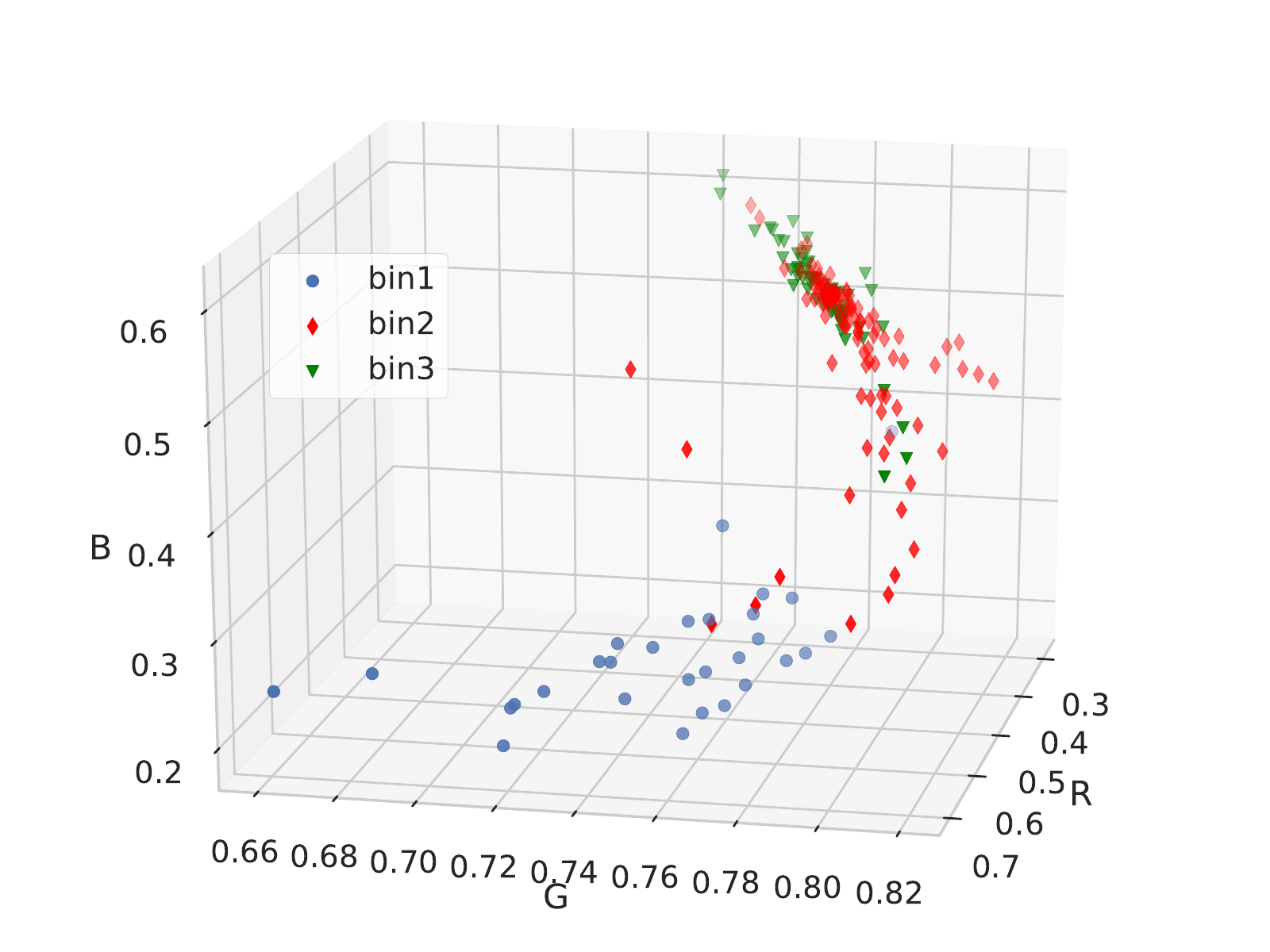}
    }
    \subfigure[Nikon D40 histogram]{
    \label{fig:supp:CCT:D40_hist}
    \includegraphics[width=0.22\linewidth]{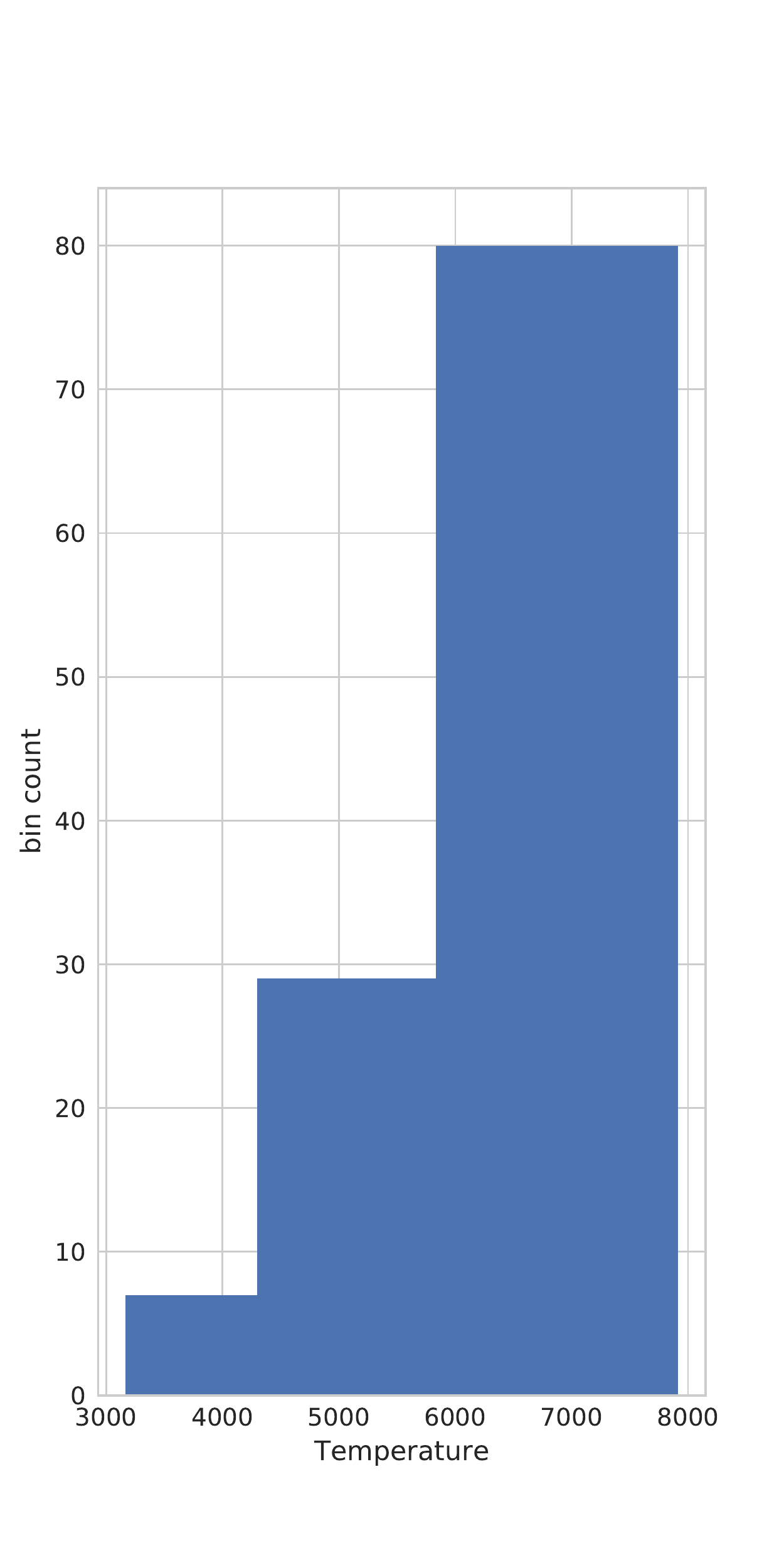}
    }
    \subfigure[Nikon D40 GT illuminants in RGB space]{
    \label{fig:supp:CCT:D40_GT}
    \includegraphics[width=0.70\linewidth]{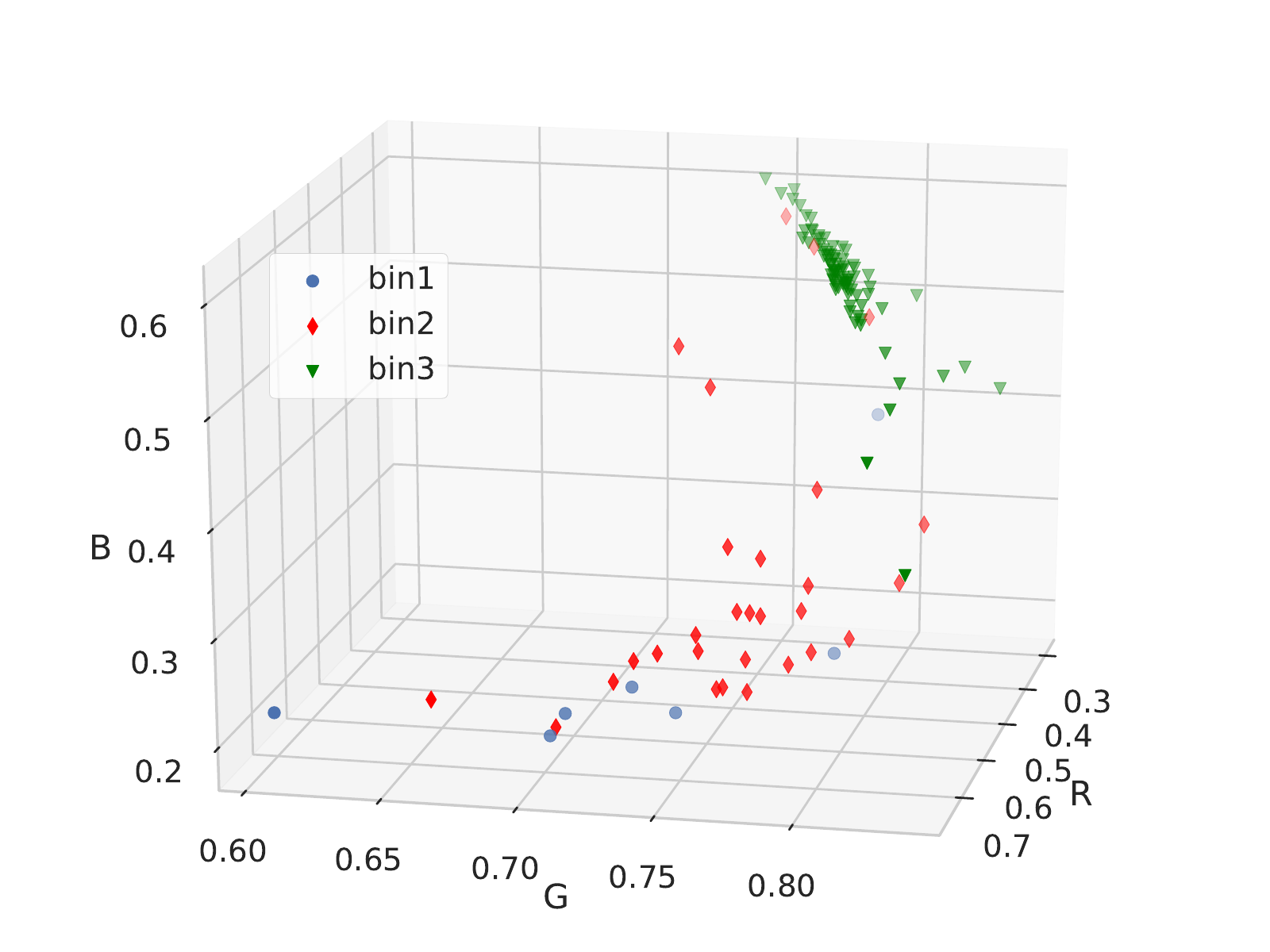}
    }
 \end{center}
\caption{CCT histograms $H_s$, 3 bin example.}
\label{fig:supp:CCT}
\end{figure}

In the main paper, we limit the number of bins to $M{}={}2$, as this allows the broad separation of images into indoor and outdoor illuminants. Here we provide additional examples of CCT histograms and their corresponding clustering of ground-truth (GT) illuminant corrections in RGB space. Results for the $M{}={}3$ bins case, for two cameras (Canon 600D and Nikon D40 from the NUS-9 dataset~\cite{cheng2014illuminant}) are shown in Fig.~\ref{fig:supp:CCT}. 

We first observe that for smaller sized datasets (e.g. Nikon D40 data), there are not enough images ($<10{}={}K$) available in the first bin as shown in~\ref{fig:supp:CCT:D40_hist}. Furthermore, while ground-truth illuminants retain reasonable cluster separability, bin edges between different cameras may be misaligned. Finally, increasing the number of histogram bins also increases training data requirements as each bin requires $10$ to $20$ training images. These results motivate the choice of $M\!=\!2$, as shown in Figure $4$ of the main paper, thus providing distinctive tasks with sufficient images in each bin of the CCT histogram.





\begin{figure}[t]
\centering
    \includegraphics[width=1\linewidth]{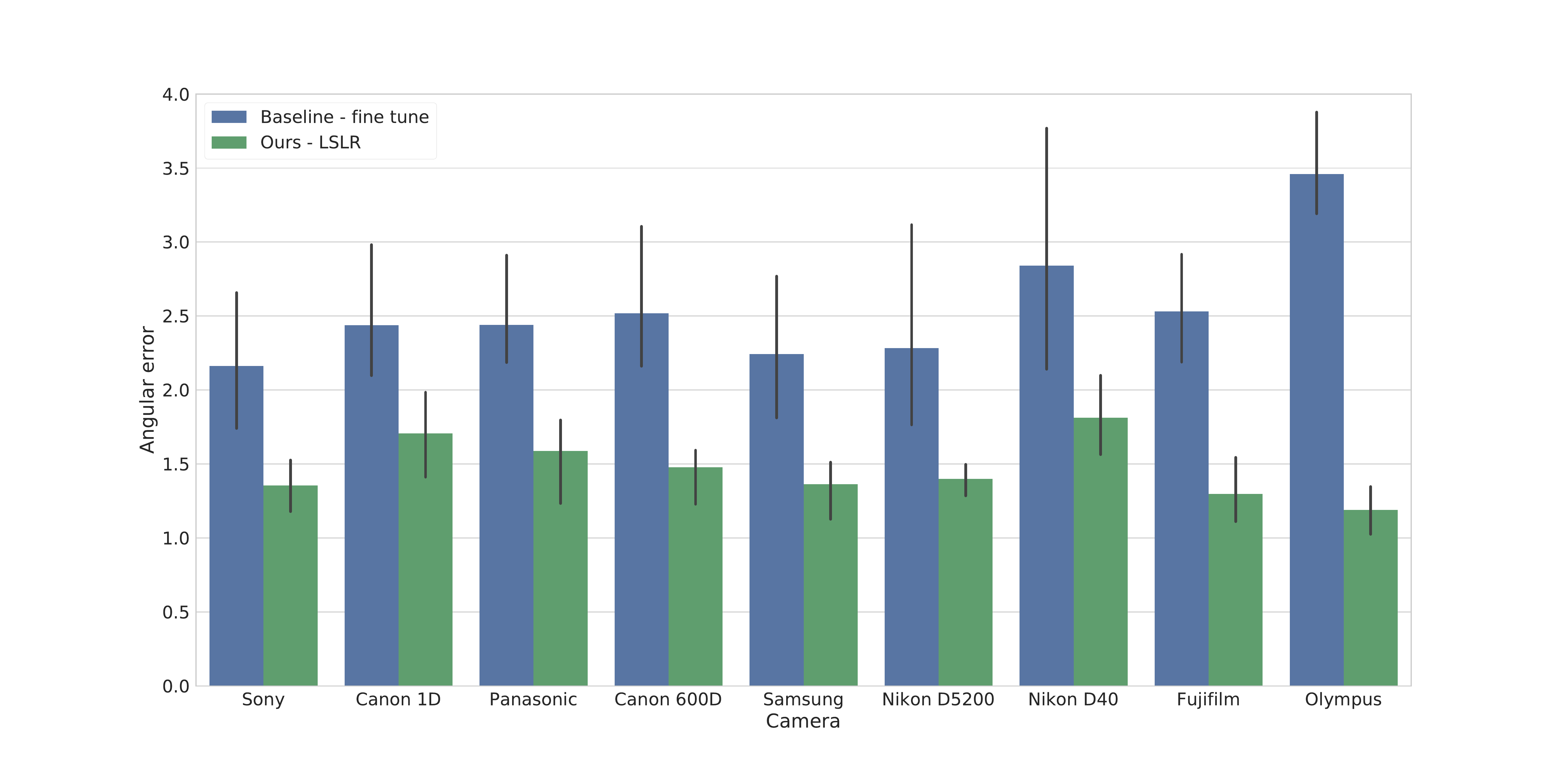}
\caption{Median angular error for each NUS-9 camera, our method (green) and fine-tuned baseline (blue). Cameras are ordered from left to right with increasing gap in performance between baseline and ours.}
\label{fig:supp:med_gain:per_camera_results}
\end{figure}
\begin{figure}[t]
\centering
\subfigure[Median ground-truth RGB correction for all cameras.]{
    \label{fig:supp:med_gain:per_camera_GT}
    \includegraphics[width=0.7\linewidth]{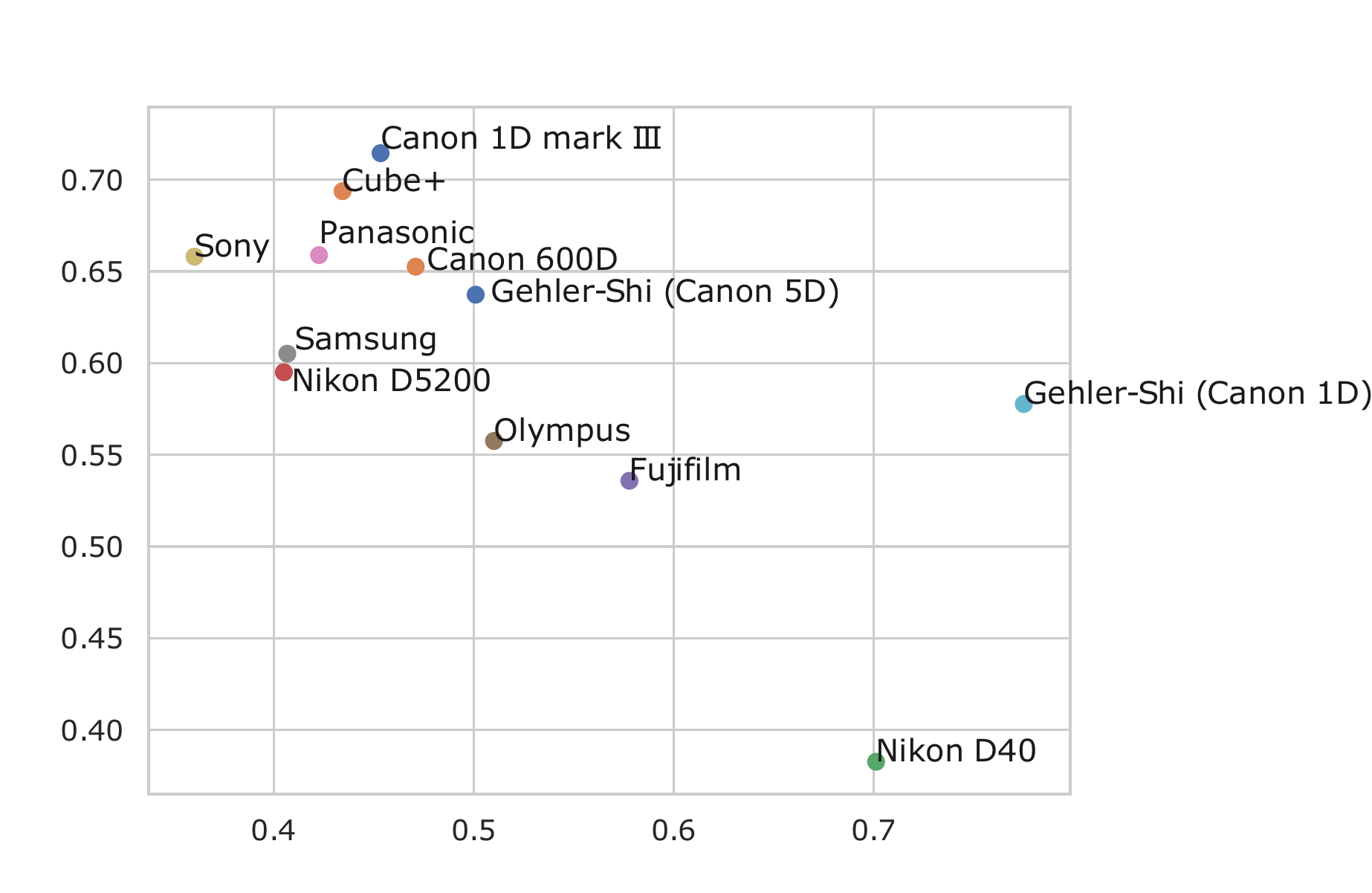}
    }
    \subfigure[Ground-truth RGB distributions for all cameras. Shifts in distributions are linked to camera effects and scene content. The color of each dataset distribution is defined by a colormap linked to the angular errors obtained over all images and cameras. Warmer colors correspond to larger median dataset errors.]{
    \label{fig:supp:med_gain:per_camera_results}
    \includegraphics[width=1\linewidth]{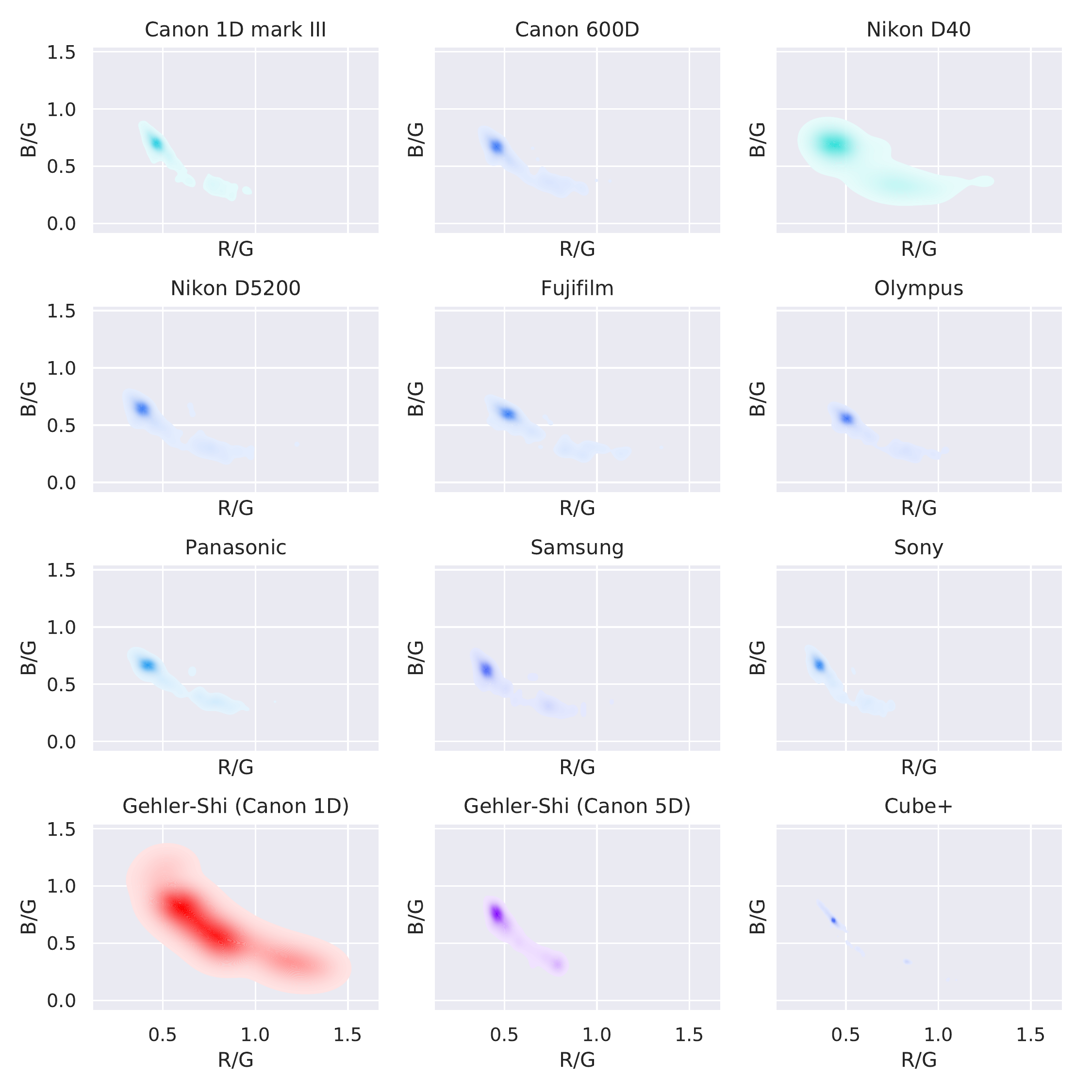}
    }
\caption{Visualisation of variability between distributions of ground-truth RGB illuminant corrections for all cameras.}
\label{fig:supp:med_gain}
\end{figure}

\section{Interpretation of camera specific results}

We visualise the distributions of ground truth RGB corrections per camera in Fig.\ref{fig:supp:med_gain} where datasets with larger median angular errors are shown with warmer colormaps, as well as their respective median value in Fig.\ref{fig:supp:med_gain:per_camera_GT}. 
Ground-truth distributions are linked to differences in scene content and camera (CSS, lens and sensor effects). In particular, camera latent effects can be considered dominant for the NUS dataset cameras due to the matching inter-camera scene content of images. Distribution difference for other cameras can be linked to both scene content and the considered hardware effects and is therefore less comparable.

Linking distributions in Fig.~\ref{fig:supp:med_gain} to the results obtained, we observe that best results are obtained for cameras with compact distributions (ie. similar scene content, as observed for the Cube+ dataset). Furthermore, the very different nature of the Gehler-Shi (Canon 1D) and Nikon D40 distributions suggest that these cameras are more difficult to adapt to. 

In particular, comparative results on the NUS-9 dataset between the fine-tuned baseline (blue) and our approach (green) are shown in Fig.~\ref{fig:supp:med_gain:per_camera_results}. Cameras are ordered from left to right from lower to larger gap between baseline and our method. This shows that our method adapts well to cameras with distributions that are shifted with respect to the majority of NUS-9 cameras (e.g. Fujifilm, Olympus). These cameras exhibit a larger increase in performance with respect to our baseline.
Similarly, we observe one of the larger gaps in performance for the Nikon D40 camera, with images that were acquired later and comprises different scenes and illuminations.

\end{appendices} 

\end{document}